\documentclass{article} % For LaTeX2e
\usepackage{iclr2023_conference,times}

\usepackage{amsmath,amsfonts,bm}

\def\eqref#1{equation~\ref{#1}}
\def\1{\bm{1}}

\def\vmu{{\bm{\mu}}}

\def\vg{{\bm{g}}}
\def\vh{{\bm{h}}}

\def\vw{{\bm{w}}}
\def\vx{{\bm{x}}}
\def\vy{{\bm{y}}}
\def\vz{{\bm{z}}}
\def\vmu{{\bm{\mu}}}
\def\vSigma{{\bm{\Sigma}}}

\newcommand{\gaussx}{\mathcal{N}(\mathbf{\vmu}^{\vz}, \vSigma^{\vz})}
\newcommand{\gaussy}{\mathcal{N}(\mathbf{\vmu}^{\vw}, \vSigma^{\vw})}
\newcommand{\gaussxo}{\mathcal{N}(\mathbf{\vmu}^{\vz}_0, \vSigma^{\vz}_0)}
\newcommand{\gaussyo}{\mathcal{N}(\mathbf{\vmu}^{\vw}_\delta, \vSigma^{\vw}_\delta)}

\def\vmu{{\bm{\mu}}}

\def\vg{{\bm{g}}}
\def\vh{{\bm{h}}}

\def\vw{{\bm{w}}}
\def\vx{{\bm{x}}}
\def\vy{{\bm{y}}}
\def\vz{{\bm{z}}}

\DeclareMathAlphabet{\mathsfit}{\encodingdefault}{\sfdefault}{m}{sl}
\SetMathAlphabet{\mathsfit}{bold}{\encodingdefault}{\sfdefault}{bx}{n}

\newcommand{\R}{\mathbb{R}}

\DeclareMathOperator*{\argmax}{arg\,max}

\usepackage{natbib}
\usepackage{graphicx}
\usepackage{caption}
\usepackage{subcaption}
\usepackage{amsmath}
\usepackage{amssymb}
\usepackage{array}
\usepackage{algorithmic}
\usepackage{algorithm}
\usepackage{xcolor} 
\definecolor{mydarkblue}{rgb}{0,0.08,0.45}
\usepackage[colorlinks,citecolor=mydarkblue,urlcolor=mydarkblue,linkcolor=mydarkblue,filecolor=mydarkblue]{hyperref}
\usepackage{url}
\usepackage{enumitem}
\usepackage{relsize}

\usepackage{multirow}
\usepackage{mdframed}
\usepackage{booktabs}
\usepackage{makecell}
\usepackage{color, colortbl}
\definecolor{Gray}{gray}{0.9}
\definecolor{light-gray}{gray}{0.30}

\usepackage{sistyle}
\SIthousandsep{,}
\usepackage{tabularx}
\newcolumntype{Y}{>{\centering\arraybackslash}X}
\newcommand\todo[1]{}

\usepackage[scaled=.8]{beramono}
\usepackage{xspace}

 \newcommand\cnn{\texttt{cnn\_dailymail}\xspace}
 \newcommand\xsum{\texttt{xsum}\xspace}
 \newcommand\reddit{\texttt{reddit\_tifu}\xspace}
 \newcommand\samsum{\texttt{samsum}\xspace}

 \newcommand\pegasuslarge{$\text{PEGASUS}_\text{LARGE}$\xspace}

\newcommand\prsum{$\text{PR}_{\text{sum}}$}
\newcommand\rougeone{ROUGE-1\xspace}
\newcommand\bleurt{BLEURT\xspace}

\newcommand{\compactparagraph}[1]{\textbf{#1}\quad}

\newcommand{\mc}{\multicolumn}

\definecolor{revisionColor}{rgb}{0.965, 0.341, 0.286}
\newcommand{\revision}[1]{#1}

  \usepackage[compact]{titlesec}
  \titlespacing{\section}{0pt}{1ex}{0ex}
  \titlespacing{\subsection}{0pt}{1ex}{0ex}
  \titlespacing{\subsubsection}{0pt}{0.5ex}{0ex}
\title{Out-of-Distribution Detection and Selective Generation for Conditional Language Models
}

\author{%
Jie Ren$^{1*}$ \quad
Jiaming Luo$^1$ \quad
Yao Zhao$^1$ \quad
Kundan Krishna$^2$\\ 
\textbf{Mohammad Saleh$^1$ \quad
Balaji Lakshminarayanan$^1$   \quad
Peter J Liu$^{1*}$} \\ 
$^1$Google Research \quad
$^2$Carnegie Mellon University, work done while at Google Research\\
$^*$Correspondence to: \texttt{\{jjren, peterjliu\}@google.com}
}

    \iclrfinalcopy % Uncomment for camera-ready version, but NOT for submission.
\begin{document}

\maketitle
\vspace{-1.5em}
\begin{abstract}
Machine learning algorithms typically assume independent and identically distributed samples in training and at test time. Much work has shown that high-performing ML classifiers can degrade significantly and provide overly-confident, wrong classification predictions,  particularly for out-of-distribution (OOD) inputs.
Conditional language models (CLMs) are predominantly trained to classify the next token in an output sequence, and may suffer even worse degradation on OOD inputs as the prediction is done auto-regressively over many steps. Furthermore, the space of potential low-quality outputs is larger as arbitrary text can be generated and it is important to know when to trust the generated output.  
We present a highly accurate and lightweight OOD detection method for CLMs, and demonstrate its effectiveness on abstractive summarization and translation.
We also show how our method can be used under the common and realistic setting of distribution shift for  \emph{selective generation} (analogous to selective prediction for classification) of high-quality outputs, while  automatically abstaining from low-quality ones, enabling safer deployment of generative language models.
\end{abstract}
\vspace{-0.8em}

\section{Introduction}
Recent progress in generative language models \citep{gnmt, radford2019language, bart, t5,  pegasus}  has led to quality approaching human-performance on research datasets and has opened up the possibility of their wide deployment beyond the academic setting. In realistic user-facing scenarios such as text summarization and translation, 
 it should be expected that user provided inputs can significantly deviate from the training data distribution. This violates the independent, identically-distributed (IID) assumption commonly used in evaluating machine learning models. 

Many have shown that performance of machine learning models can degrade significantly and in surprising ways on OOD inputs  \citep{nguyen2014deep,goodfellow2014explaining,ovadia2019can}. For example an image classifier may detect cows in images with very high accuracy on its IID test set but confidently fails to detect a cow when paired with an unseen background \citep{pml2Book,nagarajan2020understanding}.
% discuss medical example %\citet{roy2022does}
This has led to active research on OOD detection
for a variety of domains, including vision and text
but focused primarily on classification.  
\citet{salehi2021unified,bulusu2020anomalous,ruff2021unifying} provide comprehensive reviews on this topic.

%OOD detection and the effect of distribution shift on the output quality of CLMs is relatively under-explored.
Conditional language models are typically trained given input sequence $\vx=x_1 \dots x_L$ to auto-regressively generate the next token in a sequence $\vy = y_1\dots y_T$ as a classification over the token-vocabulary $V$,  $p_\theta(\vy|\vx) = \prod_{t=1}^T p_{\theta}(y_t | y_{<t}, \vx)$,  $y_t \in V$.
%, and $T$ is the sequence length. 
%The negative log-likelihood of the output sequence averaged over tokens, $-\frac{1}{L} \sum_{t=1}^{L} \log p(y^{t}|y^{<t}, \vx)$, is called perplexity. Perplexity is the most commonly used summary statistic for the output sequence of conditional generative model. It is analogous to the softmax probability $p(y|\vx)$ in classification problem.
Consequently,
the perils of out-of-distribution are arguably more severe as (a) errors propagate and magnify through auto-regression, and (b) the space of low-quality outputs is greatly increased as arbitrary text sequences can be generated.
Common errors from text generation models include disfluencies \citep{Holtzman2020The} and factual inaccuracies \citep{goodrichfactuality, maynez-etal-2020-faithfulness}.
 \revision{A common failure case we observed in abstractive summarization is for the model to output ``All images are copyrighted'' as the summary for news articles from a publisher (CNN) different than what it was trained on (BBC) (see Figure \ref{fig:low_quality_ex_1}).}
%and the ability to selectively generate high-quality output while abstaining from generating bad output is desirable. 

In this work, we propose OOD detection methods for CLMs using abstractive summarization and translation as case studies. 
Similar to classification, we show in Section \ref{sec:perplexity_for_ood} that CLMs have untrustworthy likelihood estimation on OOD examples, making perplexity a poor choice for OOD detection. 
In Section \ref{sec:ood_detection_proposed_method}, we propose a highly-accurate, simple, and lightweight OOD score based on the model’s input and output representations (or embeddings) to detect OOD examples, requiring negligible additional compute beyond the model itself. 

While accurate OOD detection enables the conservative option of abstaining from generation on OOD examples, 
  it may be desirable to generate on known near-domain data, e.g. generate summaries for articles from news publishers that differ from our fine-tuning set. Thus the ability to selectively generate where the model is more likely to produce higher-quality outputs, enables safer deployment of conditional language models. %safer off-label (in the traditional machine learning sense) use of the model. 
We call this procedure \emph{selective generation}, analogous to the commonly used term \emph{selective prediction} in classification \citep{chow1957optimum,bartlett2008classification,geifman2017selective}.
In Section \ref{sec:quality}, we show that while model perplexity is a reasonable choice for \revision{performing selective generation with} in-domain examples, combining with our OOD score works much better when the input distribution is shifted.
%\todo{introduce the terminology "selective generation" here and draw the analogy to "selective prediction". }

In summary, our contributions are:
\begin{itemize}
\item Propose lightweight and accurate scores  derived from a CLM's embeddings for OOD detection, significantly outperforming baselines on abstractive summarization and translation tasks, without the need for a separate detection model.
% \todo{add concreate \% improvment}
\item Show that model perplexity can be an unreliable signal for quality estimation on OOD examples, but combined with our OOD scores can be used effectively to selectively generate higher-quality outputs while abstaining on lower ones.
\item Propose an evaluation framework for OOD detection and selective generation for CLMs, including human quality ratings for summarization.
\end{itemize}

\section{OOD Detection in Conditional Language Models}

The maximum softmax probability (MSP), $p(y|\vx)$, $y=\argmax_{k=1,\dots,K} p(k|\vx)$ is a simple, commonly used OOD score for $K$-class classification problem \citep{hendrycks2016baseline,lakshminarayanan2017simple}.
For CLMs, the perplexity, which is monotonically related to the negative log-likelihood of the output sequence averaged over tokens $-\frac{1}{T} \sum_{t=1}^{T} \log p(y_{t}|y_{<t}, \vx)$ is a natural OOD score to consider, and analogous to the negative MSP in classification \revision{because both are based on softmax probabilities}.
%The negative log-likelihood of the output sequence averaged over tokens, $-\frac{1}{L} \sum_{t=1}^{L} \log p(y^{t}|y^{<t}, \vx)$, is called perplexity. Perplexity is the most commonly used summary statistic for the output sequence of conditional generative model. It is analogous to the softmax probability $p(y|\vx)$ in classification problem.
We first study how well the perplexity performs for OOD detection tasks.

\subsection{Perplexity is ill-suited for OOD detection}
\label{sec:perplexity_for_ood}
\begin{figure}[h]
\centering
\vspace{-0.5em}
\begin{subfigure}[t]{0.40\textwidth} % changed for arxiv
\centering
\caption{Summarization}
\vspace{-0.2em}
\includegraphics[width=0.85\textwidth]{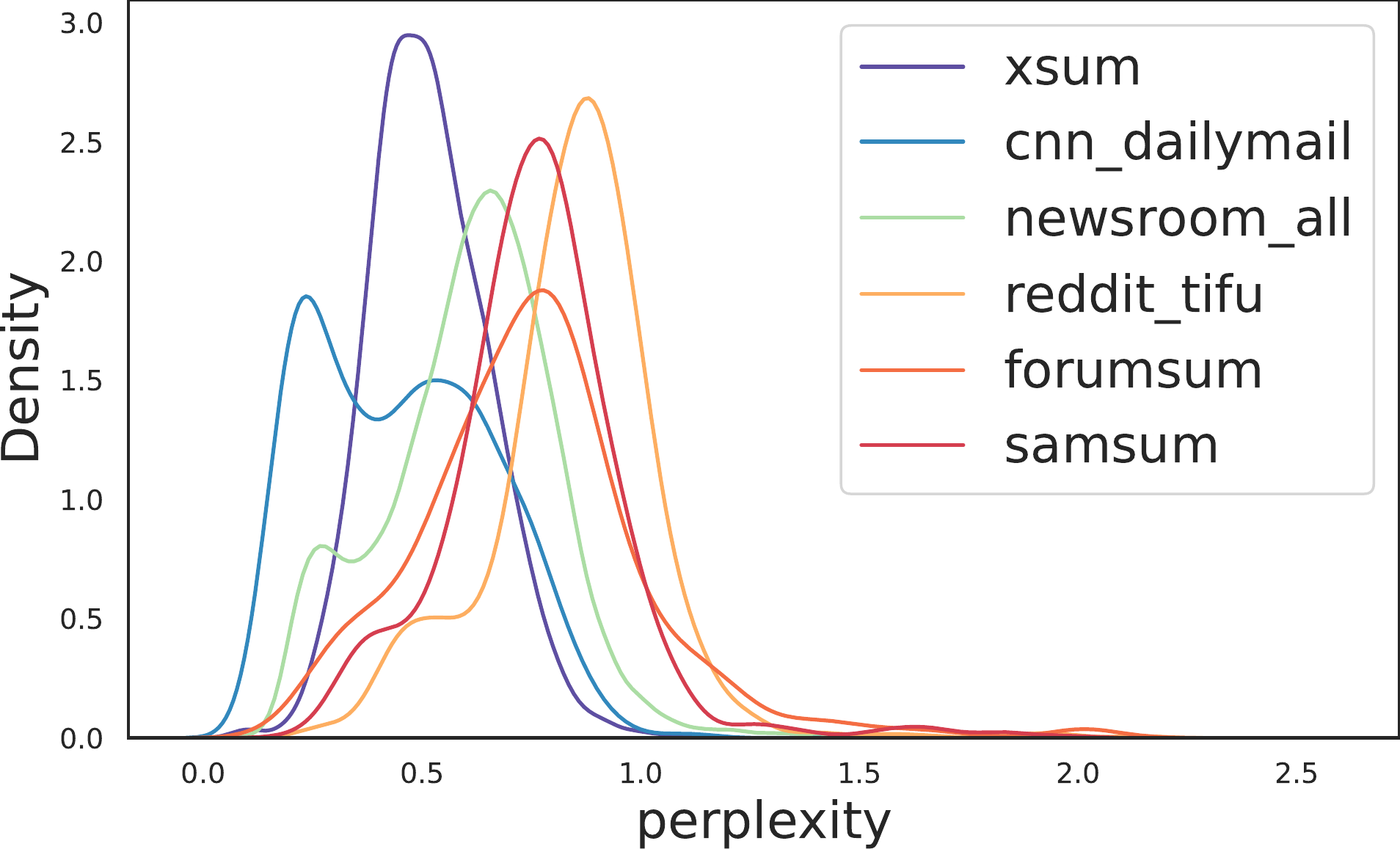} 
\end{subfigure} 
\vspace{-0.5em}
\begin{subfigure}[t]{0.40\textwidth} % changed for arxiv
\centering
\caption{Translation}
\vspace{-0.2em}
\includegraphics[width=0.85\textwidth]{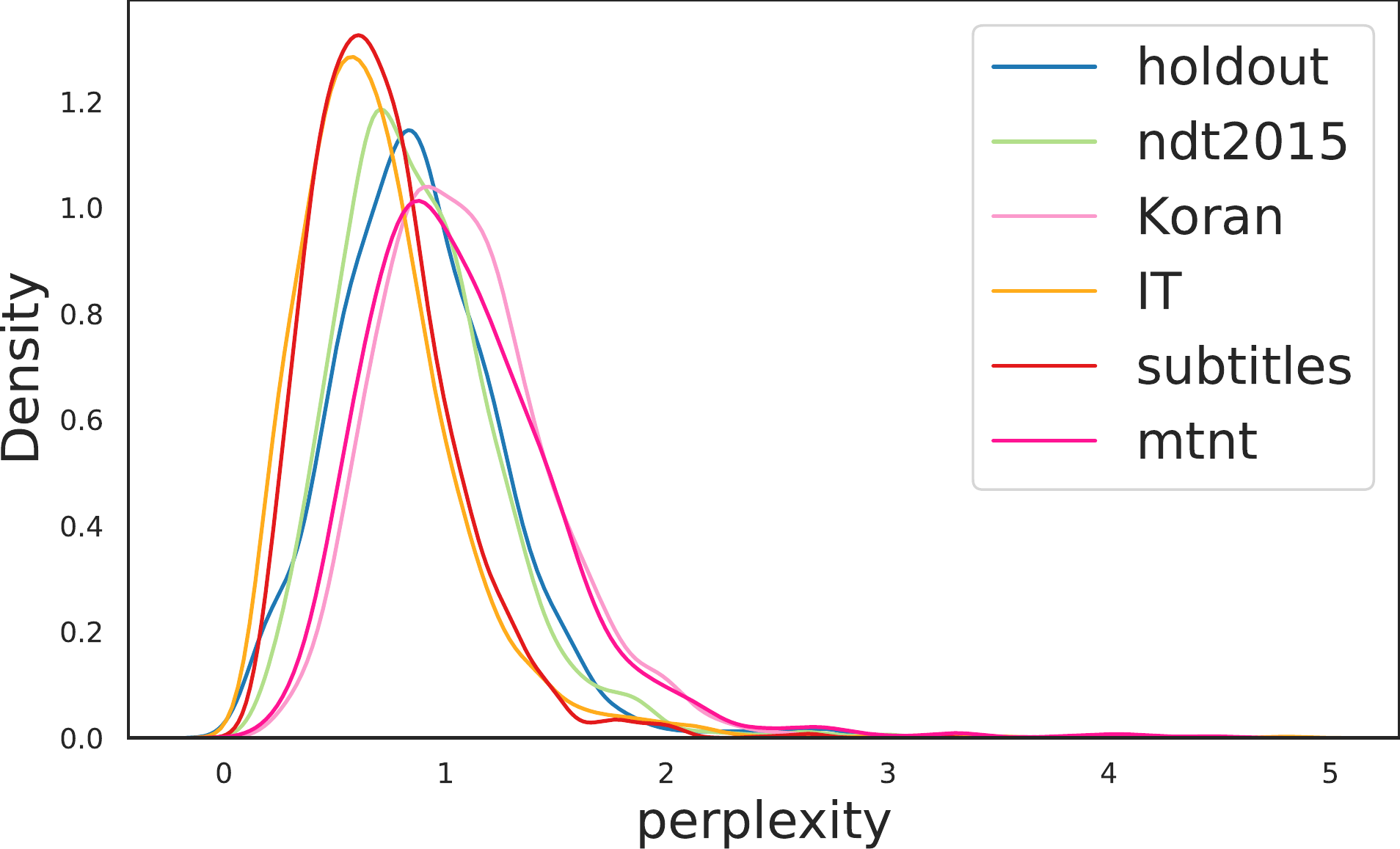}
\end{subfigure} 
% \vspace{-0.1em}
% \vspace{-0.8em}
% \begin{subfigure}[t]{0.30\textwidth} % changed for arxiv
%     \centering
%     \caption{Correlation (ppx, rouge1) decreases as OOD increases}
%     % \vspace{-0.1em}
%     \includegraphics[width=0.95\textwidth]{figures/corr_ppx_decrease_as_ood_increase.pdf}
%     \label{fig:train_vs_auc}
% \end{subfigure} 
\caption{
Perplexity scores density of a CLM trained on (a) \texttt{xsum} for summarization, and (b) \texttt{WMT} for translation, evaluated on other datasets/domains. Perplexity is not well suited for OOD detection due to significant overlap between in-domain and OOD scores. 
%A subset of translation datasets is shown in (b) for simplicity.
%For summarization, perplexity assigns cnn\_dailymail even lower scores than in-domain xsum, which is not expected. For translation, in-domain and OOD datasets have the similar distribution under perplexity, largely overlapping to each other.
}
\label{fig:ppx_hist}
\end{figure}

%, since perplexity score can be regarded as a natural extension of the the MSP score in conditional generative models. 
%the output is a sequence of predicted tokens $\{y_0 y_1 \dots, y_L\}$, and each has a conditional softmax probability $p(y^{t}|y^{<t}, \vx)$, $y^{t}=\argmax_{v\in V} p(v|y^{<t}, \vx)$,
%$t=1,\dots,L$ and $V$ is the token vocabulary. 
% the perplexity score,
% which is the average negative log-probability over tokens in the sequence 
% $-\frac{1}{L} \sum_{t=1}^{L} \log p(y^{t}|\hat{y}^{<t}, \vx)$, 
% can be regarded as a natural extension of the the MSP score for OOD detection in conditional generative models. 
% We first study how well the perplexity score performs for OOD detection tasks, using summarization and translation as two different tasks. 
% Perplexity score is not well suited for OOD detection. 
In Figure \ref{fig:ppx_hist}, we compare the distribution of perplexity of (a) a summarization model and (b) a translation model trained on in-domain dataset and evaluated on multiple OOD datasets, respectively.
For summarization, a model is trained on \texttt{xsum} and evaluated on other news datasets including \texttt{cnn\_dailymail} and \texttt{newsroom} as near-OOD datasets, and forum (\texttt{forumsum}) and dialogue (\texttt{samsum} and \texttt{reddit\_tifu}) datasets as far-OOD (see Section \ref{sec:exp_detection} for details). 
The perplexity distributions overlap significantly with each other even though the input documents are significantly different. Furthermore, perplexity assigns \texttt{cnn\_dailymail} even lower scores than the in-domain \texttt{xsum}. 

For translation, the model is trained on \texttt{WMT15} dataset and evaluated on other \texttt{WMT} test splits \citep{bojar-etal-2015-findings}, \texttt{OPUS100} \citep{aulamo-tiedemann-2019-opus}, and \texttt{MTNT} \citep{michel-neubig-2018-mtnt}. The in-domain and OOD datasets perplexity densities overlap even more.
Overall, these results suggest that perplexity is not well suited for OOD detection. 

% \todo{prism}

\subsection{Detecting OOD using CLM's embeddings}
\label{sec:ood_detection_proposed_method}
\begin{figure}[htb]
\centering
% \vspace{-1em}
\includegraphics[width=0.62\textwidth]{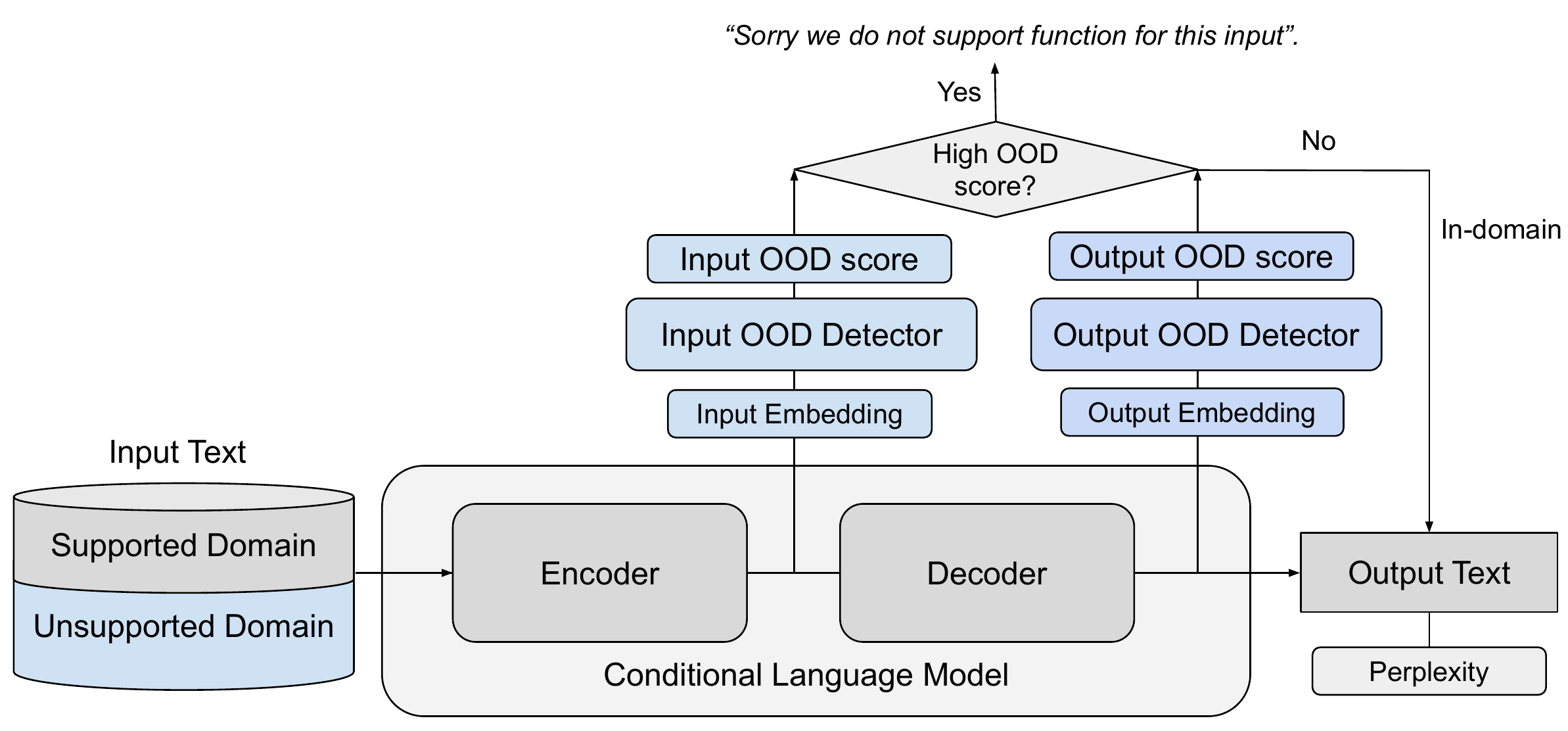}
% \vspace{-1em}
\caption{The proposed OOD detector based on input and output embeddings. 
}
% (b) The proposed combined quality score based on OOD scores and Perplexity that predicts output quality (rouge, bleu, huamn eval etc.) better than individual scores. \todo{Put figures next to each other?}
\label{fig:diagram}
\end{figure}

Given a trained conditional language model, we propose using the input and output representations/embeddings computed as part of the inference/generation process to detect OOD examples.
In this work, we use Transformer encoder-decoder
%(denoted by $\text{enc}(\cdot)$ and $\text{dec}(\cdot)$) 
models and obtain the \textbf{input embedding} $\vz$ by averaging the encoder's final-layer hidden state vectors  $\vh_i \in \R^{d}$ ($d$ is the hidden dimension)
%$\text{encoder}(\vx) = \mH \in \R^{L \times D}$ 
corresponding to the input sequence token $x_i$.
To obtain the \textbf{output embedding} $\vw$ we average the decoder's final-layer hidden state vectors $\vg_i \in \R^{d}$ corresponding to the output token $y_i$. Thus
% $\vw=\frac{1}{T}\sum^T_{i=1} \vg_i$.
\begin{equation*}
    \vz:=\frac{1}{L}\sum^L_{i=1} \vh_i \quad\quad \vw:=\frac{1}{T}\sum^T_{i=1} \vg_i, \quad\quad \vz, \vw \in \R^d
\end{equation*} 
where $L$ and $T$ are the input and output sequence lengths respectively.
%%$\text{encoder}(\vx) = \mH \in \R^{L \times D}$ 
%corresponding to the input sequence tokens $ . 
%To obtain the \textbf{output embedding} $\vw=\frac{1}{T}\sum^T_{i=1} \text{dec}_i(\vy)$, we average the decoder final-layer hidden state vectors corresponding to the output tokens $\vy=y_1\dots y_T$.
%This is illustrated in Figure \ref{fig:diagram}. 
Figure \ref{fig:diagram} illustrates the idea.

Intuitively, if the embedding of a test input or output is far from the embedding distribution of the training data, it is more likely to be OOD. One way of measuring this distance is to fit a Gaussian,
$\mathcal{N}(\bm{\mu}, \bm{\Sigma})$, $\bm{\mu} \in \R^d, \bm{\Sigma} \in \R^{d\times d}$,
to the training embeddings and use the
\emph{Mahalanobis distance (MD)}:
\begin{equation*}
    \text{MD}(x; \bm{\mu}, \bm{\Sigma}):=(x-\bm{\mu})^T \bm{\Sigma}^{-1}(x-\bm{\mu}),
\end{equation*}
This has been used for OOD detection using the representations from classification models \citep{lee2018simple} and computing the distances to class-conditional Gaussians.

Unlike classification, which has class labels, in conditional language modeling we have paired input and output text sequences. 
We fit one Gaussian on the training input embeddings, $\gaussx$, and a second Gaussian on the embeddings of the training ground-truth outputs,  $\gaussy$.

For a test input and output embedding pair $(\vz_{\text{test}}, \vw_{\text{test}})$, the input MD is computed as 
\begin{align*}
\text{MD}_{\text{input}}({\vz_{\text{test}}}) :=
    \text{MD}(\vz_{\text{test}}; \bm{\vmu^\vz}, \bm{\vSigma^\vz})
 \tag{\footnotesize \textbf{Input MD OOD score}}
\end{align*}
The output MD  is computed similarly:
\begin{align*}
\text{MD}_{\text{output}}({\vw_{\text{test}}}) :=
    \text{MD}(\vw_{\text{test}}; \bm{\vmu^\vw}, \bm{\vSigma^\vw})
 \tag{\footnotesize \textbf{Output MD OOD score}}
\end{align*}

Mahalanobis distance is equivalent to computing a negative log-likelihood of the Gaussian distribution (up to a constant and a scalar), i.e. $-\log p(\vz)=\frac{d}{2}\log(2\pi)+\frac{1}{2}\log|\vSigma|+\frac{1}{2}(\vz-\vmu)^T \vSigma^{-1}(\vz-\vmu) = \texttt{const.} + \frac{1}{2} \text{MD}(\vz)$. 
\citet{ren2019likelihood} showed that normalizing the likelihood with the likelihood of a background model works better for OOD detection. In a similar vein,  
\citet{ren2021simple} proposed an analogous \textbf{Relative Mahalanobis Distance (RMD)} for classification: using the relative distance between the class-conditional Gaussians  and a single background Gaussian using data from all classes. 
That method cannot be directly applied for CLMs because outputs are not just class labels.  
%  while lot of OOD detection scores have been proposed for classification, there hasn’t been a comprehensive evaluation of which methods work best for OOD detection in conditional language models.
Thus in this work, we extend the RMD idea to conditional language models, 
\begin{align}
\text{RMD}_{\text{input}}({\vz_{\text{test}}}) := \text{MD}_{\text{input}}({\vz_{\text{test}}}) - \text{MD}_0({\vz_{\text{test}}}), \tag{\footnotesize \textbf{Input RMD OOD score}}
\label{eq:ratio_input}
\end{align}
where $\text{MD}_0({\vz_{\text{test}}}):=\text{MD}(\vz_{\text{test}}; \mathbf{\vmu}^{\vz}_0,  \vSigma^{\vz}_0)$
is the MD  to a background Gaussian $\gaussxo$, fit using a large, broad dataset to approximately represent all  domains. 
In practice, we use \texttt{C4}, a large Common Crawl-based English dataset \citep{t5}\footnote{\url{https://www.tensorflow.org/datasets/catalog/c4}} and \texttt{ParaCrawl}'s English-French dataset  \citep{banon-etal-2020-paracrawl}\footnote{\url{https://www.tensorflow.org/datasets/catalog/para_crawl}}, as the data for fitting the background distributions for summarization and translation in our experiments, respectively. 

While we use the ground-truth outputs to fit $\gaussy$, we decode outputs from the trained CLMs and use those output embeddings to fit the background output Gaussian, $\gaussyo$. 
\begin{align*}
\text{RMD}_{\text{output}}({\vw_{\text{test}}}) := \text{MD}_{\text{output}}({\vw_{\text{test}}}) - \text{MD}_\delta({\vw_{\text{test}}}), \tag{\footnotesize \textbf{Output RMD OOD score}}
\label{eq:ratio_output}
\end{align*}
where $\text{MD}_\delta({\vw_{\text{test}}}):=\text{MD}(\vw_{\text{test}}; \mathbf{\vmu}^{\vw}_\delta,  \vSigma^{\vw}_\delta)$
 is the MD  to the decoded output background distribution $\gaussyo$.
 See Algorithm \ref{alg:rmd_fit} and \ref{alg:rmd_infer} for the detailed steps.
 \revision{Using decoded outputs serves two purposes: (1) We do not require supervised data (e.g. document-summary pairs) to fit the background Gaussian. (2)  Decoded outputs may exhibit increased deficiencies that result from running the model on out-of-distribution data, which provides greater contrast with the in-domain ground-truth labels.}

The RMD score can be regarded as a background contrastive score that indicates how close the test example is to the training domain compared to the background domains.
A negative score suggests the example is relatively in-domain, while a positive score suggests the example is OOD. A higher score indicates greater OOD-ness.

\compactparagraph{Binary classifier for OOD detection}
Since we have \revision{explicitly} defined two classes, in-domain and background/general domain, another option
is to train a binary classifier to discriminate embeddings from the two classes.
We train a logistic regression model and use the un-normalized logit for the background as an OOD score. 
The \textbf{Input Binary logits OOD score} uses the input embeddings as features,
whereas the \textbf{Output Binary logits OOD score} uses the decoded output embeddings as features.
A higher score suggests higher likelihood of OOD. The preferred use of the logits over probability was also recommended by previous OOD studies for classification problems~\citep{hendrycks2019scaling}.  
Though RMD is a generative-model based approach and the binary classifier is a discriminative model, 
we show that RMD is a generalized version of binary logistic regression and can be reduced to a binary classification model under certain conditions (see Section \ref{sec:conect_rmd_binary} for details).

\section{Experiments: OOD detection}
\label{sec:exp_detection}
\subsection{Experiment setup}
\label{sec:experiment_setup}
We run our experiments using Transformer \citep{vaswani2017attention} encoder-decoder models trained for
abstractive summarization and translation. Below we specify the dataset used for training/fine-tuning (i.e. in-domain) and the OOD datasets.
% \item input embedding: average of the token embeddings
% \item output embedding

In the case of summarization, OOD datasets can be intuitively  categorized as \emph{near} or \emph{far OOD} based on the nature of the documents. For example, news articles from different publishers may be considered as sourced from different distributions, but are closer than  news articles are to  dialogue transcripts. 
\revision{We also quantitatively showed that using $n$-gram overalp analysis in Table \ref{tab:summary_ngram}.}
In contrast, the translation datasets we use consist of English-French sentence pairs with less variation between datasets due to the shorter length of sentences.

\compactparagraph{Summarization model} We \revision{fine-tuned \pegasuslarge \citep{pegasus}} on the \texttt{xsum} \citep{narayan-etal-2018-dont} dataset, consisting of BBC News articles with short, abstractive summaries.

\compactparagraph{Summarization datasets}
We use \num{10000} examples from \texttt{xsum} and \texttt{C4} training split to fit in-domain/foreground and background Gaussian distributions, respectively. 
For test datasets, we have
 \texttt{cnn\_dailymail} \citep{hermann2015cnndm, see-etal-2017-get}, news articles and summaries from CNN and DailyMail;
  \texttt{newsroom} \citep{Grusky_2018}, article-summary pairs from 38 major news publications;
\texttt{reddit\_tifu} \citep{reddit_tifu}, informal stories from sub-reddit TIFU with author written summaries of very diverse styles;
\texttt{samsum} \citep{gliwa-etal-2019-samsum} and \texttt{forumsum} \citep{khalman2021forumsum}, high-quality summaries of casual dialogues.

\compactparagraph{Translation model} We train a Transformer base model~\citep{vaswani2017attention} with embedding size 512 on WMT15 English-French~\citep{bojar-etal-2015-findings}.
% We additionally set aside \todo{one million?} translations from the training split to form a \texttt{holdout} subset that is 
% unseen during training and reserved only for analysis.
The model is trained with \texttt{Adafactor} optimizer~\citep{shazeer2018adafactor} for 2M steps with 0.1 dropout and 1024 batch size. 
Decoding is done using beam search with 10 beam size and $\alpha=0.6$ length normalization~\citep{wu2016google}.
The best checkpoint scores 39.9 \texttt{BLEU} on newstest2014.
%\footnote{Obtained through \texttt{SacreBLEU}~\citep{post-2018-call}. In a comparable setting, \cite{vaswani2017attention} reported 38.1 BLEU, although not using \texttt{SacreBLEU}.} 

\compactparagraph{Translation datasets}
We use \num{100000} examples from \texttt{WMT15} En-Fr and the same number of examples from \texttt{ParaCrawl} En-Fr to fit the foreground and background Gaussians, respectively. 
For test, we use newstest2014 (\texttt{nt14}), newsdiscussdev2015 (\texttt{ndd15}), and 
newsdiscusstest2015 (\texttt{ndt15}) from WMT15~\citep{bojar-etal-2015-findings} and the
\texttt{law}, \texttt{Koran}, \texttt{medical}, 
\texttt{IT}, and subtitles (\texttt{sub}) subsets from
OPUS~\citep{tiedemann-2012-parallel,aulamo-tiedemann-2019-opus}. We also use the English-French test set of MTNT \citep{michel-neubig-2018-mtnt}, consisting of noisy comments from Reddit.

%More specifically for WMT15, we use the \texttt{holdout} split,

\compactparagraph{Evaluation metric} We use the area under the ROC curve (AUROC) between the in-domain test data as negative and the OOD test data as positive sets to evaluate and compare the OOD detection performance. AUROC 1.0 means a perfect separation, and 0.5 means the two are not distinguishable.

\compactparagraph{Baseline methods} We compare our proposed OOD scores with various baseline methods, including (1) the model perplexity score, (2) the embedding-based Mahalanobis distance. In addition, we also compare with (3) Natural Language Inference (NLI) score \citep{honovich-etal-2022-true-evaluating} for summarization, and (4) COMET \citep{rei2020comet} and (5) Prism \citep{thompson2020automatic} for translation.
NLI score measures the factual consistency by treating the input document as a premise and the generated summary as a hypothesis. 
Both COMET and Prism are quality estimation metrics designed to measure
translation quality without access to a human reference. More specifically,
COMET finetunes the large XLM-R model~\citep{conneau-etal-2020-unsupervised} on human evaluation data, and Prism is the perplexity
score from a multilingual NMT model trained on 99.8M sentence pairs in 39 languages.

\subsection{Results}
\begin{table}[t]
\caption{AUROCs for OOD detection. For summarization task (a), \texttt{cnn\_dailymail} and \texttt{newsroom} are considered as near shift OOD since it shares news topics as \texttt{xsum}, and \texttt{reddit\_tifu}, \texttt{forumsum}, and \texttt{samsum} are far shift OOD. For translation (b), WMT dataset contains various test WMT datasets collected from different years, OPUS contains five different domains (the degree of shift varies), and MTNT contains noisy data from Reddit.}
\begin{subtable}[c]{1\textwidth}
\centering
\vspace{-0.8em}
\subcaption{Summarization}
\scriptsize
\begin{tabular}{lcccccc}
\toprule
                 & \multicolumn{2}{c}{Near Shift OOD}       & \multicolumn{3}{c}{Far Shift OOD}      \\
                 \cmidrule(lr){2-3} \cmidrule(lr){4-6} 
Measure          & \texttt{cnn\_dailymail} & \texttt{newsroom} & \texttt{reddit\_tifu} & \texttt{forumsum} & \texttt{samsum} \\ \toprule
\multicolumn{6}{c}{\textsc{Input OOD}}                                        \\ 
% GPT-3 likelihood (baseline) &   TBD     &    TBD    &        TBD      &     TBD     &   TBD  \\   
MD          & 0.651          & 0.799         & 0.974        & 0.977    & 0.995 \\
RMD   & 0.828          & 0.930         & \underline{0.998}        & 0.997    & \textbf{0.999}  \\
Binary logits    & \textbf{0.997}          & 0.959         & \textbf{1.000}        & \textbf{0.999}    & 0.998  \\ \midrule
\multicolumn{6}{c}{\textsc{Output OOD}}   \\ 
Perplexity (baseline)      & 0.424          & 0.665         & 0.909        & 0.800    & 0.851  \\
NLI score (baseline)           & 0.440          & 0.469         & 0.709        & 0.638    & 0.743 \\ 
MD           & 0.944          & 0.933         & 0.985        & 0.973    & 0.985  \\
RMD  & 0.958          & \underline{0.962}         & \underline{0.998}        & 0.993    & \underline{0.998}  \\
Binary logits     & \underline{0.989}          & \textbf{0.982}         & \textbf{1.000}        & \underline{0.998}    & 0.997  \\ \bottomrule 
\end{tabular}
\vspace{2em}
\end{subtable}
\begin{subtable}[c]{1\textwidth}
\centering
\vspace{-1em}
\subcaption{Translation}
\scriptsize
\begin{tabular}{lccccccccc}
\toprule
& \mc{3}{c}{WMT}  & \mc{5}{c}{OPUS}        &   \multicolumn{1}{l}{\multirow{2}{*}{MTNT}}  \\
\cmidrule(lr){2-4} \cmidrule(lr){5-9} 
Measure               & \texttt{nt2014} & \texttt{ndd2015} & \texttt{ndt2015} & \texttt{law} & \texttt{medical} & \texttt{Koran} & \texttt{IT} & \texttt{sub}  & \multicolumn{1}{l}{} \\ 
\toprule
\mc{10}{c}{\textsc{Input OOD}} \\
MD       & 0.534           & 0.671            & 0.670            & 0.511        & 0.704            & 0.737          & 0.828       & 0.900            & 0.668  \\
RMD  & 0.798           & \underline{0.866}            & 0.863            & 0.389        & \underline{0.840}            & \underline{0.957}          & \textbf{0.959}       & \textbf{0.969}        & \underline{0.943}      \\
Binary logits  & \textbf{0.864}           & \textbf{0.904}            & \textbf{0.904}            & 0.485        & 0.813            & \textbf{0.963}          & 0.928       & 0.950      & \textbf{0.963}        \\ 
\midrule
\mc{10}{c}{\textsc{Output OOD}} \\
Perplexity (baseline) & 0.570           & 0.496            & 0.494            & 0.392        & 0.363            & 0.657          & 0.343       & 0.359        & 0.633      \\
COMET (baseline)     & 0.484           & 0.514            & 0.525            & 0.435        & 0.543            & 0.632          & 0.619       & 0.518        & 0.724      \\
Prism (baseline)    & 0.445           & 0.504            & 0.505            & 0.459        & 0.565            & 0.716          & 0.604       & 0.577        & 0.699      \\
MD       & 0.609           & 0.733            & 0.739            & 0.482        & 0.784            & 0.838          & 0.900       & 0.935        & 0.794      \\
RMD  & 0.786           & 0.858            & 0.861            & 0.355        & \textbf{0.845}            & 0.939          & \underline{0.951}       & \underline{0.959}      & 0.922        \\
Binary logits & \underline{0.822}           & 0.860            & \underline{0.865}            & 0.507        & 0.783            & 0.942          & 0.890       & 0.910        & 0.931      \\ \bottomrule
\end{tabular}
\vspace{-1em}
\end{subtable}
\label{tab:ood_aucroc}
\end{table}

\compactparagraph{RMD and Binary classifier are better at OOD detection than baselines} Table \ref{tab:ood_aucroc} shows the AUROCs for OOD detection on the (a) summarization and (b) translation datasets. Overall, our proposed OOD scores RMD and Binary logits outperform the baselines with high AUROCs (above 0.8).
The commonly used output metrics, perplexity, NLI, COMET and Prism, have generally low AUROC scores (many have values around 0.5-0.6), suggesting they are not suited for OOD detection. 
Interestingly, we noticed that the output OOD scores perform better for summarization, while the input OOD scores perform better for translation.
One possible reason is that when summarization outputs are low-quality (e.g. producing repeated text or irrelevant summaries) they look very different than reference summaries, making OOD output score more sensitive to the contrast. %\todo{reword following:} One possible reason is that the input document for summarization task have various lengths and formats that could potentially confounds the input OOD score slightly, while the output is mostly one sentence summary, which could make it easy for fair comparison. For translation, there is no such effect because the input and output are both one sentence.

\begin{figure}[!htb]
\centering
\begin{subfigure}[t]{0.40\textwidth} % changed for arxiv
\centering
% \vspace{-0.05em}
\includegraphics[width=0.9\textwidth]{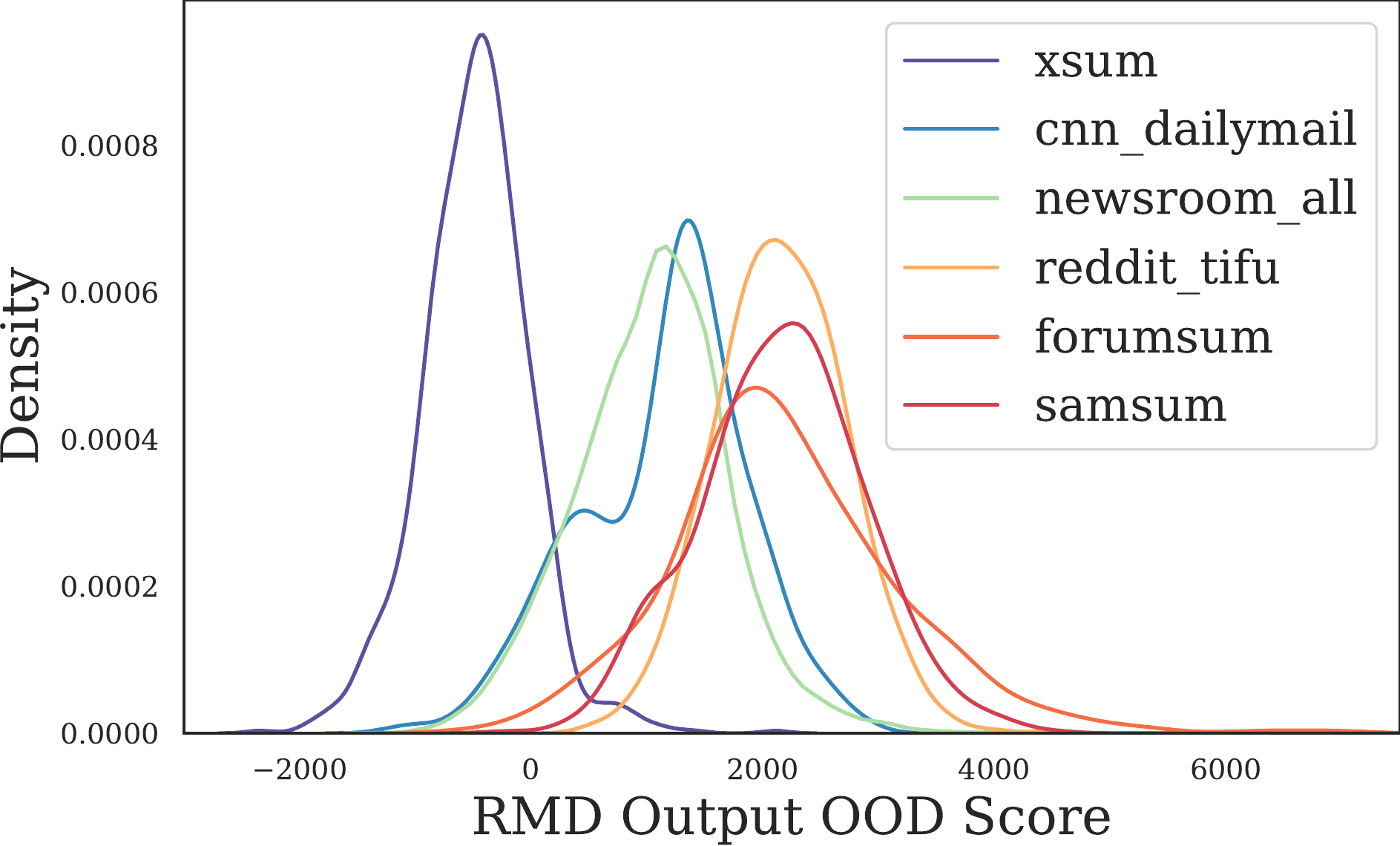} 
\end{subfigure} 
% \vspace{-0.8em}
\begin{subfigure}[t]{0.40\textwidth} % changed for arxiv
\centering
% \vspace{-0.05em}
\includegraphics[width=0.9\textwidth]{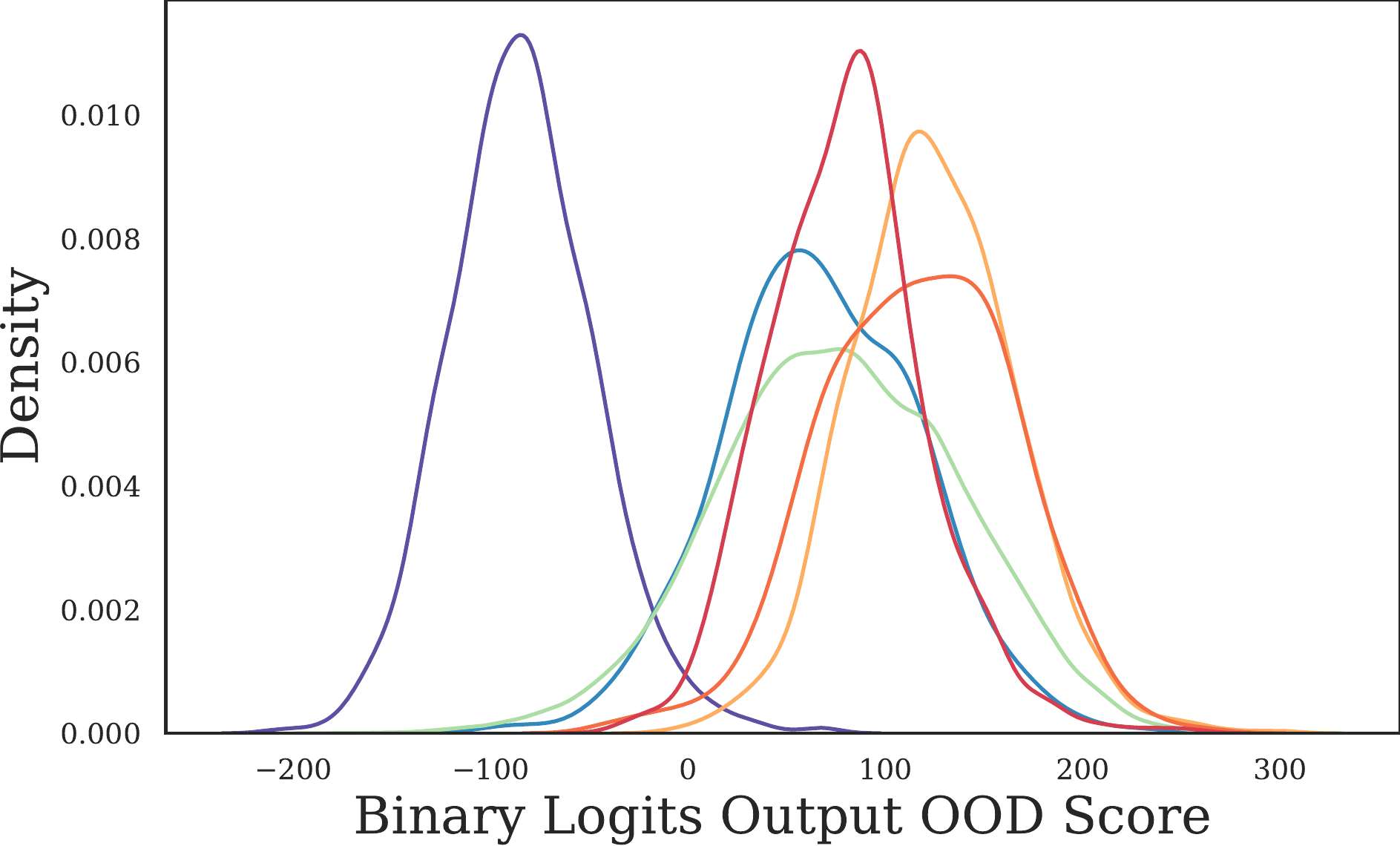}
\end{subfigure} 
% \vspace{-0.1em}
% \vspace{0.8em}
\caption{Density of  RMD (left) and Binary logits (right) OOD scores evaluated on summarization datasets. RMD is better at distinguishing near-OOD from far-OOD.}
\label{fig:rmd_vs_binary}
\end{figure}

Though RMD and Binary logits OOD scores both perform well at OOD detection, \textbf{RMD OOD score is better at distinguishing near-OOD from far-OOD}. This can be seen in Figure \ref{fig:rmd_vs_binary} where near-OOD datasets have scores distributed in between in-domain and far-OOD. In the summarization task, near-OOD (news articles) datasets \texttt{cnn\_dailymail} and \texttt{newsroom} have their RMD scores distributed in the middle of \texttt{xsum} and \texttt{reddit\_tifu}, \texttt{forumsum} and \texttt{samsum}. In contrast, under the binary logits score, the near-OOD and far-OOD datasets have largely overlapping score distributions making it hard to distinguish between the two.
%In other words, binary logits score are very sensitive to detect distributional shift, even if the shift is small and is most useful in  the very conservative application scenario where the model cannot tolerate any shifted data.
In practice, RMD OOD score may be better suited for selective generation where domain shifts are expected. We explore this in more detail in Section~\ref{sec:quality}. 

% \vspace{-0.425em}
% \vspace{-0.6em}

% All methods except for perplexity and NLI score have good OOD detection performance with AUROC $>0.95$ for far-OOD datasets. Among them, relative Mahalanobis distance and binary classifier perform the best and have similar AUROC values. For near shift OOD \texttt{cnn\_dailymail} and \texttt{newsroom}, binary classifier has the highest AUROCs $0.997$ for input and $0.985$ for output.

For the translation task, we additionally note that all methods have small AUROC for \texttt{law} dataset,  suggesting that none of the methods are detecting the dataset as OOD.
To better understand the special characteristics of the \texttt{law} dataset, we conducted an $n$-gram overlap analysis between the various test sets including \texttt{law} and the in-domain training data. We observed that \texttt{law} has the highest unigram overlap rate (48.8\%) and the second highest overall overlap with the in-domain data (Table \ref{tab:translate_ngram}).\footnote{We define overlap rate as the percentage of unique $n$-grams in the test set that are also present in the in-domain data. 
The overall overlap is defined as the geometric mean of all the $n$-gram overlap rates up to $n=4$.
All domains/splits including the in-domain data are subsampled to 1K for this analysis.}
%This confirms that \textbf{\texttt{law} is actually not OOD} data and explains why no method can detect it.
This shows that \texttt{law} is close to in-domain data in terms of surface features, which might contribute to the low AUROC scores for all tested methods.

\revision{We use \texttt{ParaCrawl} instead of \texttt{C4} for translation because our translation model is trained on the sentence level, unlike the summarization model that takes the document as input. 
To further explore the effect of the background data on the performance, we split \texttt{C4} documents into sentences and use that as the background data
to compute the scores.
%and compare that with the version using \texttt{ParaCrawl}. 
The OOD detection performance using \texttt{C4} sentences is very similar to that using \texttt{ParaCrawl}, as shown in Table \ref{tab:ood_c4sent}, suggesting that our method is not particularly sensitive to the choice of background data.}

\section{Using OOD scores for selective generation}
\label{sec:quality}

The most conservative option for deployment of a conditional language model is to completely abstain from generating on inputs that are detected as out-of-distribution, for which we have shown in Section \ref{sec:exp_detection} our OOD scores are fairly accurate. However, it is often desirable to expand the use of models beyond strictly in-distribution examples,  if the quality of outputs is sufficiently high. In classification, this has been framed as determining when to trust a classifier, or \emph{selective prediction} \citep{geifman2017selective,lakshminarayanan2017simple,tran2022plex}. 
In this section, we seek to predict the quality of generation given an example, which may be out-of-distribution and \emph{abstain} if the predicted quality is low. We call this \emph{selective generation}.
In practice, abstaining may correspond to hiding the model's generated text, or turning off a summarization/translation feature.

% Using OOD score for quality prediction; Combining perplexity and OOD score for better quality prediction

\subsection{Experiment setup}
We use the same models and datasets described in Section \ref{sec:experiment_setup} but instead of simply detecting out-of-distribution examples, our focus now is to \emph{predict the quality of generation} for examples possibly outside the training distribution.

\compactparagraph{Measuring Translation quality}\quad We use \bleurt \citep{pu-etal-2021-learning} as the main metric to measure translation quality. 
Previous work has demonstrated that neural metrics such as \bleurt are much better correlated with human evaluation, 
on both the system level and the sentence level~\citep{freitag-etal-2021-results}. 
\bleurt scores range from 0 to 1, with higher scores indicating better quality.

\compactparagraph{Measuring Summarization quality} 
In general, it is unclear how to automatically measure the quality of summaries generated by a model on out-of-distribution examples (in this case, examples from different datasets). The reason is summarization datasets have dataset-specific summary styles that may be difficult to compare.
For example, \xsum summaries are typically single-sentence whereas \cnn summaries consist of multiple sentences. Thus we report \rougeone score as an automatic measure but primarily use human evaluation to assess the quality. Amazon Mechanical Turk workers were asked to evaluate summaries generated by the \xsum model on a scale of 1-5 (bad-good) using 100 examples from \xsum, \cnn, \reddit, and \samsum. We collected 3 ratings per example and \revision{computed the median}. % to reduce inter-rater noise. 
\revision{See Section \ref{sec:mturk} for more details}.

\subsection{Perplexity has diminishing capability in predicting quality on OOD data}

\begin{figure}[htb]
\centering
\begin{subfigure}[t]{0.29\textwidth} % changed for arxiv
\centering
\caption{Summarization, \rougeone}
\vspace{-0.2em}
\includegraphics[width=\textwidth]{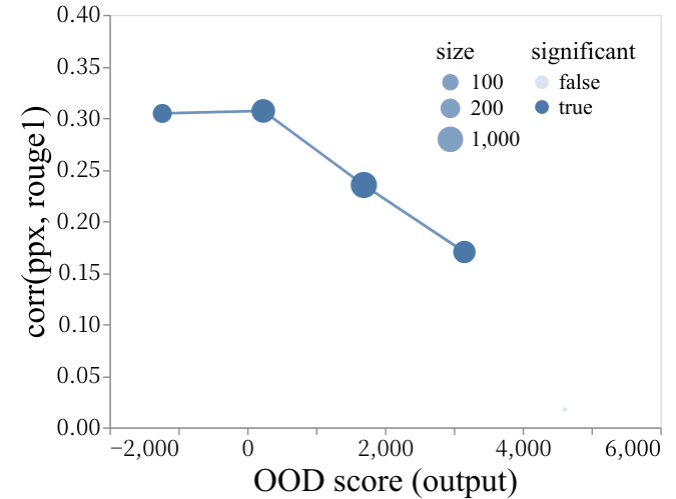} 
\end{subfigure}
\begin{subfigure}[t]{0.30\textwidth} % changed for arxiv
\centering
\caption{Summarization, human rating}
\vspace{-0.1em}
\includegraphics[width=\textwidth]{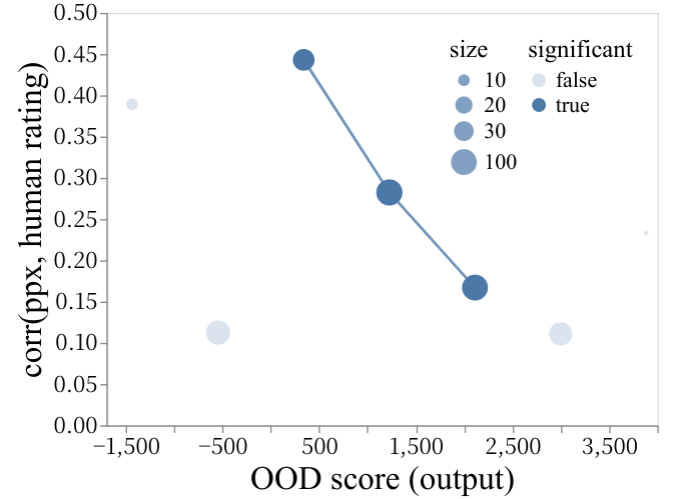} 
\end{subfigure} 
% \vspace{-0.8em}
\begin{subfigure}[t]{0.30\textwidth} % changed for arxiv
\centering
\caption{Translation, \bleurt}
\vspace{-0.1em}
\includegraphics[width=\textwidth]{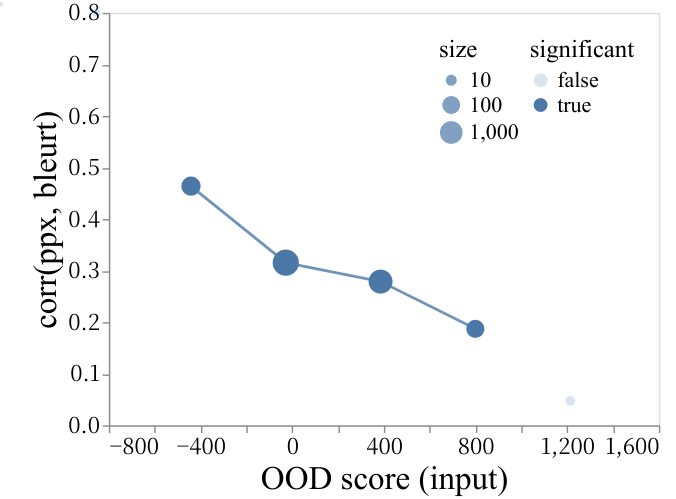}
\end{subfigure} 
% \vspace{-0.1em}
\vspace{-0.5em}
\caption{The Kendall's $\tau$ correlation between perplexity and (a) \rougeone, (b) human evaluation median rating, and (c) \bleurt decreases as OOD score increases respectively. Note that we use output RMD OOD score for summarization and input RMD OOD score for translation.}
\label{fig:corr_ppx_as_ood}
\end{figure}

Since the models are trained using negative log-likelihood as the loss, perplexity (which is monotonically related) is a good predictor of output quality for in-domain data. In fact, the Kendall rank correlation coefficient $\tau$ between perplexity and human judged quality score is 0.256 (See Table \ref{tab:quality_corr_reduced}) for in-domain \texttt{xsum} for summarization. 
However, when including shifted datasets to test, we found that the perplexity score is worse at predicting quality on OOD data. For example the Kendall's $\tau$ decreases to 0.068 for OOD dataset \texttt{samsum} (see Table \ref{tab:corr_heval_sum}). 
We observed similar trend in translation, although less severe, as data shifted from in-domain to OOD, the Kendall's $\tau$ between perplexity and \bleurt decreases
% from 0.309 to 0.227
(see Table \ref{tab:quality_corr_MT}).
Figure \ref{fig:corr_ppx_as_ood} further shows the correlation between perplexity and the quality score (\rougeone, human rating, and \bleurt, respectively) as a function of OOD score. It is clear to see the correlation decreasing as OOD score increases and the trend is consistent for both summarization and translation. 

\vspace{-0.3em}

\vspace{-1em}
\subsection{Combining OOD scores and perplexity}

While model perplexity for quality estimation is worse for OOD examples, we observed that our OOD scores and perplexity are complementary in quality prediction. Figure \ref{fig:2d_ppx_ood} shows a 2-D plot between the OOD score and perplexity regarding quality. 
We can see that neither perplexity nor OOD score can perfectly separate good and bad examples, and the combination of the two can work much better. 
Our observation echos  work in uncertainty estimation in classification models \citep{mukhoti2021deep}: perplexity based on softmax predictive distribution is regarded as an estimation for \emph{aleatoric} uncertainty (caused by inherent noise or ambiguity in data), and the OOD distance based on representation estimates the \emph{epistemic} uncertainty (caused by a lack of training data), and combining the two provides a comprehensive estimation of uncertainty.

We propose two simple methods to combine perplexity and OOD scores. (1) A simple linear regression, trained on a random 10\% data split using \rougeone or \bleurt as the quality score, and evaluated on the test split and human evaluation split.
(2) the sum of the percentile ranks (PR) of the scores, i.e. \prsum$=\text{PR}_\text{perplexity} + \text{PR}_{\text{OOD}}$. We sum PRs instead of their raw values because the two scores are in different ranges, $\text{PR}(x)=\frac{R(x)}{N} \times 100$, where $R(x)$ is $x$'s rank in the list of size $N$.
%$y = a + b_1 \text{Perplexity} + b_2 \text{OOD}_\text{input} + b_3 \text{OOD}_\text{output} + \epsilon$. The regression model is trained on a random 10\% data split using \texttt{rouge1} as the quality score. We then evaluate on the test split and human evaluation split. 
%When evaculuaing on human evaluation split, we use human judged median rating as the final quality metric.
%(2) Inspired by Figure \ref{fig:2d_ppx_ood}, we propose to use the sum of the percentile ranks (PR) of the scores, i.e. $y = \text{PR}_\text{perplexity} + \text{PR}_{\text{OOD}}$, as the predictor for the quality.  \todo{Use consistent notation with table}

Table \ref{tab:quality_corr_reduced} shows the Kendall's $\tau$ correlation coefficient between the various single and combined scores and the quality metric with only in-domain and all examples from all datasets.
When all datasets are merged, \textbf{the combined scores significantly improve the correlation over perplexity by up to 12\% (absolute) for summarization and 8\% for translation, while the gains over the best external model-based (and much more expensive) baselines are 4\% and 3\%}.
The two combination methods perform similarly. See Tables \ref{tab:corr_heval_sum} and \ref{tab:quality_corr_MT} for an expanded table of scores.

% Table copied and reduced from table_quality_correlation_combined.tex
\begin{table}[t]
\vspace{-0.67em}
\caption{Kendall’s $\tau$ correlation ($p$-value $<0.05$ are grayed out) between various measures and human evaluation for summarization and \bleurt for translation. The ``All'' column shows the correlation when both in-domain and OOD examples are merged. 
Note for negatively correlated scores (e.g. perplexity (ppx), RMD), we take the negative value of the score for easier comparison.}
\vspace{-0.8em}
\begin{minipage}{0.48\linewidth}
\caption*{(a) Summarization }
\vspace{-0.5em}
\centering
\scriptsize
\begin{tabular}{lcc}
\textbf{Measure}         &  \textbf{In-domain} &  \textbf{All} \\ \toprule
\multicolumn{3}{c}{Single Score} \\ \midrule    
Perplexity (baseline)                                 & 0.256   & 0.300 \\
NLI score (baseline)                                   & 0.337   & 0.381 \\
Input RMD                                & \textcolor{gray}{0.015}  & 0.336 \\
Output RMD                                 & \textcolor{gray}{0.053}  & 0.385 \\
\midrule
\multicolumn{3}{c}{Combined Score}  \\ \midrule
\prsum (ppx, input RMD)                    & 0.186   & 0.358 \\
\prsum (ppx, output RMD)                   & \textcolor{gray}{0.250}  &   \underline{0.415} \\
Linear Reg. (ppx, input \& output)                                       & 0.235   & \textbf{0.422} \\
\bottomrule
\end{tabular}
% \vspace{1em}
\end{minipage}
\vspace{-0.5em}
\begin{minipage}{0.48\linewidth}
\caption*{(b) Translation }
\vspace{-1em}
\scriptsize
\centering
\begin{tabular}{lcc}
\textbf{Measure}         &  \textbf{In-domain} &  \textbf{All} \\
\toprule
\multicolumn{3}{c}{Single Score} \\
\midrule
Perplexity (baseline) & 0.309 &  0.286 \\
COMET (baseline) & 0.184 &  0.336 \\
Prism (baseline) & 0.184 & 0.301 \\
Input RMD & 0.147 & 0.195 \\
Output RMD & 0.086 & 0.170 \\
\midrule
\multicolumn{3}{c}{Combined Score} \\ 
\midrule
\prsum (ppx, input RMD) & 0.321  & \textbf{0.361} \\
\prsum (ppx, output RMD) & 0.323  & \underline{0.356} \\
Linear Reg. (ppx, input \& output) & 0.318  & 0.352 \\
\bottomrule
\end{tabular}
\end{minipage}
\label{tab:quality_corr_reduced}
\vspace{-0.6em}

 \end{table}

\subsection{Selective generation using the combined score}
In selective generation, our goal is to generate when the model is more likely to produce high-quality output, and \emph{abstain} otherwise, enabling safer deployment of generative language models.
To evaluate that, we propose using the \textbf{\emph{Quality vs Abstention Curve} (QA)}, analogous to accuracy versus rejection curve used for selective prediction in the classification \citep{chow1957optimum,bartlett2008classification,geifman2017selective}.  Similar concepts were proposed also in \cite{malinin2020uncertainty,xiao2020wat}, \revision{but they only use automatic quality metrics for the analysis while we consider human evaluation to assess the quality as well}.
Specifically, at a given abstention rate $\alpha$,  the highest $\alpha$-fraction scoring examples are removed and the average quality of remaining examples is computed. We want to maximize the quality of what is selectively generated and a better curve is one that tends to the upper-left which corresponds to removing bad examples earlier than good ones.

\begin{figure}[!htb]
\centering
% \begin{subfigure}[t]{0.30\textwidth} % changed for arxiv
%     \centering
%     \caption{Perplexity vs RMD output OOD}
%     % \vspace{-0.1em}
%     \includegraphics[width=1.1\textwidth]{figures/2d_ppx_heval_vs_output_ood.png}
% \end{subfigure} 
% \vspace{-1.2em}
\captionsetup[subfigure]{oneside,margin={0.7cm,0cm}}

\begin{subfigure}[t]{0.36\textwidth} % changed for arxiv
  
\centering
\caption{}
\vspace{-0.8em}
\includegraphics[width=0.91\textwidth]{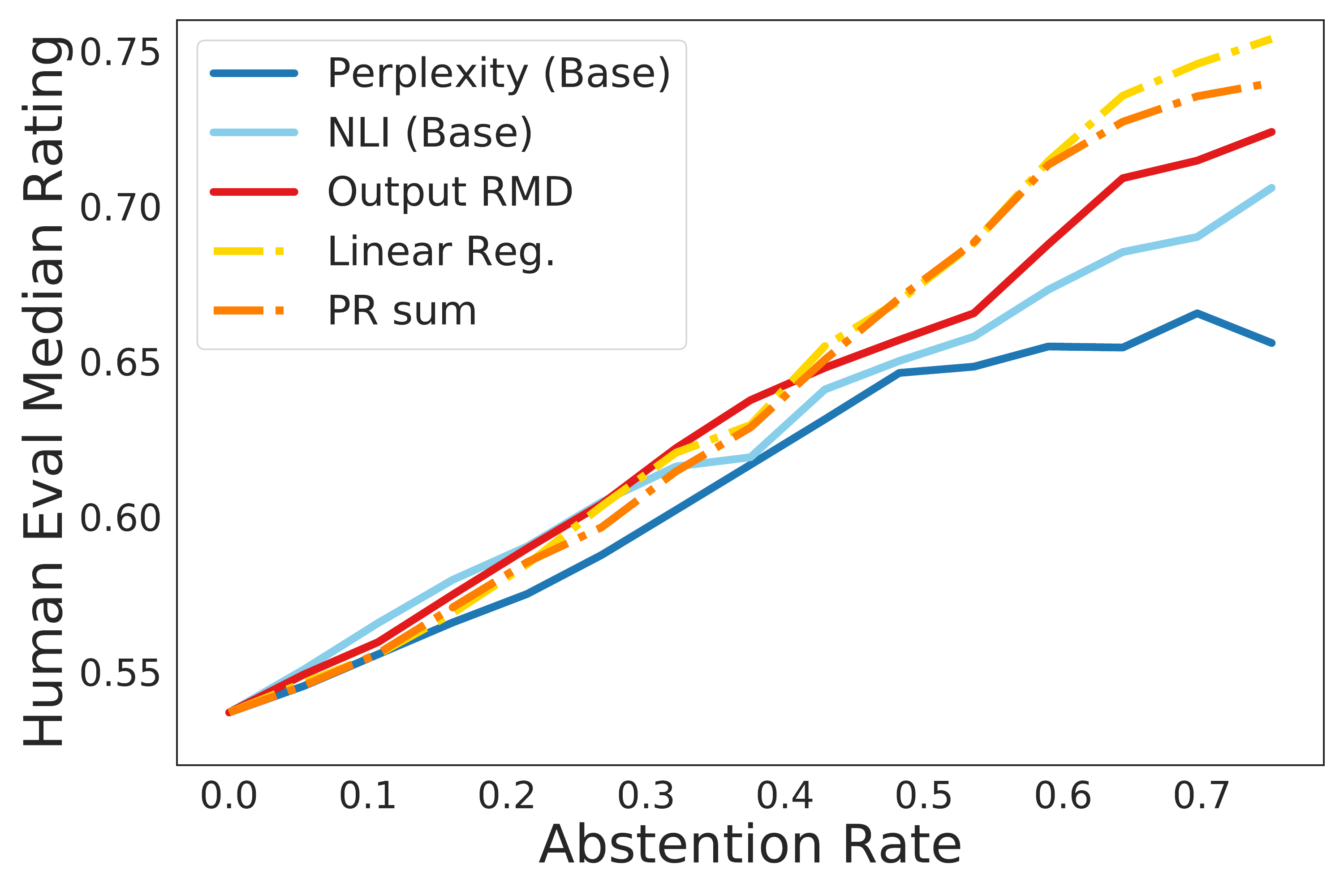} % s_abstain_vs_quality_methods8.pdf}
\end{subfigure} 
% \vspace{0.8em}
\begin{subfigure}[t]{0.36\textwidth} % changed for arxiv
\centering
\caption{}
\vspace{-0.8em}
\includegraphics[width=0.91\textwidth]{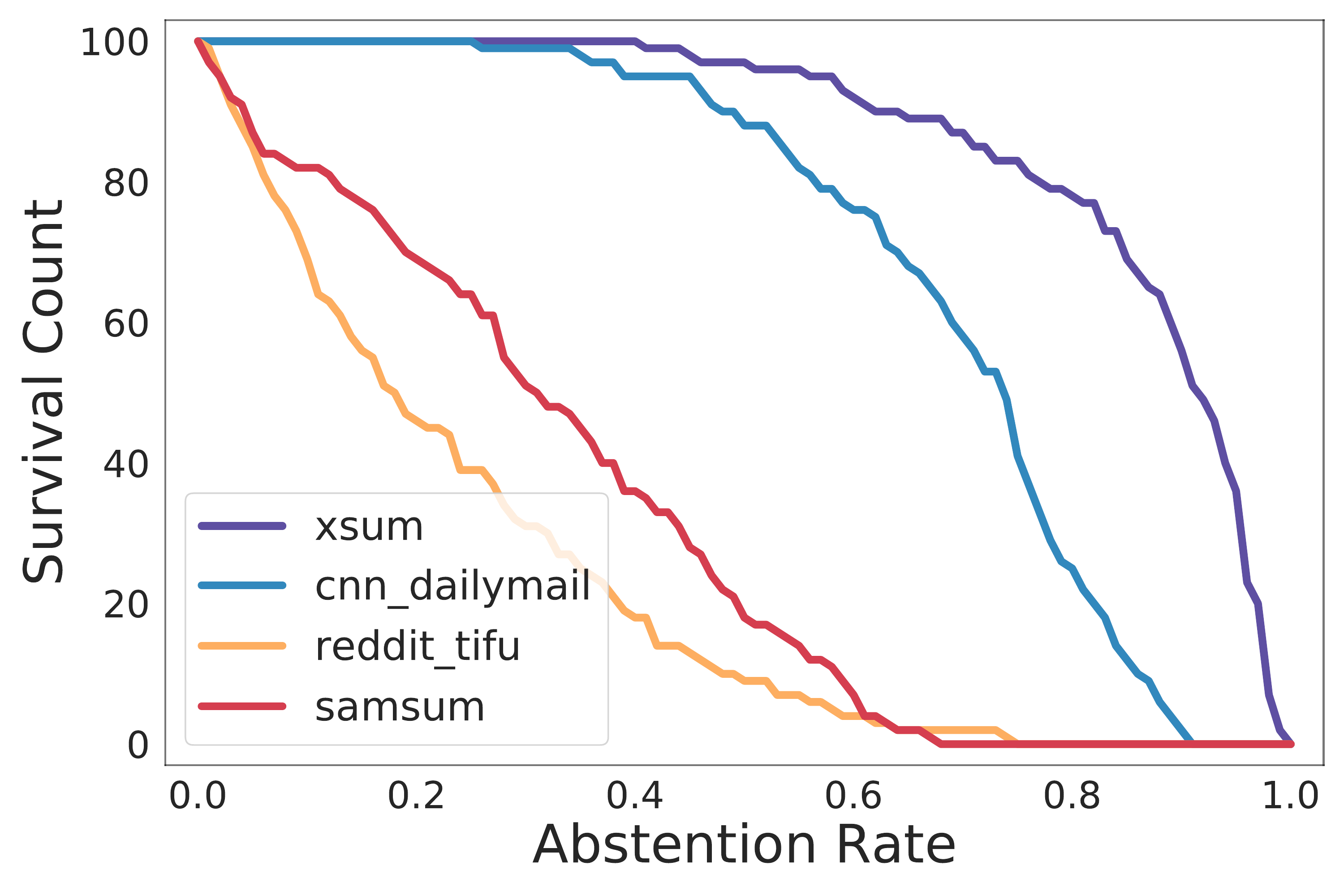}
\end{subfigure} 
% \begin{subfigure}[t]{0.30\textwidth} % changed for arxiv
%     \centering
%     \caption{\toaddjl{caption}}
%     % \vspace{-0.1em}
%     \includegraphics[width=0.9\textwidth]{figures/translation/cm_MT.png}
% \end{subfigure} 
% \vspace{-0.1em}
% \vspace{-0.8em}
\begin{subfigure}[t]{0.36\textwidth} % changed for arxiv
\centering
\caption{}
\vspace{-0.55em}
% \vspace{-0.1em}
\includegraphics[width=0.91\textwidth]{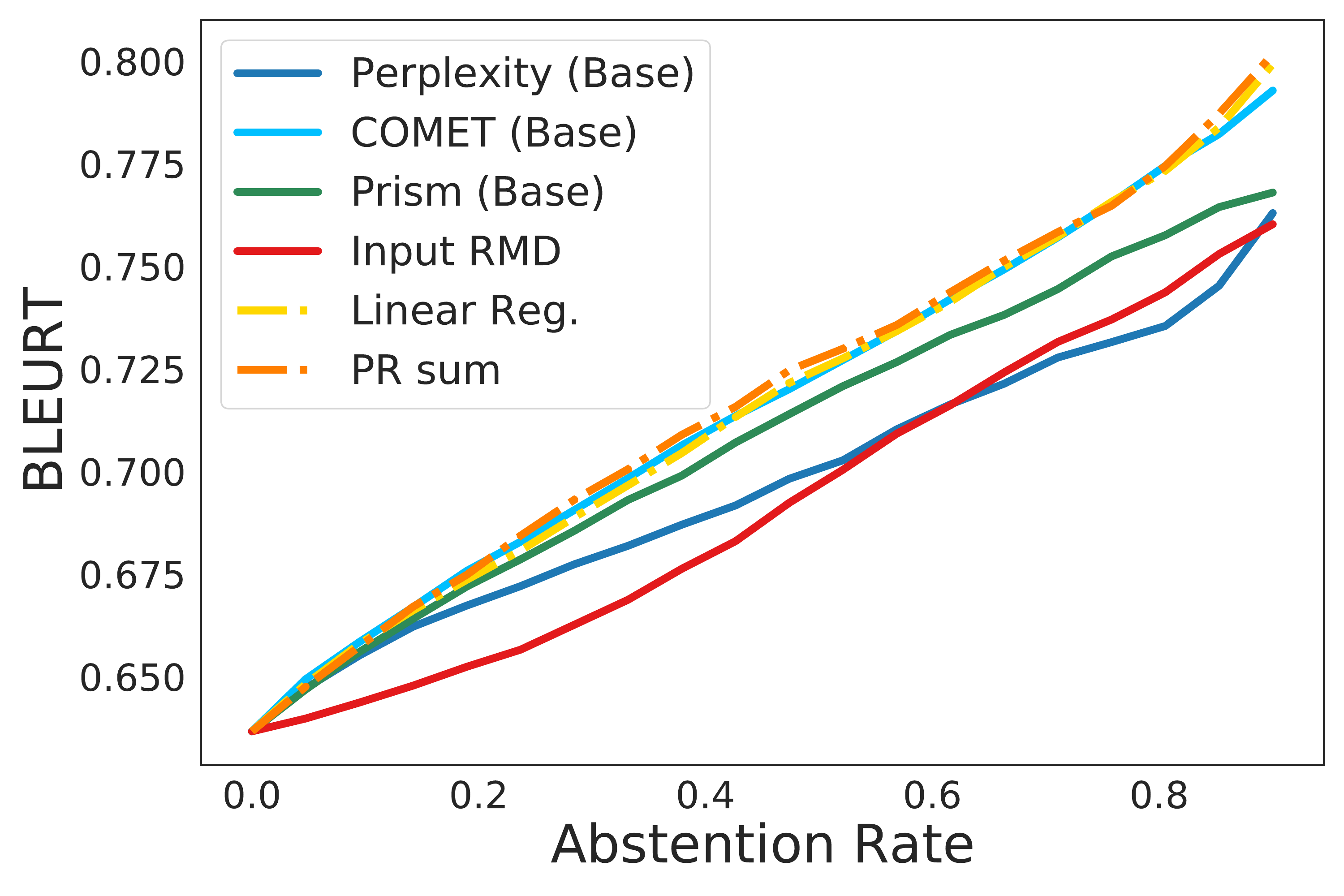}
\end{subfigure} 
\begin{subfigure}[t]{0.36\textwidth} % changed for arxiv
\centering
\caption{}
\vspace{-0.8em}
\includegraphics[width=0.91\textwidth]{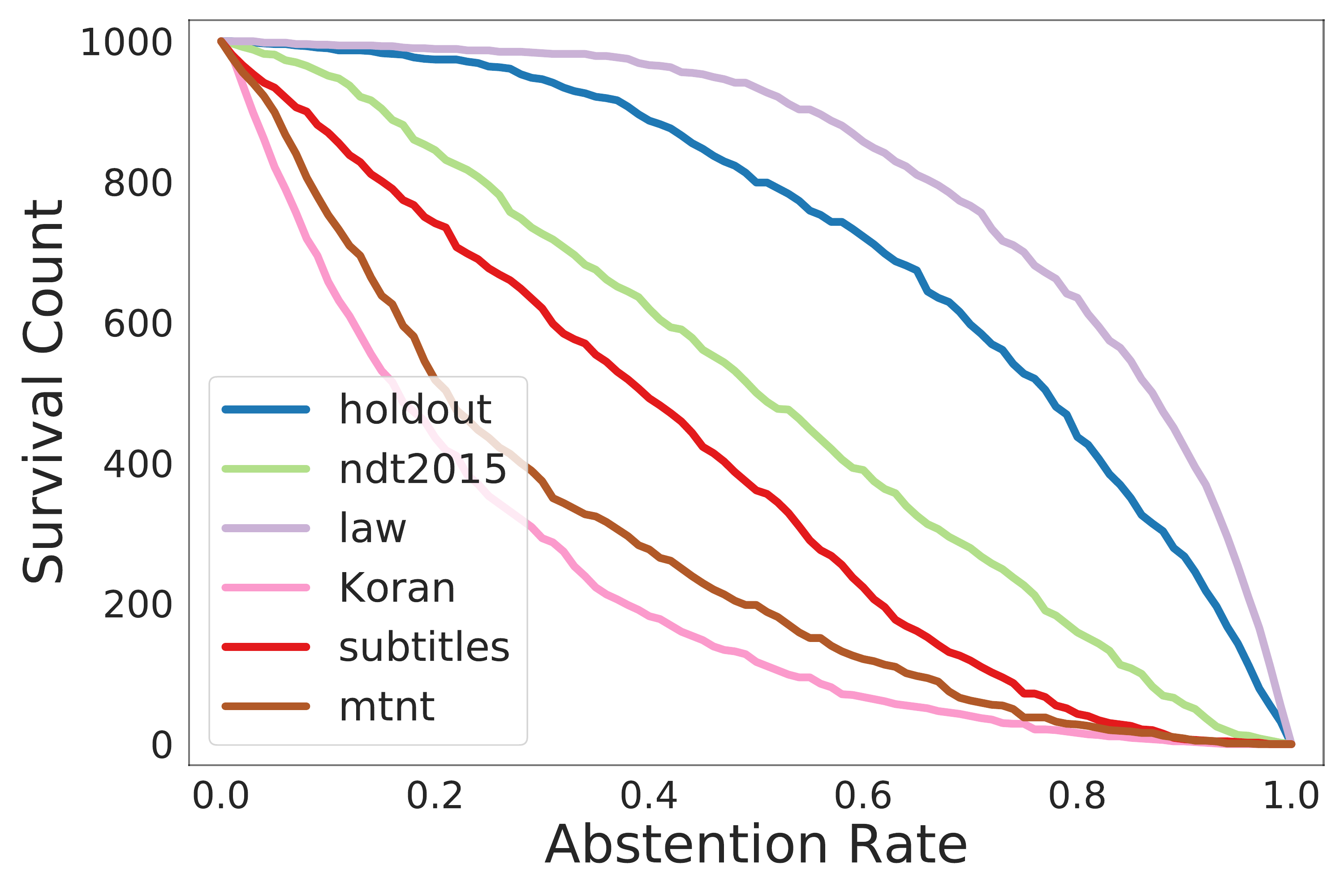}
\end{subfigure} 
% \vspace{-1em}
\vspace{-0.5em}
\caption{(a) The Quality (\texttt{human eval}) vs Abstention curve for summarization. Combined scores have the highest quality at almost all abstention rates. (b) Survival count of each dataset as a function of abstention rate, using \prsum{} (we use output/input RMD for summarization/translation to pair with perplexity). OOD data is abstained earlier than in-domain. (c, d) The same as (a, b) for translation. }
\vspace{-0.1em}
\label{fig:abs_vs_quality}
\end{figure}

Figure \ref{fig:abs_vs_quality} shows the QA curves for various methods on summarization and translation. 
%Various scores are evaluated, including input and output RMD OOD, output perplexity, NLI (for summarization only), and COMET and Prism (for translation only), and combined scores like linear regression model, and the sum of the percentile ranks of RMD OOD and perplexity scores. 
Quality is measured by human evaluation for summarization (see Figure \ref{fig:abs_vs_quality_rouge1} for similar \rougeone plot), and \bleurt for  translation.
\textbf{The combined scores have the highest quality score at almost all abstention rates for both summarization and translation}, while linear regression and \prsum{} perform similarly.
For single scores, the OOD score performs better than perplexity and NLI scores at almost all abstention rates for summarization.
For translation, the OOD score is better than perplexity when abstention rate \revision{$\alpha>0.65$} and worse than perplexity when \revision{$\alpha<0.65$}. In other words, OOD score is better at abstaining slightly far-OOD while perplexity is better at abstaining near-OOD examples. 
Interestingly, our combined score is even marginally better than COMET that requires a separate neural network trained on human evaluation data. 
Prism is better than single scores, but much worse than our combined score. 
Area under the QA curves are shown in Tables \ref{table:abstention_AUC_sum_heval} and \ref{table:abstention_AUC_translate} for reference.

Figures \ref{fig:abs_vs_quality} (b, d) are the corresponding survival curves showing how many examples per dataset are selected for generation as a function of abstention rate, based on the \prsum{} score. For summarization, the samples from far-OOD datasets \texttt{reddit\_tifu} and \texttt{samsum} are eliminated first with their sample count decreasing rapidly. The near-OOD dataset \texttt{cnn\_dailymail} and in-domain \texttt{xsum} are kept intact until $\alpha>0.3$, and in-domain \texttt{xsum} examples survive the longest. 
Similarly for translation, the out-of-domain and worst-quality (as seen in Table \ref{tab:quality_by_dataset_MT}) \texttt{Koran}, \texttt{MTNT}, and \texttt{subtitles} examples are eliminated first, and the best-performing \texttt{law} and in-domain datasets are abstained last. \textbf{The order in which datasets are eliminated corresponds to the aggregate quality by dataset}, which we report in Table \ref{tab:quality_by_dataset}.
\revision{Besides the quantitative results, we show a few real examples in Section \ref{sec:cnn_example} to better demonstrate how our predicted quality score helps selective generation. }

%%%%%%%%%%%%%%%%%%%%%% MT Section %%%%%%%%%%%%%%%%%%%%%%
% \include{tex/figure_translation_quality} # have merged to summarization
% \include{tex/table_abstention_AUC_translate}

% \vspace{1em}
\section{Related Work}
%\citep{hendrycks2016baseline,liang2017enhancing,lakshminarayanan2017simple,lee2018simple,hendrycks2018deep,ren2019likelihood,hendrycks2019scaling} and later text \citep{hendrycks2020pretrained,arora2021types,liu2020simple,tran2022plex,rawat2021pnpood},b
% vision classification problems \citep{hendrycks2016baseline,liang2017enhancing,lakshminarayanan2017simple,lee2018simple,hendrycks2018deep,ren2019likelihood,hendrycks2019scaling}, and later in text classification problems such as sentiment prediction \citep{hendrycks2020pretrained}, entailment prediction \citep{hendrycks2020pretrained,arora2021types,tran2022plex}, intent prediction \citep{liu2020simple,tran2022plex}, and topic prediction \citep{rawat2021pnpood}. 
% \todo{Add back stuff removed from intro?}
% \compactparagraph{OOD detection for classification problem} 
OOD detection problem was first proposed and studied in vision classification problems \citep{hendrycks2016baseline,liang2017enhancing,lakshminarayanan2017simple,lee2018simple,hendrycks2018deep,hendrycks2019scaling}, and later in text classification problems such as sentiment analysis \citep{hendrycks2020pretrained}, natural language inference \citep{arora2021types}, intent prediction \citep{liu2020simple,tran2022plex}, and topic prediction \citep{rawat2021pnpood}. 
The widely used OOD methods can be characterized roughly into two categories (1) softmax probability or logits-based scores \citep{hendrycks2016baseline,liang2017enhancing,hendrycks2019scaling,liu2020energy}, (2) embedding-based methods that measure the distance to the training distribution in the embedding space \citep{lee2018simple,ren2021simple,sun2022out}, 
% A few studies develop generative model-based OOD detection methods that evaluate the likelihood of the input under training distribution \citep{ren2019likelihood,nalisnick2018deep,morningstar2021density,choi2018waic}, but this approach is found to have some failure modes, and it requires training of an additional generative model besides the classification model, which is often time consuming. 
\revision{(3) contrastive learning based methods which incorporate the contrastive loss into the classification cross-entropy loss to improve representation learning and consequently improve OOD detection  \citep{winkens2020contrastive,zhou2021contrastive}.} 
\revision{Though it is not straightforward to extend those classifier-based scores to CLMs especially for input OOD detection, we extend three of them based on our understanding
% (1) the averaged maximum softmax over output tokens, (2) the averaged energy score over the output tokens, (3) KNN-score based on input and output embeddings, 
as baselines for comparison with our methods. 
% We also include one ensemble-based method, (4) the ensemble of the output perplexity from multiple Monte-Carlo dropout samples, as another baseline. 
See Section \ref{sec:add_baselines} for details. 
The results in Table \ref{tab:ood_more_baseline} show that those methods are in general not competitive with our proposed methods RMD and Binary logits, especially %and the performance gap between those methods and our methods is quite big especially 
on near-OOD datasets. }

% Note that a different version of relative Mahalanobis distance was previously proposed specifically for classifier-based model for OOD detection in \cite{ren2021simple}, but the relative distance proposed there was the difference between the class-conditional Gaussian with a single background Gaussian fitted using the same dataset but ignoring class labels. That is different from the proposed method in this work, since we fit foreground and background Gaussian using different datasets, and the foreground Gaussian is not class-conditional but a single Gaussian as the conditional language model does not have class labels. 

% \compactparagraph{OOD detection for sequence model} 
OOD detection problem is less studied in CLMs. 
A few studies explored OOD detection in semantic parsing \citep{lukovnikov2021detecting,lin2022towards}, speech recognition \citep{malinin2020uncertainty}, and machine translation \citep{malinin2021shifts,xiao2020wat}, but many of them focus on ensemble-based methods like Monte Carlo dropout or deep ensemble which use the averaged perplexity after sampling multiple output sequences.%as the uncertainty score. 
%\cite{malinin2020uncertainty} showed the ensemble based methods improve the error detection and OOD detection in machine translation and speech recognition model, and \cite{xiao2020wat} similarly showed that the proposed output variance score ensembled from Bayeisan neural networks improves output quality prediction, but 
The ensembling method costs $N$ times of the inference time, which is not feasible in practice. 
In this work, we focus on developing scores that can be readily derived from the generative model itself, without much increase in computation.
\revision{We include an ensemble-based baseline in Section \ref{sec:add_baselines} and show that its performance is worse than our methods.}

\section{Conclusion and Future work}
We have proposed lightweight and accurate scores to detect out-of-distribution examples for conditional language generation tasks. For real-world deployment, we have also shown how our OOD scores can be combined with language model perplexity to selectively generate high-quality outputs while abstaining from low-quality ones  in the setting of input distribution shift.

Although our experiments focus on summarization and translation, our methods do not make any assumptions about the task modality, and we believe our method is widely applicable to other tasks where the model output is a sequence, e.g. image captioning.
While our analysis was restricted to conditional language modeling with encoder-decoder Transformers, we expect our method to also work with decoder-only \citep{generating_wikipedia} architectures, used by some large language models such as GPT-3 \citep{gpt3}, PaLM \citep{palm}, and LaMDA \citep{thoppilan2022lamda}.
% \todo{summarize results, and talk about possible future work with decoder-only, and other modalities, e.g. image-captioning}

Finally, analyzing why certain examples are OOD could lead to insights in how to make models more robust. Section \ref{sec:visualization} presents one possible way to attribute OOD scores to sentences. % along with a visualization.

%\input{TODO.tex}
% Removed during Review Period

\clearpage

\section*{Acknowledgements}

\revision{The authors would like to thank Jeremiah Zhe Liu, Sharat Chikkerur, and the anonymous reviewers for their helpful feedback on the manuscript. The authors would also like to thank Colin Cherry, George Foster, and Polina Zablotskaia for their feedback throughout the project. }

% \clearpage
\bibliography{main}%, anthology}
\bibliographystyle{iclr2023_conference}

\clearpage

\appendix

% Set special A.X figure numbering for appendix.
\renewcommand\thefigure{\thesection.\arabic{figure}}    
\setcounter{figure}{0}    
\renewcommand\thetable{\thesection.\arabic{table}}    
\setcounter{table}{0}    

\clearpage

% \counterwithin{figure}{section}
% \counterwithin{table}{section}
% \renewcommand\thefigure{\thesection.\arabic{figure}}
% \renewcommand\thetable{\thesection.\arabic{table}}

\section{Appendix}

\subsection{The output quality for summarization and translation datasets. }

\begin{table}[h]
\caption{The output quality for summarization and translation datasets. (a) Summarization quality (higher is better) for \texttt{xsum} model. \rougeone is based on all samples in the test split per dataset, while human evaluation is based on 100 samples.
The raw human evaluation rating ranges from 1 to 5. We normalized the score by dividing 5.0, and toke the median of the ratings over 3 raters to reduce inter-rater noise. \revision{The standard deviation among 3 ratings are reported in brackets.}  (b) Translation quality for different datasets (higher is better). All datasets are sub-sampled to 1000 sentence pairs.}
\begin{subtable}[c]{1\textwidth}
\centering
\subcaption{Summarization}
\scriptsize
\begin{tabular}{lrr}
\toprule
Dataset          & \rougeone & Human evaluation \\ 
\midrule
\texttt{xsum}     &    0.474        &     0.698 (0.182)                \\
\texttt{cnn\_dailymail}     &      0.226   &    0.624 (0.145)                 \\
\texttt{reddit\_tifu}     &      0.140 & 0.450 (0.152)                     \\
\texttt{samsum}     &      0.210   &           0.376 (0.147)          \\
\bottomrule
\end{tabular}
\label{tab:quality_by_dataset_sum}
\vspace{2em}
\end{subtable}
\begin{subtable}[c]{1\textwidth}
\centering
\subcaption{Translation}
\scriptsize
\begin{tabular}{lrr}
\toprule
Dataset          & \texttt{BLEURT} & \texttt{BLEU} \\ 
\midrule
\texttt{law}     & 0.781                      & 53.8                     \\
\texttt{nt2014}  & 0.731                      & 39.8                     \\ 
\texttt{holdout} & 0.674                      & 41.8                     \\
\texttt{ndt2015} & 0.671                      & 37.9                     \\
\texttt{ndd2015} & 0.664                      & 30.9                     \\
\texttt{medical} & 0.643                      & 34.2                     \\
\texttt{IT}      & 0.588                      & 28.3                     \\
\texttt{MTNT}      & 0.565                      & 32.0                     \\
\texttt{sub}     & 0.552                      & 22.8                     \\
\texttt{Koran}   & 0.491                      & 12.9                     \\
\bottomrule
\end{tabular}
\label{tab:quality_by_dataset_MT}
\end{subtable}
\label{tab:quality_by_dataset}
\end{table}

\subsection{OOD score and perplexity are complementary for predicting output quality.}
\begin{figure}[!htb]
\centering
\begin{subfigure}[t]{0.32\textwidth} % changed for arxiv
\centering
\caption{Summarization, \rougeone}
% \vspace{-0.1em}
\includegraphics[width=0.92\textwidth]{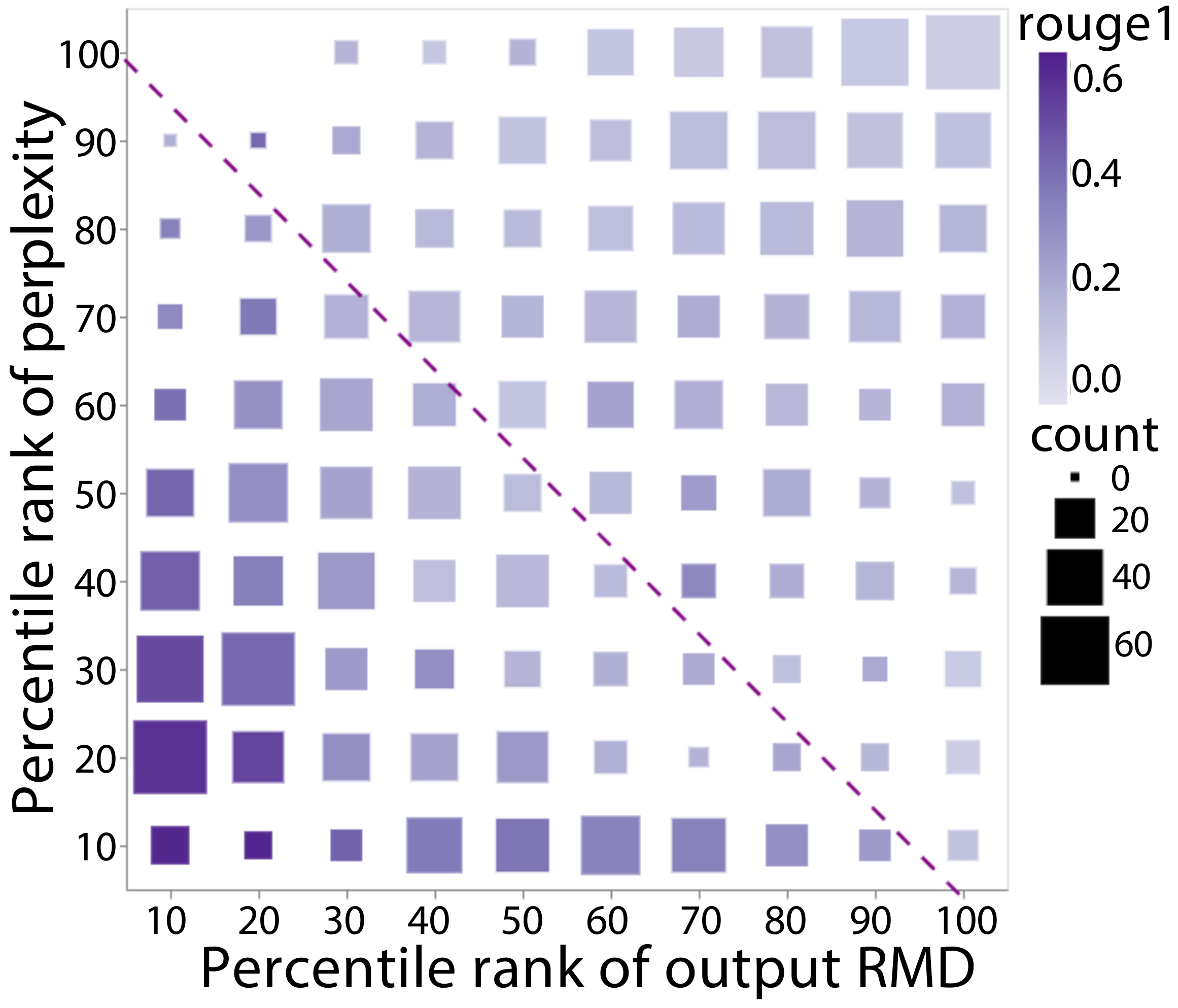} 
\end{subfigure}
\begin{subfigure}[t]{0.32\textwidth} % changed for arxiv
\centering
\caption{Summarization, human rating}
% \vspace{-0.1em}
\includegraphics[width=1.02\textwidth]{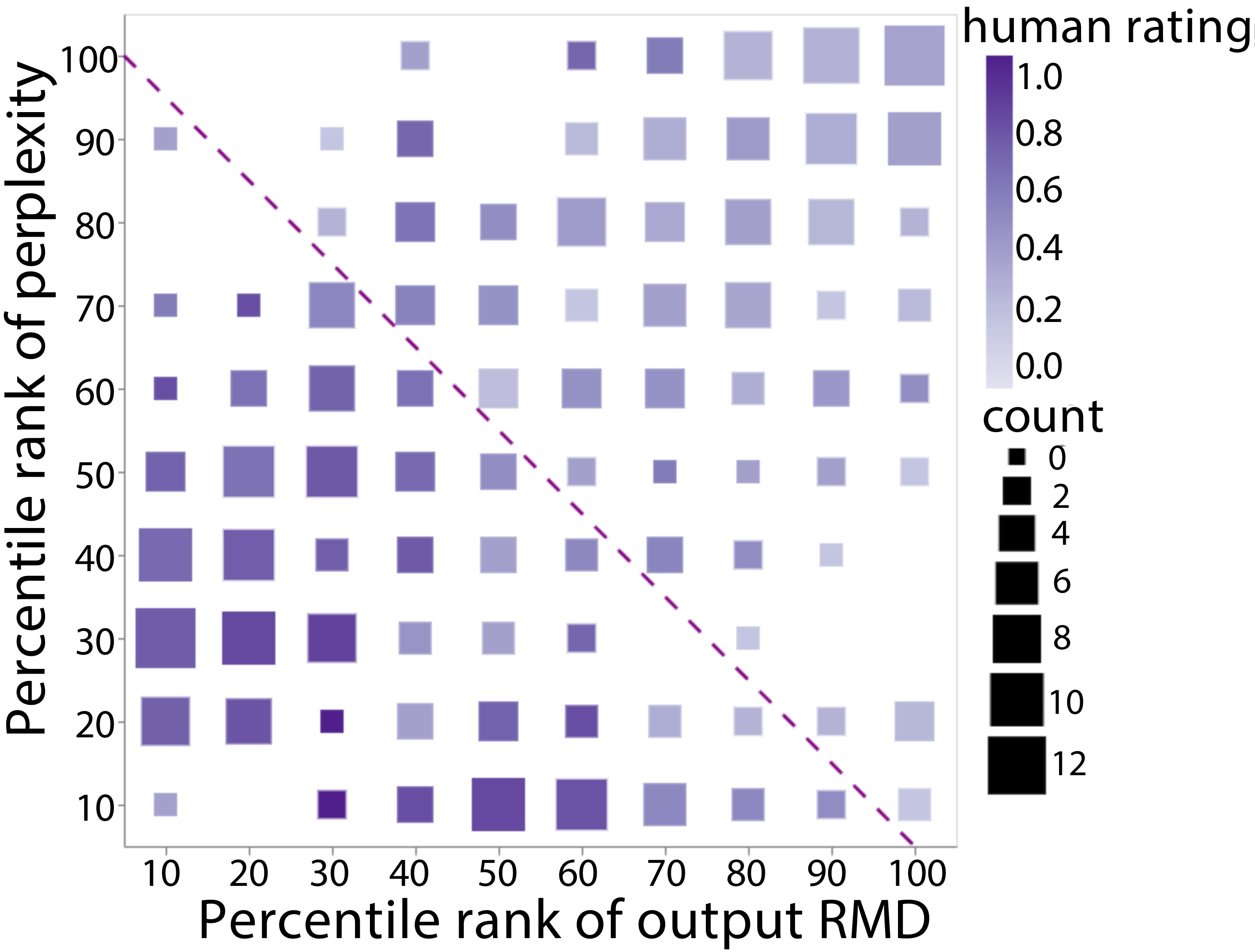} 
\end{subfigure}
\begin{subfigure}[t]{0.32\textwidth} % changed for arxiv
\centering
\caption{Translation, \bleurt}
% \vspace{-0.1em}
\includegraphics[width=0.92\textwidth]{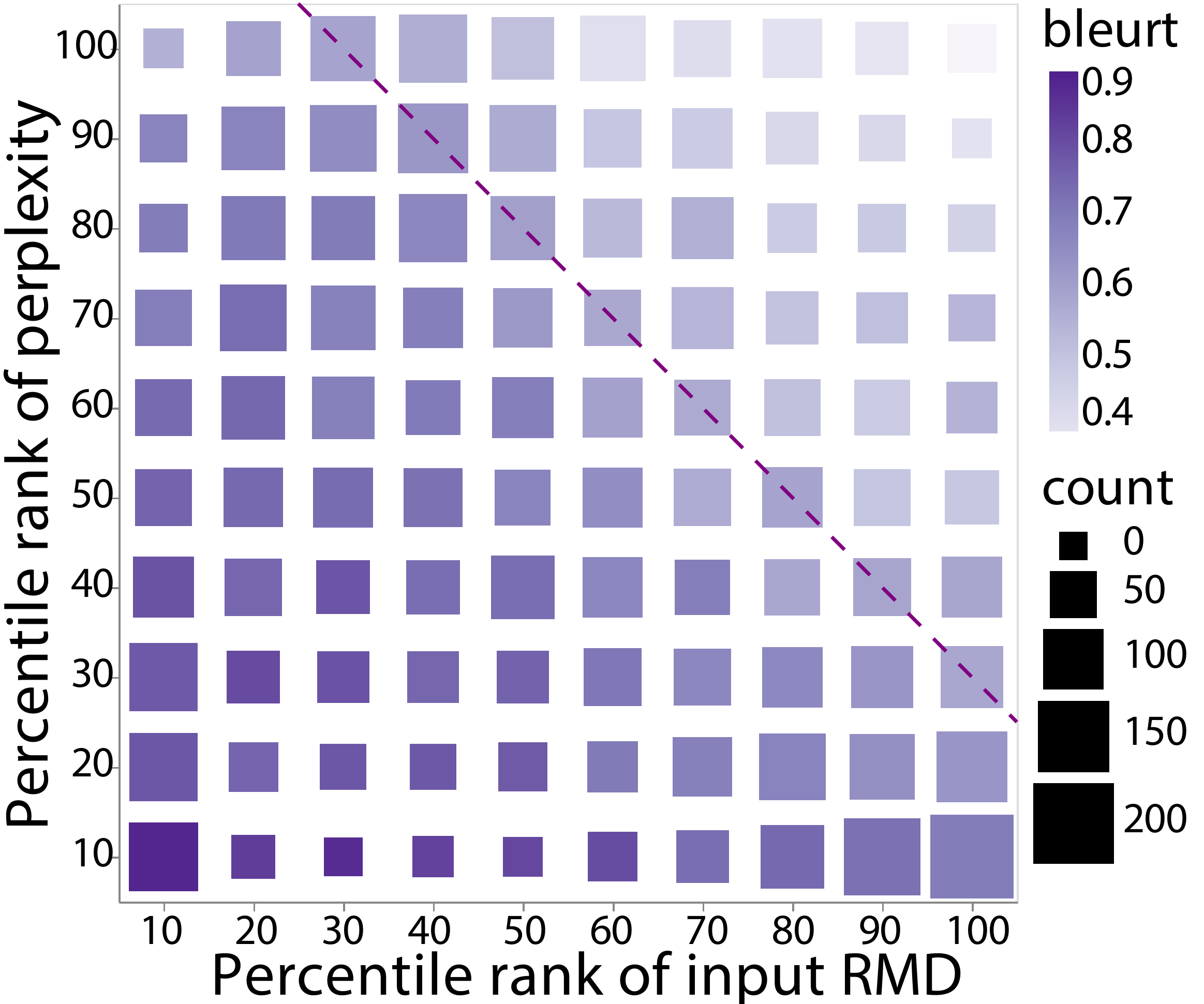} 
\end{subfigure} 
% \vspace{-0.1em}
% \vspace{-0.8em}
\caption{2D plot between OOD and perplexity. The two scores are self-normalized by its percentile rank respectively. Each square corresponds to a subset of samples whose OOD and perplexity scores are within the percentile bin. The size of the square represents the size of the bin where the color indicates the quality of the model's output. The OOD score and perplexity capture different properties of model outputs, and combining both scores can be beneficial for quality prediction. 
 %Since the two scores are in different scale, we normalize the score by itself percentile rank ($\text{PR}$), $\text{PR}(x)=\frac{R(x)}{N} \times 100$, where $R$ is $x$'s rank in the list, and $N$ is the sample size.
 %Samples are binned into 0-10, 10-20, \dots, 90-100 percentile bins. 
 %A square in the 2D plot $[x_1, x_2] \times [y_1, y_2]$ corresponds to a subset of samples whose OOD score is within the percentile bin $[x_1, x_2]$ and perplexity score is within $[y_1, y_2]$. The darkness of the square indicates the quality of the model's output for that subset of samples.
}
\label{fig:2d_ppx_ood}
\end{figure}

\clearpage
\subsection{Amazon Mechanical Turk assessment of summary quality}
\label{sec:mturk}
\begin{figure}[h]
\centering
\includegraphics[width=0.95\textwidth]{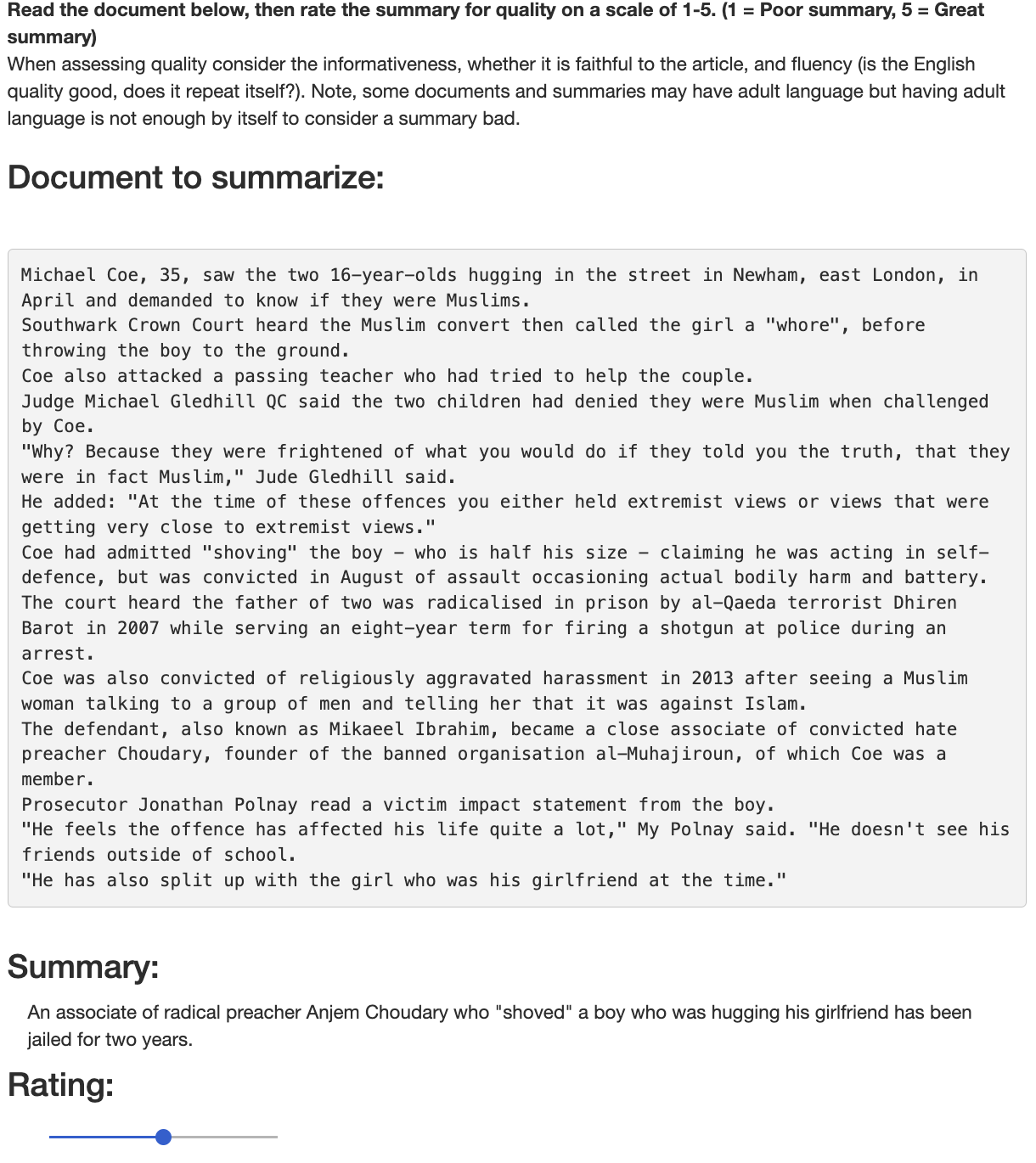}
\caption{AMT template for summarization human evaluation.}
\label{fig:mturk}
\end{figure}

A \pegasuslarge model fine-tuned on \xsum was run on 
a random sample of 100 examples from the test split of four datasets: \xsum, \cnn, \reddit, \samsum. Each example was rated for general summarization quality on a rating of 1-5 by 3 AMT workers using the template shown in Figure \ref{fig:mturk}.
Workers were required to be Masters located in the US with greater than 95\% HIT Approval Rate, with at least 1000 HITs approved and were paid \$0.80 per rating. %The median rating was used to reduce noise.

\clearpage

\subsection{Algorithm for RMD OOD scores}\label{sec:algorithm}

\begin{center}
% \vspace{-1em}
% \scalebox{.9}{%
%\begin{minipage}{0.58\textwidth}
% \begin{minipage}{0.626\textwidth}
\begin{algorithm}[H]
   \caption{Fitting Gaussians for input and output embeddings}
   \label{alg:rmd_fit}
\begin{algorithmic}[1]
   \STATE {\bfseries Input:}
   CLM $M$ with encoder $f_e$ and decoder $f_d$ trained on in-domain train set $\mathcal{D}_{\mathrm{train}}^{\mathrm{in}} = \{ (\vx,\vy) \}$. 
   A large and background dataset such as \texttt{C4} or \texttt{ParaCrawl} $\mathcal{D}_{\mathrm{train}}^{\mathrm{bg}} = \{ (\vx,\hat{\vy}) \}$, where $\hat{\vy} = M(\vx)$. 
   \STATE Generate the input embeddings
   %$\vz$ of $\vx$ from the encoder $f_e$,
   $\mathcal{S}_{\mathrm{train}}^{\mathrm{in}}=\{\vz | f_e(\vx), \vx \in \mathcal{D}_{\mathrm{train}}^{\mathrm{in}} \}$ and $\mathcal{S}_{\mathrm{train}}^{\mathrm{bg}}=\{\vz | f_e(\vx), \vx \in \mathcal{D}_{\mathrm{train}}^{\mathrm{bg}} \}$.
   \STATE Fit a Gaussian distribution $\gaussx$ using $\mathcal{S}_{\mathrm{train}}^{\mathrm{in}}$, and a background Gaussian $\gaussxo$ using $\mathcal{S}_{\mathrm{train}}^{\mathrm{bg}}$.
   \STATE Similarly, generate output embeddings 
%   $\vw$ of $\vy$ from the decoder $f_d$, 
   $\mathcal{E}_{\mathrm{train}}^{\mathrm{in}}=\{\vw | f_d(\vy), \vy \in \mathcal{D}_{\mathrm{train}}^{\mathrm{in}} \}$, and $\mathcal{E}_{\mathrm{train}}^{\mathrm{bg}}=\{\vw | f_d(\hat{\vy}), \hat{\vy} \in \mathcal{D}_{\mathrm{train}}^{\mathrm{bg}} \}$.
   \STATE Fit a Gaussian distribution $\gaussy$ using $\mathcal{E}_{\mathrm{train}}^{\mathrm{in}}$ and a background Gaussian $\gaussyo$ using $\mathcal{E}_{\mathrm{train}}^{\mathrm{bg}}$.
\end{algorithmic}
% \label{alg:rmd}
\end{algorithm}
% \end{minipage}

% \hspace{1em}

% \begin{minipage}{0.50\textwidth}
\begin{algorithm}[H]
   \caption{OOD score inference}
   \label{alg:rmd_infer}
\begin{algorithmic}[1]
   \STATE {\bfseries Input:}
   In-domain test set $\mathcal{D}_{\mathrm{test}}^{\mathrm{in}} = \{ (\vx, \hat{\vy}) \}$. OOD test set $\mathcal{D}_{\mathrm{test}}^{\mathrm{ood}} = \{ (\vx,\hat{\vy}) \}$, where $\hat{\vy}=M(\vx)$.
   %, a pre-trained $f(\cdot): \vx\to\vz$ from inputs to embedding vectors, a trained classification head $h(\cdot): \vz \to \mathbf{p} \in \mathbb{R}^{K+O}$.   
  % \vspace{0.2em}
   \STATE Generate input embeddings $\mathcal{S}_{\mathrm{test}}^{\mathrm{in}}=\{\vz | f_e(\vx), \vx \in \mathcal{D}_{\mathrm{test}}^{\mathrm{in}} \}$ and $\mathcal{S}_{\mathrm{test}}^{\mathrm{ood}}=\{\vz | f_e(\vx), \vx \in \mathcal{D}_{\mathrm{test}}^{\mathrm{ood}} \}$.
   \STATE Compute input OOD score $\text{RMD}_{\text{input}}(\vz)$ for $\vz\in$ $\mathcal{S}_{\mathrm{test}}^{\mathrm{in}}$ and $\mathcal{S}_{\mathrm{test}}^{\mathrm{ood}}$, respectively. Compute AUROC based on the input OOD scores.
   \STATE Similarly, generate output embeddings $\mathcal{E}_{\mathrm{test}}^{\mathrm{in}}=\{\vw | f_d(\hat{\vy}), \hat{\vy} \in \mathcal{D}_{\mathrm{test}}^{\mathrm{in}} \}$ and $\mathcal{E}_{\mathrm{test}}^{\mathrm{ood}}=\{\vw | f_d(\hat{\vy}), \hat{\vy} \in \mathcal{D}_{\mathrm{test}}^{\mathrm{ood}} \}$.
   Compute output OOD score $\text{RMD}_{\text{output}}(\vw)$ for $\vw\in$ $\mathcal{E}_{\mathrm{test}}^{\mathrm{in}}$ and $\mathcal{E}_{\mathrm{test}}^{\mathrm{ood}}$, respectively. Compute AUROC based on the output OOD scores.
\end{algorithmic}
\end{algorithm}
% \end{minipage}
% }
\end{center}
\subsection{The connection between RMD and Binary classifier}\label{sec:conect_rmd_binary}

RMD is a generative model based approach which assumes the distributions of the two classes are Gaussian, while the binary classifier is a discriminative model which learns the decision boundary between two classes. 
Though they have different settings, under certain condition, the Gaussian generative model can be reduced to a binary classifier.  
To see the connection, let us assume the label $y=0$ if the sample is from in-domain, and $y=1$ if the sample is from the general domain. Let us also assume the two classes have balanced sample size without loss of generality $p(y=1)=p(y=0)$. 
Since the log-probability of $\log p(y=1|\vz)$ can be rewritten using the Bayes rule $\log p(y=1|\vz) = \log p(\vz|y=1) + \log p(y=1) - \log p(\vz)$, the logit (log odds) can be written as,
\begin{align*}
\text{logit} = \log \left(\frac{p(y=1|\vz)}{p(y=0|\vz)} \right) &= \log p(y=1|\vz) - \log p(y=0|\vz) \\
&= \log p(\vz|y=1) - \log p(\vz|y=0) \\
&= -\frac{1}{2} \left(\text{MD}(\vz; \vmu_{y=1}, \vSigma_{y=1}) - \text{MD}(\vz; \vmu_{y=0}, \vSigma_{y=0}) \right) + \texttt{const.}
% &= -\frac{1}{2} [ \left(\log|\vSigma_{y=1}|-\log|\vSigma_{y=0}|\right) - \left((\vz-\vmu_{y=1})^T \vSigma_{y=1}^{-1}(\vz-\vmu_{y=1}) - (\vz-\vmu_{y=0})^T \vSigma_{y=0}^{-1}(\vz-\vmu_{y=0}) \right) ]
\end{align*}
When $\vSigma = \vSigma_{y=1}=\vSigma_{y=0}$, the equation can be further simplified as 
\begin{align*}
\text{logit} &= \vSigma^{-1}(\vmu_{y=1}-\vmu_{y=0})^T\vz - \frac{1}{2} \left( \vmu_{y=1}^T \vSigma^{-1} \vmu_{y=1} - \vmu_{y=0}^T \vSigma^{-1} \vmu_{y=0} \right) + \texttt{const.} \\
&= \bm{\beta_1} \vz + \beta_0.
\end{align*}
Therefore, when assuming the covariance matrices are identical for the two Gaussian distributions, the Gaussian generative model can be reduced to a binary classification model. 
However, our RMD does not assume the same covariance matrix in both distributions. We estimate the covariance matrix individually for each class. So our RMD is different from binary classifier, and it has higher model capacity than the binary classifier.

\clearpage

\subsection{Comparison with more baseline methods} \label{sec:add_baselines}
\begin{table}[t]
\caption{AUROCs for OOD detection for comparing our proposed method with more baseline}
\centering
\scriptsize
\begin{tabular}{lcccccc}
\toprule
                 & \multicolumn{2}{c}{Near Shift OOD}       & \multicolumn{3}{c}{Far Shift OOD}      \\
                 \cmidrule(lr){2-3} \cmidrule(lr){4-6} 
Measure          & \texttt{cnn\_dailymail} & \texttt{newsroom} & \texttt{reddit\_tifu} & \texttt{forumsum} & \texttt{samsum} \\ \toprule
\multicolumn{6}{c}{\textsc{Input OOD}}   \\
KNN ($\alpha$=100\%, $k$=1000)        & 0.887                & 0.743                & 0.944                & 0.961                & 0.955                \\
MD                               & 0.651                & 0.799                & 0.974                & 0.977                & 0.995                \\
RMD                              & 0.828                & 0.930                & 0.998                & 0.997                & 0.999                \\
Binary logits                    & 0.997                & 0.959                & 1.000                & 0.999                & 0.998                \\ \midrule
\multicolumn{6}{c}{\textsc{Output OOD}}   \\
NLI score                        & 0.440                & 0.469                & 0.709                & 0.638                & 0.743                \\
Perplexity                       & 0.424                & 0.665                & 0.909                & 0.800                & 0.851                \\
Mean(MSP)                      & 0.343                & 0.616                & 0.877                & 0.715                & 0.826                \\
Energy score                     & 0.460                & 0.592                & 0.960                & 0.899                & 0.981                \\
Ensemble using MC dropout ($N$=5)  & 0.496                & 0.768                & 0.970                & 0.937                & 0.944                \\
Ensemble using MC dropout ($N$=10) & 0.497                & 0.774                & 0.976                & 0.947                & 0.956                \\
KNN ($\alpha$=100\%, $k$=1000)        & 0.860                & 0.791                & 0.948                & 0.926                & 0.968                \\
MD                               & 0.944                & 0.933                & 0.985                & 0.973                & 0.985                \\
RMD                              & 0.958                & 0.962                & 0.998                & 0.993                & 0.998                \\
Binary logits                    & 0.989                & 0.982                & 1.000                & 0.998                & 0.997 \\ \bottomrule
\end{tabular}
\label{tab:ood_more_baseline}
\end{table}
\revision{As we discussed in the related works, OOD detection problem was mainly studied in classification problems, and less studied in CLMs. 
Though it is not straight forward to extend classifier-based scores to CLMs especially for the input OOD detection, we would like to include as many possible methods as we can to present a comprehensive comparison for different methods. }

\revision{For those methods which rely on classification head derived logits, MSP \citep{hendrycks2016baseline}, max-logit \citep{hendrycks2019scaling}, and energy score \citep{liu2020energy}, we simply consider the output decoding process as a sequence of classifications over tokens, and take the average of the corresponding score over the generated output tokens $y_1, \dots, y_T$ as the output OOD scores. 
Therefore we added the following scores for CLMs, 
\begin{itemize}
    \item Mean(MSP) $-\frac{1}{T} \sum_{t=1}^{T} p(y_t|y_{<t}, \vx)$.
    \item Energy score $\frac{1}{T} \sum_{t=1}^T E(\vx, f_t)$, where $E(\vx, f_t) = - \tau \log \sum_{v\in V} e^{{f(y_t=v|y_{<t}, \vx)} / \tau}$, $f(y_t=v|y_{<t}, \vx)$ is the logit corresponding to the $v$-th token at the $t$-th decoding step, $V$ is the token-vocabulary, and $\tau$ is the temperature parameter. We set $\tau=1$ since the original paper \citep{liu2020energy} suggested the energy score can be used parameter-free by simply setting $\tau = 1$. 
    \item Ensemble estimation of the output perplexity from multiple Monte-Carlo dropout samples. \cite{malinin2020uncertainty,xiao2020wat} propose to turn on the MC dropout layer at the inference time and sample multiple times ($N$) using different random seeds as a way to approximate the Bayesian neural networks. We follow their idea and generate multiple output sequences and use the averaged perplexity as the uncertainty score. Note that the inference time for ensemble based method is $N$ times of that for the single model based score. 
    \item KNN-based OOD score. \cite{sun2022out} propose to use the distance to the k-th nearest neighbour in the training set in the embedding space as an OOD score. There are two hyper-parameters in the KNN-based method, $\alpha$ and $k$. $\alpha$ is the proportion of training data sampled for nearest neighbor calculation, and $k$ refers to the $k$-th nearest neighbor. We use the optimal $k=1000$ and $\alpha=100$ as suggested by the paper. We also normalize the embedding features since the paper showed the feature normalization is critical for good performance. 
\end{itemize}
Mean(MSP), energy score, and ensembled perplexity score, are all derived from the logits of the tokens in output sequences, so they are output OOD scores. The KNN-based method can be applied for both input sequence embeddings and output sequence embeddings. }

\revision{Table \ref{tab:ood_more_baseline} shows the AUROCs for OOD detection for the above newly added baselines, as a comparison to our methods. First, the logits based output OOD scores, perplexity, mean(MSP), energy score, even the ensembled perplexity score which costs $N$ times of the inference time, are in general not competitive with our proposed method RMD and Binary logits. 
Though the energy score is a bit better than perplexity and mean(MSP), and ensembled score is better than energy score, the performance gap between those methods and our proposed method is still big, especially for the near-OOD datasets. 
Second, KNN-based methods are not as good as MD and RMD either. Though it is possible that the optimal hyper-paramaters suggested by the paper may not be the optimal ones for our problem, searching for the optimal hyper-parameters requires a separate validation set. In contrast, our proposed methods have no hyperparameters.}

\subsection{Effect of the choice of the background dataset}
\begin{table}
\scriptsize
\caption{Comparison of the OOD detection performance using two different background data, \texttt{ParaCrawl} and C4 sentence. }
\begin{tabular}{lccccccccc}
\toprule
& \mc{3}{c}{WMT}  & \mc{5}{c}{OPUS}        &   \multicolumn{1}{l}{\multirow{2}{*}{MTNT}}  \\
\cmidrule(lr){2-4} \cmidrule(lr){5-9} 
Measure               & \texttt{nt2014} & \texttt{ndd2015} & \texttt{ndt2015} & \texttt{law} & \texttt{medical} & \texttt{Koran} & \texttt{IT} & \texttt{sub}  & \multicolumn{1}{l}{} \\ 
\toprule
\mc{9}{c}{\textsc{Input OOD}} \\
% MD  (Paracrawl)     & 0.534           & 0.671            & 0.670            & 0.511        & 0.704            & 0.737          & 0.828       & 0.900            & 0.668  \\
% MD  (C4 sent)     & 0.535 &	0.671 &	0.670 &	0.512 &	0.704 &	0.737 &	0.828 &	0.901 &	0.668  \\ \\
RMD (\texttt{ParaCrawl})  & 0.798           & 0.866           & 0.863            & 0.389        & \underline{0.840}            & 0.957          & \textbf{0.959}       & 0.969        & 0.943      \\
RMD (C4 sent)  & 0.833 & 	\underline{0.916} &	0.911 &	0.269 &	0.811	& 0.954	& 0.924	& 0.985	& 0.953   \\ \\
Binary logits (\texttt{ParaCrawl}) & \textbf{0.864}           & 0.904            & 0.904           & 0.485        & 0.813            & \underline{0.963}          & 0.928       & 0.950      & 0.963        \\ 
Binary logits (C4 sent)  & 0.848 &	\underline{0.916} &	\underline{0.916} &	0.285 &	0.808 &	0.944 &	0.918 &	\textbf{0.987} &	\textbf{0.976}     \\ 
\midrule
\mc{9}{c}{\textsc{Output OOD}} \\
% Perplexity (baseline) & 0.570           & 0.496            & 0.494            & 0.392        & 0.363            & 0.657          & 0.343       & 0.359        & 0.633      \\
% COMET (baseline)     & 0.484           & 0.514            & 0.525            & 0.435        & 0.543            & 0.632          & 0.619       & 0.518        & 0.724      \\
% Prism (baseline)    & 0.445           & 0.504            & 0.505            & 0.459        & 0.565            & 0.716          & 0.604       & 0.577        & 0.699      \\
% MD (Paracrawl)      & 0.609           & 0.733            & 0.739            & 0.482        & 0.784            & 0.838          & 0.900       & 0.935        & 0.794      \\
% MD (C4 sent)       & 0.611 &	0.735 &	0.741 &	0.483 &	0.782 &	0.839 &	0.900 &	0.936 &	0.794      \\ \\
RMD (\texttt{ParaCrawl})  & 0.786           & 0.858            & 0.861            & 0.355        & \textbf{0.845}            & 0.939          & \underline{0.951}       & 0.959      & 0.922        \\
RMD (C4 sent)  & 0.818	& 0.901	& 0.898	& 0.259	& \textbf{0.845} &	0.953 &	0.947 &	0.979 &	0.947      \\ \\
Binary logits (\texttt{ParaCrawl}) & 0.822           & 0.860            & 0.865            & 0.507        & 0.783            & 0.942          & 0.890       & 0.910        & 0.931      \\ 
Binary logits (C4 sent) & \underline{0.853}	& \textbf{0.925} &	\textbf{0.919} &	0.294 &	0.809 &	\textbf{0.964} &	0.901 &	\underline{0.981} &	\underline{0.975} \\ \midrule
\mc{9}{c}{\textsc{Other baselines}} \\
Input MD     & 0.534           & 0.671            & 0.670            & 0.511        & 0.704            & 0.737          & 0.828       & 0.900            & 0.668  \\
Output MD      & 0.609           & 0.733            & 0.739            & 0.482        & 0.784            & 0.838          & 0.900       & 0.935        & 0.794      \\
Perplexity  & 0.570 & 0.496 & 0.494 & 0.392 & 0.363 & 0.657 & 0.343 & 0.359 & 0.633 \\
COMET & 0.484 & 0.514 & 0.525 & 0.435 & 0.543 & 0.632 & 0.619 & 0.518 & 0.724 \\
Prism & 0.445 & 0.504 & 0.505 & 0.459 & 0.565 & 0.716 & 0.604 & 0.577 & 0.699 \\
\bottomrule
\end{tabular}
\label{tab:ood_c4sent}
\end{table}

\revision{Our principle for choosing the background data is to make it as general as possible. For summarization we use the \texttt{C4} dataset, which contains a large amount of web crawl documents, to represent a broad range of topics. Similarly for translation, we use \texttt{ParaCrawl} dataset, which is also a large web crawl of sentences, because our translation model is a sentence to sentence model, unlike the summarization model that takes the document as the input. 
% From both studies, we found the two background datasets worked well, and there was no need for further tuning on the background data. 
To further explore the effect of the background data on the performance, we split \texttt{C4} documents into sentences and use that as the background data to compute the scores, and compare that with the version using \texttt{ParaCrawl} dataset. The OOD detection performance using \texttt{C4} sentences is very similar to that using \texttt{ParaCrawl}, as shown in Table \ref{tab:ood_c4sent}. For example, \texttt{ParaCrawl}-based input OOD score has slightly better performance on \texttt{medial}, \texttt{Koran}, \texttt{IT} datasets, while \texttt{C4} based input score is slightly better at the other datasets. Both are significantly better than the baseline methods, and both give the same ranking of datasets on their OOD-ness, so our conclusion remains. Those results verify that our method is robust to the choice of background data.  }

\todo{add a sentence in the main text}

\subsection{ROC plots for the corresponding AUROC scores for OOD detection}
\begin{figure}[!htb]
\centering
\begin{subfigure}[t]{0.32\textwidth} % changed for arxiv
\centering
\caption{Input MD}
% \vspace{-0.1em}
\includegraphics[width=0.92\textwidth]{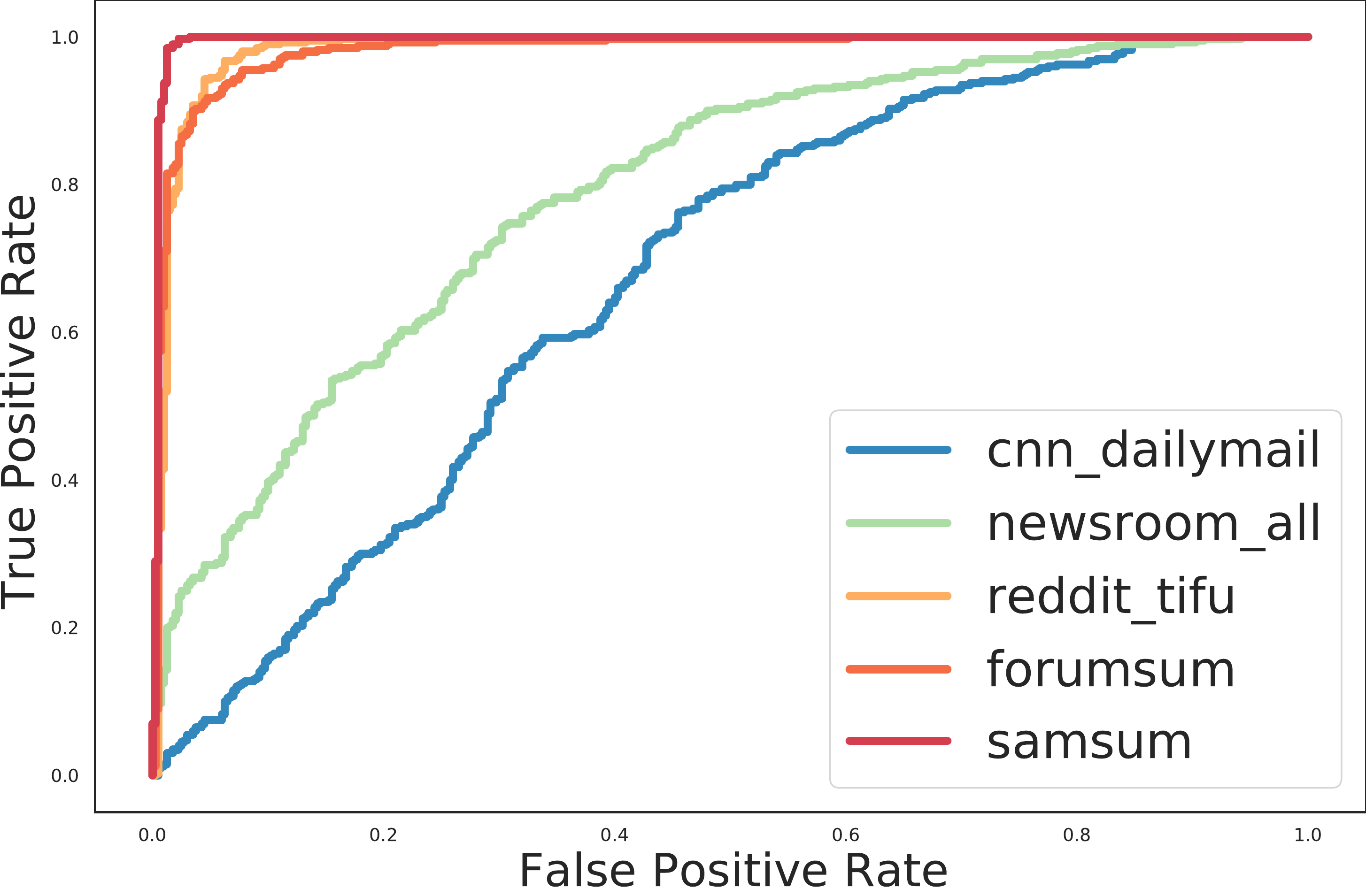} 
\end{subfigure}
\begin{subfigure}[t]{0.32\textwidth} % changed for arxiv
\centering
\caption{Input RMD}
% \vspace{-0.1em}
\includegraphics[width=0.92\textwidth]{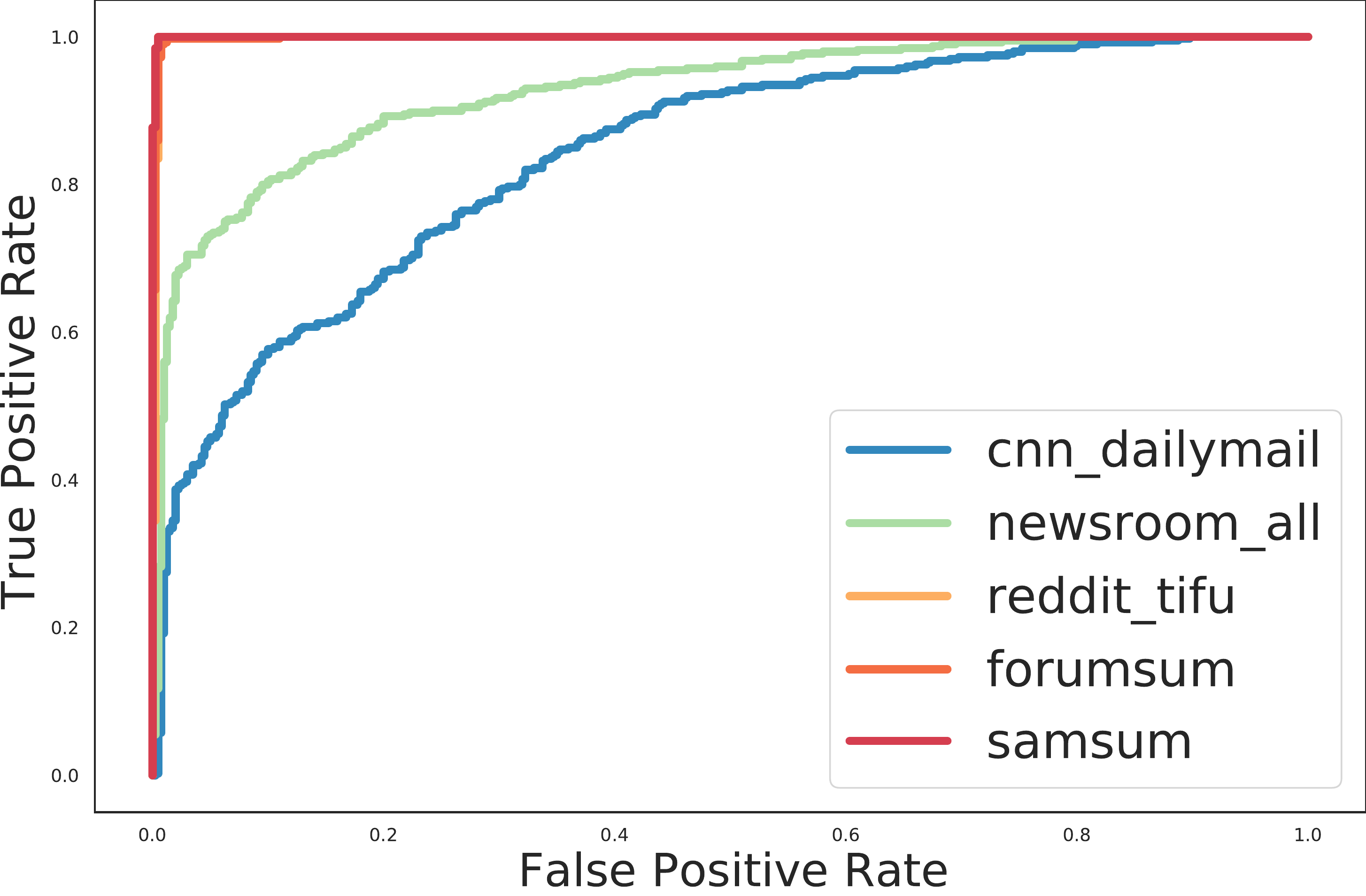} 
\end{subfigure}
\begin{subfigure}[t]{0.32\textwidth} % changed for arxiv
\centering
\caption{Input Binary logits}
% \vspace{-0.1em}
\includegraphics[width=0.92\textwidth]{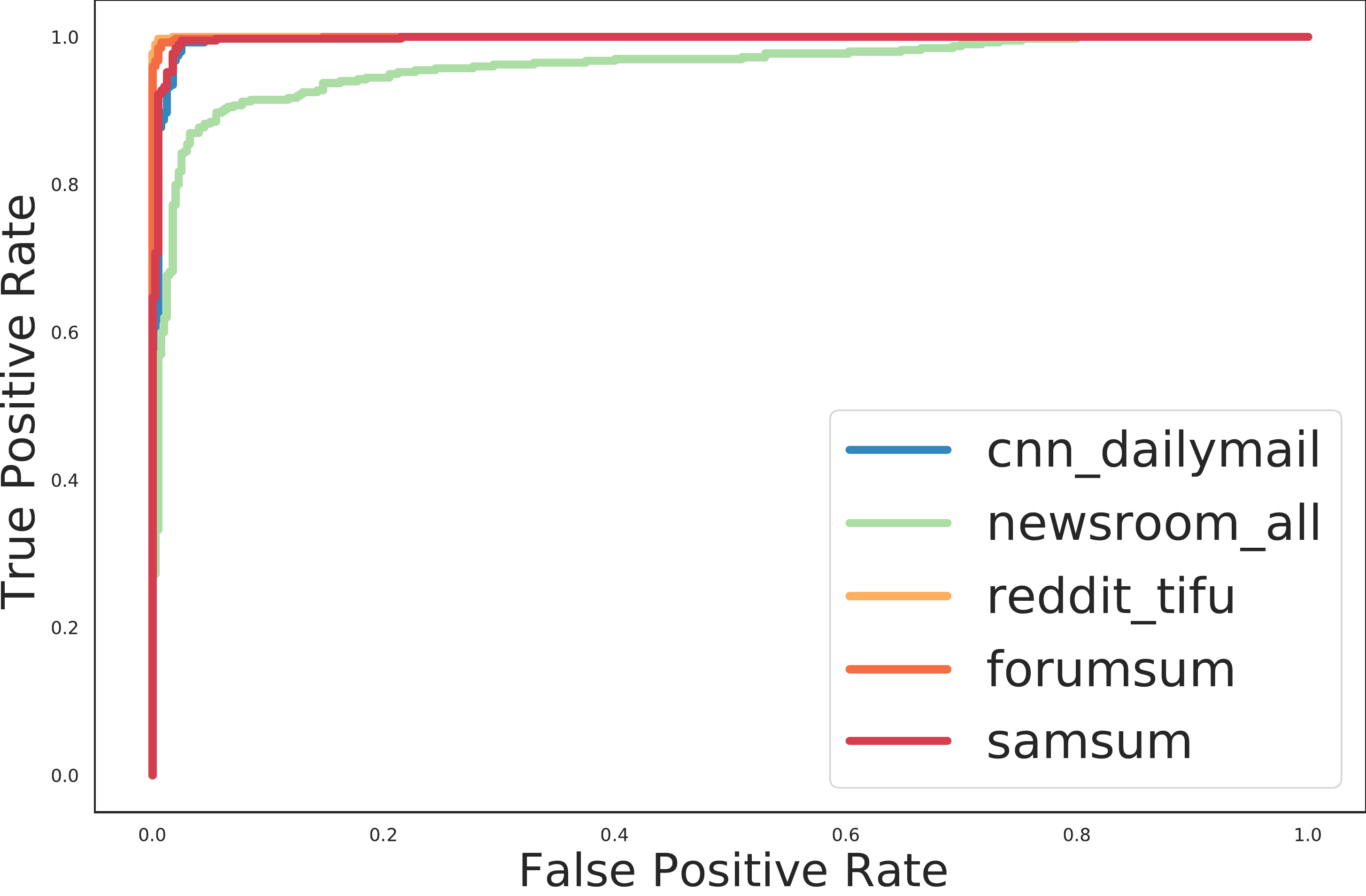} 
\end{subfigure} 
\begin{subfigure}[t]{0.32\textwidth} % changed for arxiv
\centering
\caption{Output MD}
% \vspace{-0.1em}
\includegraphics[width=0.92\textwidth]{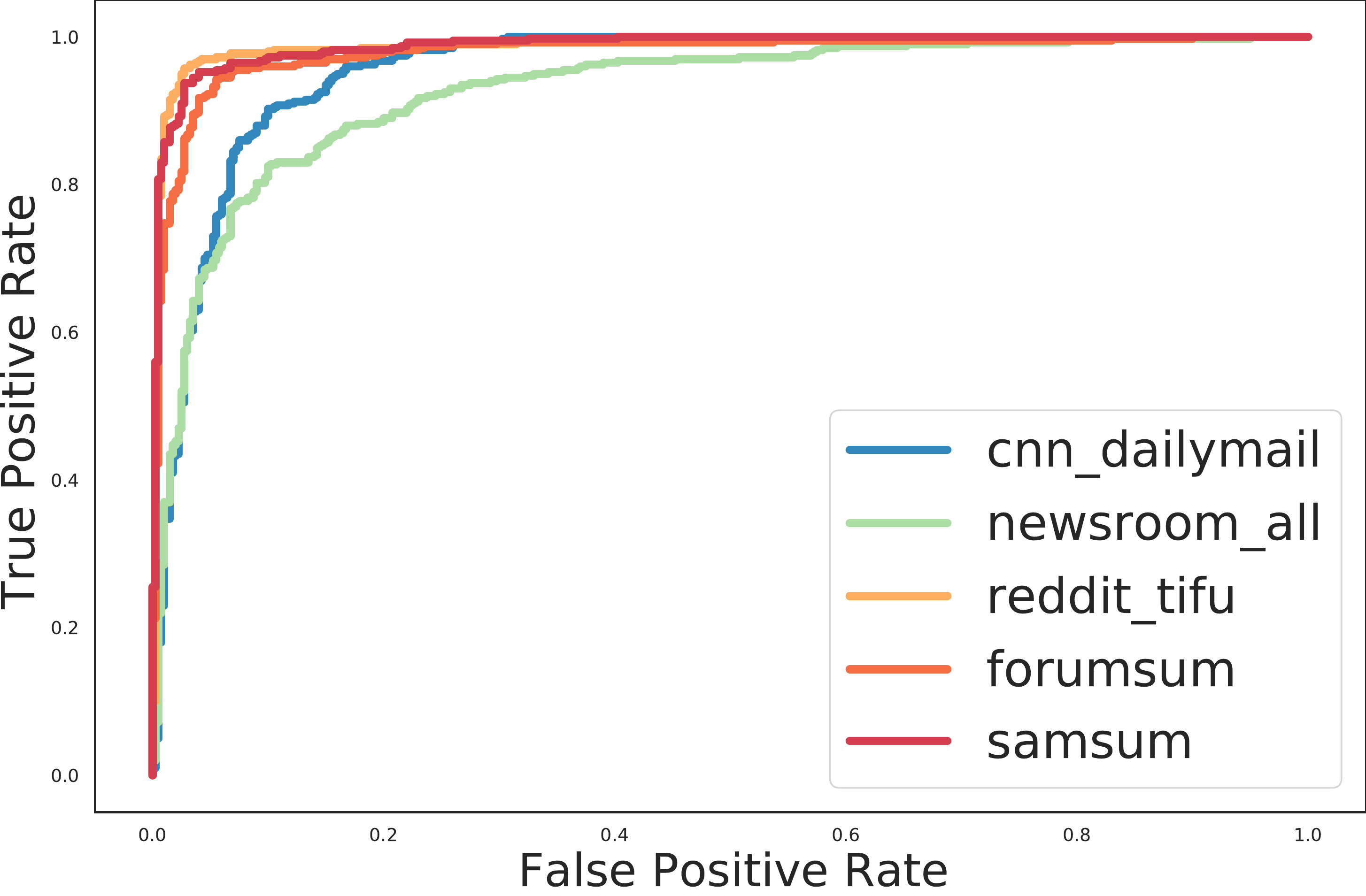} 
\end{subfigure}
\begin{subfigure}[t]{0.32\textwidth} % changed for arxiv
\centering
\caption{Output RMD}
% \vspace{-0.1em}
\includegraphics[width=0.92\textwidth]{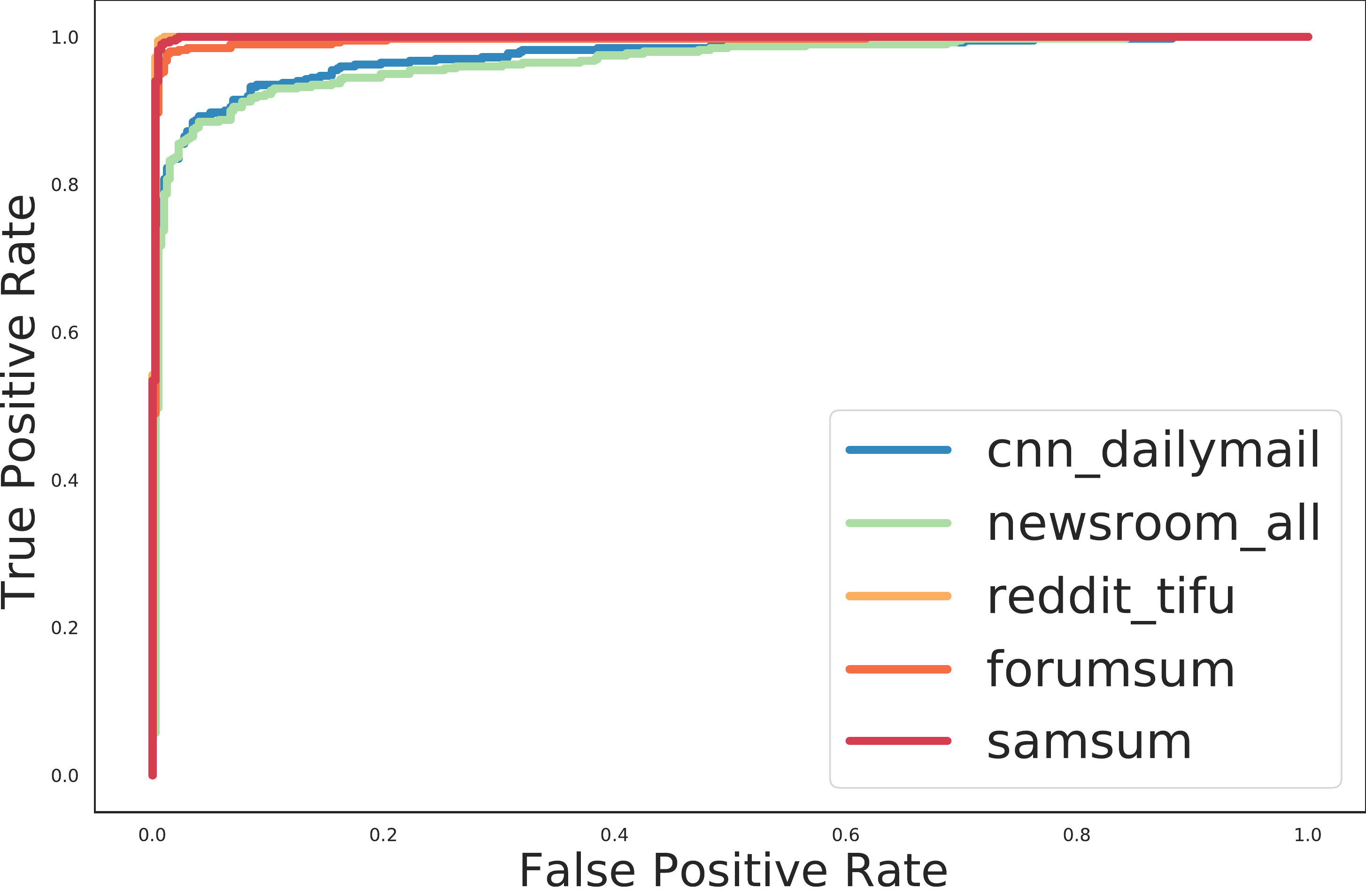} 
\end{subfigure}
\begin{subfigure}[t]{0.32\textwidth} % changed for arxiv
\centering
\caption{Output Binary logits}
% \vspace{-0.1em}
\includegraphics[width=0.92\textwidth]{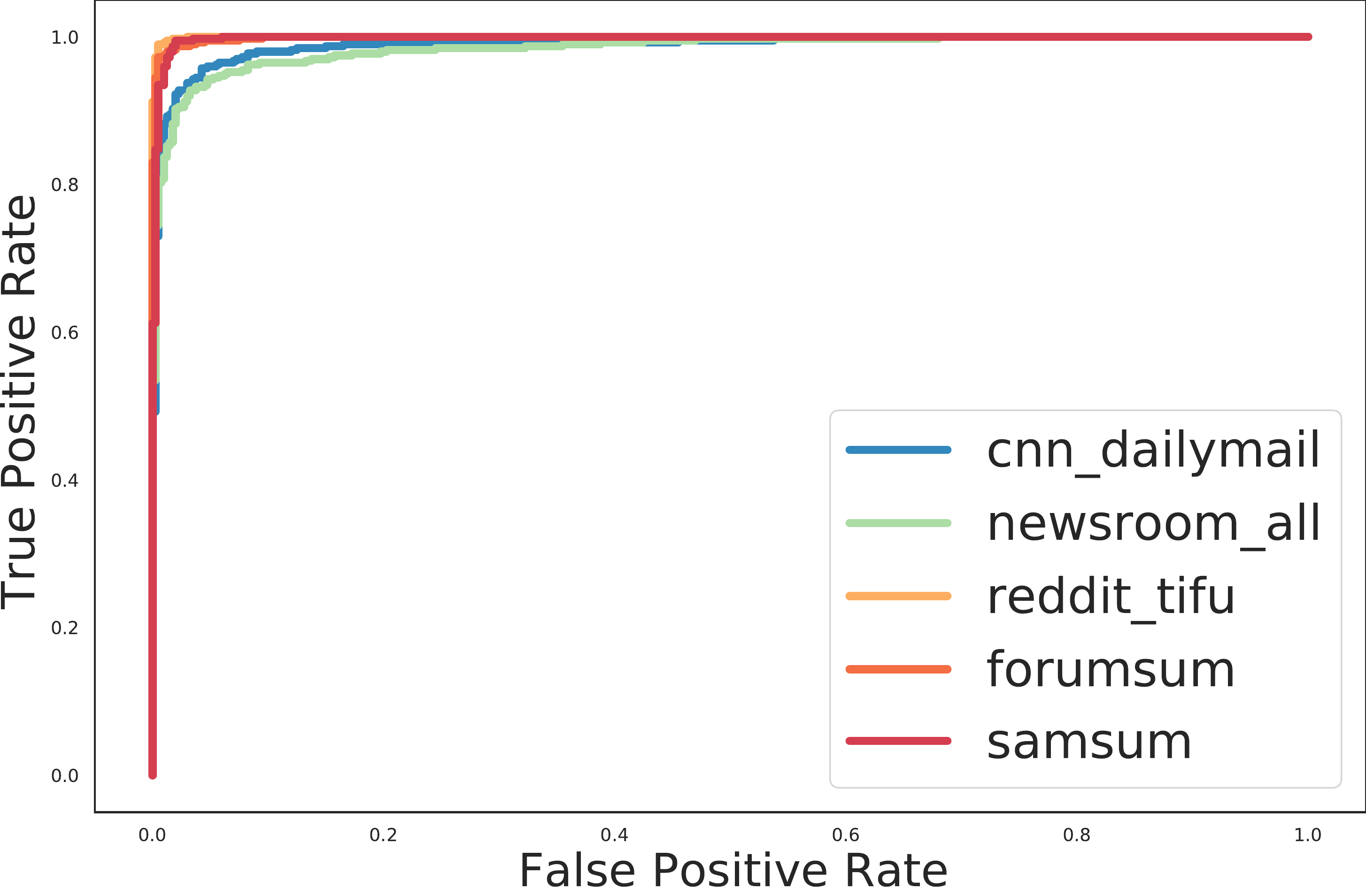} 
\end{subfigure} 
\begin{subfigure}[t]{0.32\textwidth} % changed for arxiv
\centering
\caption{Perplexity}
% \vspace{-0.1em}
\includegraphics[width=0.92\textwidth]{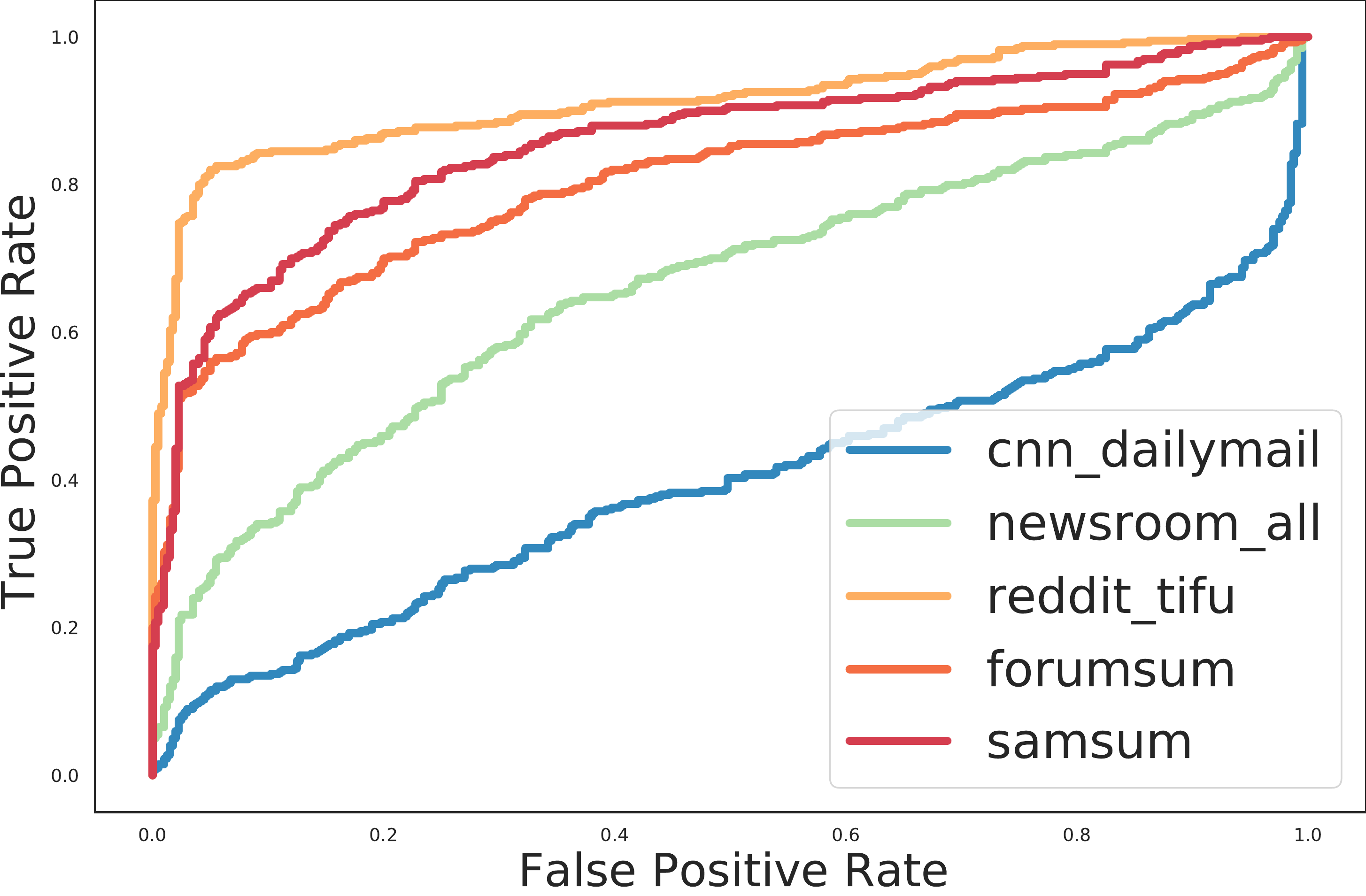} 
\end{subfigure} 
\begin{subfigure}[t]{0.32\textwidth} % changed for arxiv
\centering
\caption{Mean(MSP)}
% \vspace{-0.1em}
\includegraphics[width=0.92\textwidth]{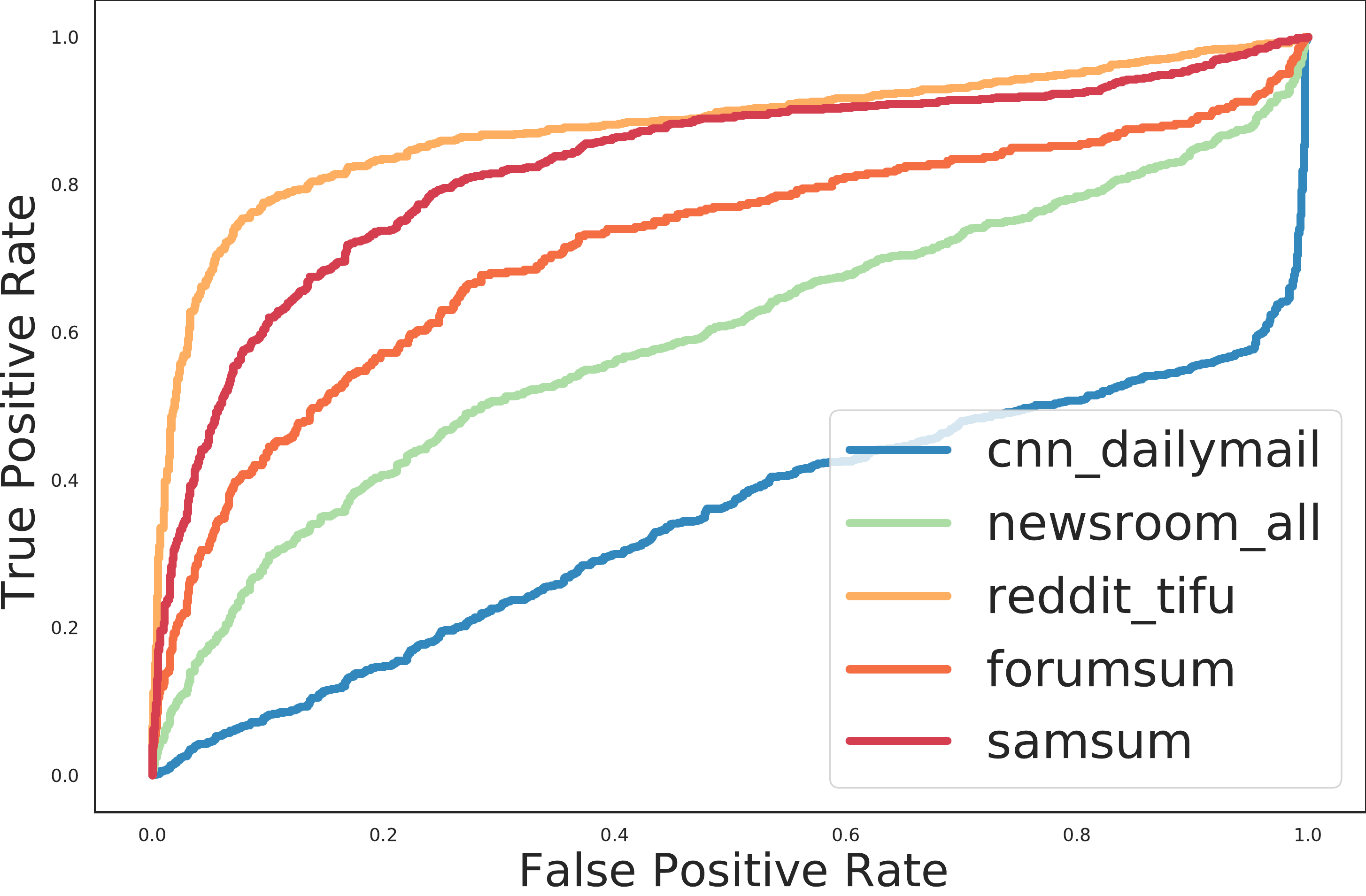} 
\end{subfigure} 
\begin{subfigure}[t]{0.32\textwidth} % changed for arxiv
\centering
\caption{NLI score}
% \vspace{-0.1em}
\includegraphics[width=0.92\textwidth]{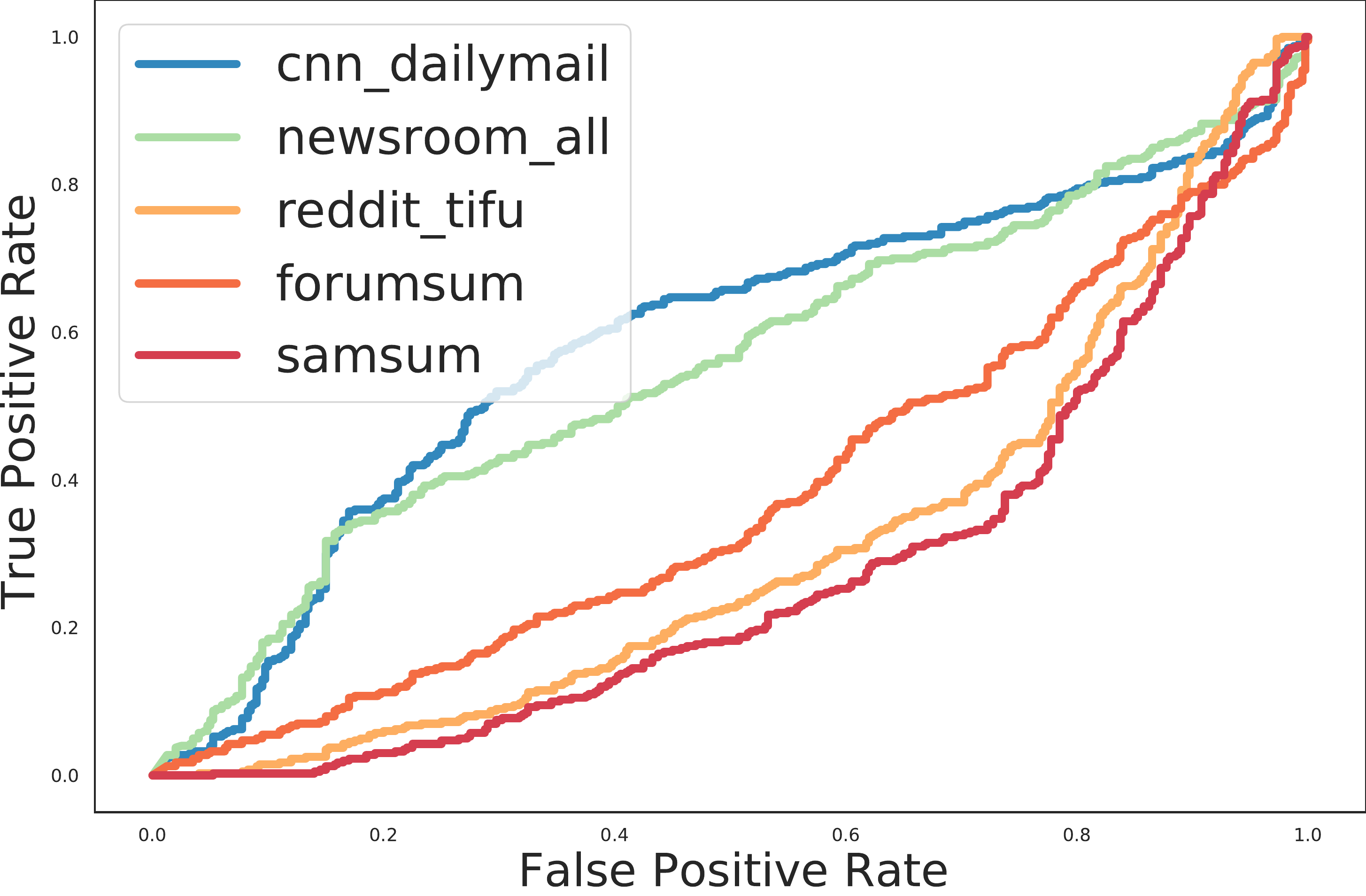} 
\end{subfigure} 
% \vspace{-0.1em}
% \vspace{-0.8em}
\caption{ROC plots for the corresponding AUROC scores in Table \ref{tab:ood_aucroc} for OOD detection in summarization
}
\label{fig:rocs}
\end{figure}

\revision{
To better visualize the OOD detection performance, we present Figure \ref{fig:rocs} to show the ROC plots for the corresponding AUROC scores for OOD detection in Table \ref{tab:ood_aucroc}. 
Each of the OOD measures is used for separating the in-domain test data as negative and the OOD test data as positive sets. The AUROC is defined as the area under the ROC curves. 
The closer an ROC curve is to the upper left corner, the larger the AUROC value is. AUROC 1.0 means a perfect separation, and 0.5 means the two are not distinguishable. AUROC is independent of the choice of threshold, so it can be used for fair comparisons among methods. }

\subsection{Correlation between different scores and the quality metrics}

\begin{table}[h]
\caption{Kendall’s $\tau$ correlation ($\text{p-value}<0.05$ are greyed out) between various measures with human-judged quality of a PEGASUS \texttt{xsum} model decoded on summarization datasets.  The ``All'' column shows the correlation when examples from all datasets are included. Note for negatively correlated scores (e.g. perplexity, OOD score), we take the negative value of the score for easier comparison. A few intra-dataset correlations have $\text{p-value}<0.05$ due to the small sample size (only 100 examples per dataset were sent for human evaluation). }
\resizebox{\textwidth}{!}{
\begin{tabular}{lccccc}
% \tiny
                 & \multicolumn{1}{c}{In-domain} & \multicolumn{1}{c}{Near Shift OOD}       & \multicolumn{2}{c}{Far Shift OOD}      \\
Measure         &  \texttt{xsum} & \texttt{cnn\_dailymail} & \texttt{reddit\_tifu} & \texttt{samsum} & All \\ \toprule
\multicolumn{6}{c}{Single Score} \\ \midrule    
\multicolumn{6}{c}{\textsc{Input OOD}}  \\
MD                                        & \textcolor{light-gray}{0.044}  & \textcolor{light-gray}{-0.018} & \textcolor{light-gray}{-0.017} & \textcolor{light-gray}{0.133} & 0.328 \\
RMD                               & \textcolor{light-gray}{0.015}  & \textcolor{light-gray}{-0.033} & \textcolor{light-gray}{0.017}  & \textcolor{light-gray}{0.133} & 0.336 \\
Binary Logits                              & \textcolor{light-gray}{-0.022} & \textcolor{light-gray}{-0.061} & \textcolor{light-gray}{0.028}  & \textcolor{light-gray}{0.106} & 0.233 \\ \midrule
\multicolumn{6}{c}{\textsc{Output OOD}}                                                                                                       \\
\rowcolor{Gray}
Perplexity (baseline)                                 & 0.256  & 0.186  & \textcolor{light-gray}{0.081}  & \textcolor{light-gray}{0.068} & 0.300 \\
NLI score (baseline)                                   & 0.337  & 0.308  & 0.226  & \textcolor{light-gray}{0.132} & 0.381 \\
MD                                  & \textcolor{light-gray}{0.106}  & \textcolor{light-gray}{-0.055} & \textcolor{light-gray}{0.202}  & 0.352 & 0.384 \\
% \rowcolor{light-gray}
RMD                                & \textcolor{light-gray}{0.053}  & 0.177  & 0.214  & 0.314 & 0.385 \\
Binary logits                       & 0.199  & \textcolor{light-gray}{-0.100} & \textcolor{light-gray}{0.091}  & \textcolor{light-gray}{0.026} & 0.213 \\ \midrule
\multicolumn{6}{c}{Combined Score}  \\ \midrule
PR sum (perplexity, input RMD)                    & 0.186  & \textcolor{light-gray}{0.134}  & \textcolor{light-gray}{0.082}  & \textcolor{light-gray}{0.109} & 0.358 \\
PR sum (perplexity, output RMD)                   & \textcolor{light-gray}{0.250}  & 0.350  & 0.168  & 0.237 & \underline{0.415} \\
PR sum (perplexity, input \& output RMD)           & 0.171  & 0.242  & 0.158  & 0.250 & 0.401 \\
PR sum (perplexity, input binary logits)          & 0.214  & \textcolor{light-gray}{0.079}  & \textcolor{light-gray}{0.126}  & \textcolor{light-gray}{0.090} & 0.322 \\
PR sum (perplexity, output binary logits)         & 0.347  & \textcolor{light-gray}{0.086}  & \textcolor{light-gray}{0.114}  & \textcolor{light-gray}{0.052} & 0.330 \\
PR sum (perplexity, input \& output binary logits) & 0.277  & \textcolor{light-gray}{0.003}  & \textcolor{light-gray}{0.127}  & \textcolor{light-gray}{0.096} & 0.307 \\
Lineare regression (perplexity, input \& output)                                       & 0.235  & 0.402  & 0.170  & 0.250 & \textbf{0.422} \\ \bottomrule
\end{tabular}
}
\label{tab:corr_heval_sum}
\end{table}

\begin{table}[h]
    \caption{Kendall $\tau$ correlation ($\text{p-value}<0.05$ are grayed out) between various measures and quality measured by \bleurt on translation datasets. For easier comparison, we negate the signs of the coefficients for measures that are expected to have negative correlation with BLEURT (e.g., OOD score). Within the same dataset, perplexity shows good correlation, but it deteriorates (with the exception of \texttt{MTNT}) as we move to more OOD datasets such as \texttt{Koran}.}
    \resizebox{\textwidth}{!}{
        \begin{tabular}{lccccccccccc}
        \toprule
        & \mc{4}{c}{\texttt{WMT}} & \mc{5}{c}{\texttt{OPUS}} & & \\
        \cmidrule(lr){2-5} \cmidrule(lr){6-10}
        \multirow{-2}{*}{Measure} & \texttt{holdout} & \texttt{nt2014} & \texttt{ndd2015} & \texttt{ndt2015} & \texttt{law} & \texttt{medical} & \texttt{Koran} & \texttt{IT} & \texttt{sub} &  \multirow{-2}{*}{\texttt{MTNT}} & \multirow{-2}{*}{All} \\
        \midrule
        \mc{11}{c}{Single Score} \\
        \midrule
        \mc{11}{c}{\textsc{Input OOD}} \\
        MD & \llap{-}0.081 & \llap{-}0.131 & \llap{-}0.129 & \llap{-}0.117 & \llap{-}0.171 & \textcolor{light-gray}{0.041} & \llap{-}0.147 & \llap{-}0.093 & \textcolor{light-gray}{0.012} & -0.117 & \textcolor{light-gray}{0.007} \\
        RMD & 0.147 & 0.091 & 0.049 & 0.115 & 0.197 & \textcolor{light-gray}{0.013} & \llap{-}0.071 & \llap{-}0.060 & 0.098 & 0.083 & 0.195 \\
        Binary logits & 0.144 & 0.116 & 0.141 & 0.162 & 0.124 & \textcolor{light-gray}{\llap{-}0.003} & \textcolor{light-gray}{0.025} & \llap{-}0.071 & 0.104 & 0.161 & 0.202 \\
        \midrule
        \mc{11}{c}{\textsc{Output OOD}} \\
        \rowcolor{Gray}
        Perplexity (baseline) & 0.309 & 0.337 & 0.352 & 0.375 & 0.389 & 0.224 & 0.222 & 0.225 & 0.227 & 0.341 & 0.286 \\
        COMET (baseline) & 0.184 & 0.397 & 0.402 & 0.443 & 0.324 & 0.253 & 0.359 & 0.174 & 0.297 & 0.414 & 0.336 \\
        Prism (baseline) & 0.184 & 0.329 & 0.337 & 0.342 & 0.179 & 0.188 & 0.192 & 0.151 & 0.286 & 0.370 & 0.301\\
        MD & \textcolor{light-gray}{\llap{-}0.029} & \llap{-}0.066 & \llap{-}0.064 & \llap{-}0.048 & \llap{-}0.096 & \textcolor{light-gray}{0.032} & \llap{-}0.105 & \llap{-}0.057 & \textcolor{light-gray}{0.041} & \textcolor{light-gray}{-0.020}  & 0.083 \\
        RMD & 0.086 & 0.049 & 0.044 & 0.095 & 0.135 & \textcolor{light-gray}{\llap{-}0.026} & \llap{-}0.077 & \llap{-}0.056 & 0.061 & 0.077 & 0.170 \\
        Binary logits & 0.106 & 0.058 & 0.075 & 0.114 & 0.094 & \textcolor{light-gray}{\llap{-}0.036} & \textcolor{light-gray}{\llap{-}0.013} & \llap{-}0.059 & \textcolor{light-gray}{\llap{-}0.012} & 0.075 & 0.151 \\
        \midrule 
        \mc{11}{c}{Combined Score} \\
        \midrule
        RR sum (perplexity, input RMD) & 0.321 & 0.361 & 0.351 & 0.410 & 0.382 & 0.230 & 0.161 & 0.154 & 0.261 & 0.354 & \textbf{0.361} \\
        PR sum(perplexity, output RMD) & 0.323 & 0.357 & 0.359 & 0.414 & 0.371 & 0.200 & 0.152 & 0.164 & 0.240 & 0.350 & \underline{0.356} \\
        PR sum(perplexity, input \& output RMD) & 0.291 & 0.284 & 0.264 & 0.329 & 0.346 & 0.119 & 0.082 & 0.084 & 0.231 & 0.290 & 0.311 \\
        PR sum(perplexity, input binary logits) & 0.323 & 0.352 & 0.372 & 0.384 & 0.391 & 0.195 & 0.211 & 0.111 & 0.234 & 0.359 & 0.335 \\
        PR sum(perplexity, output binary logits) & 0.318 & 0.302 & 0.314 & 0.350 & 0.356 & 0.168 & 0.162 & 0.127 & 0.156 & 0.293 & 0.299 \\
        PR sum(perplexity, input \& output binary logits) & 0.300 & 0.262 & 0.288 & 0.309 & 0.340 & 0.125 & 0.145 & 0.053 & 0.163 & 0.287 & 0.288 \\
        Linear regression (perplexity, input \& output) & 0.318 & 0.370 & 0.355 & 0.414 & 0.383 & 0.243 & 0.180 & 0.119 & 0.268 & 0.367 & 0.352 \\ \bottomrule
        \end{tabular}
    }
    \label{tab:quality_corr_MT}
\end{table}
\clearpage

\subsection{Selective generation and output quality prediction}
\begin{figure}[!htb]
\centering
% \begin{subfigure}[t]{0.30\textwidth} % changed for arxiv
%     \centering
%     \caption{Perplexity vs RMD output OOD}
%     % \vspace{-0.1em}
%     \includegraphics[width=1\textwidth]{figures/2d_ppx_vs_output_ood.png}
% \end{subfigure} 
% \vspace{0.8em}
\begin{subfigure}[t]{0.45\textwidth} % changed for arxiv
\centering
\caption{\rougeone vs Abstention Rate}
% \vspace{-0.1em}
\includegraphics[width=1\textwidth]{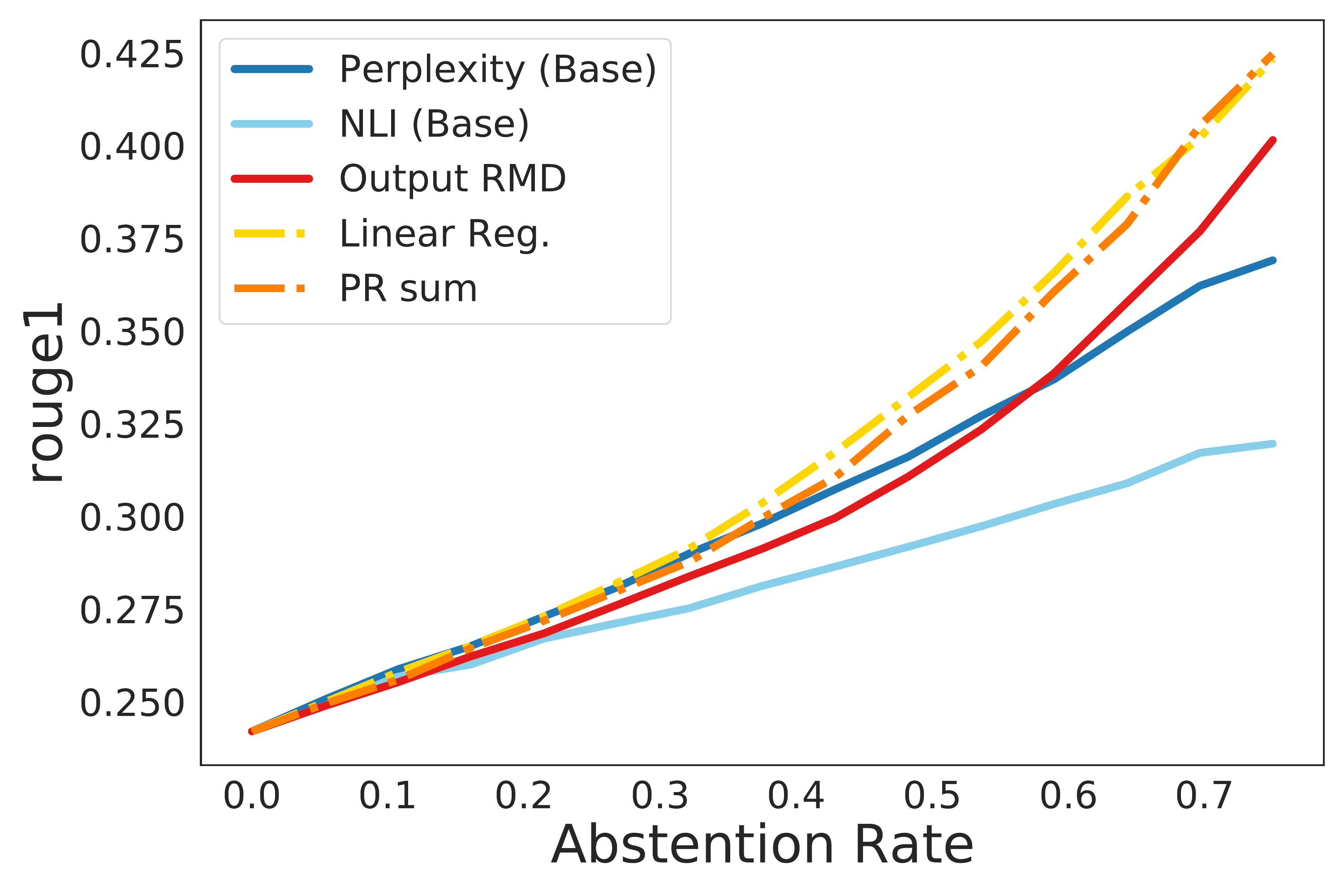} % s_abstain_vs_quality_methods8.pdf}
\end{subfigure} 
% \vspace{0.8em}
\begin{subfigure}[t]{0.45\textwidth} % changed for arxiv
\centering
\caption{Survival count per dataset}
% \vspace{0.6em}
\includegraphics[width=1\textwidth]{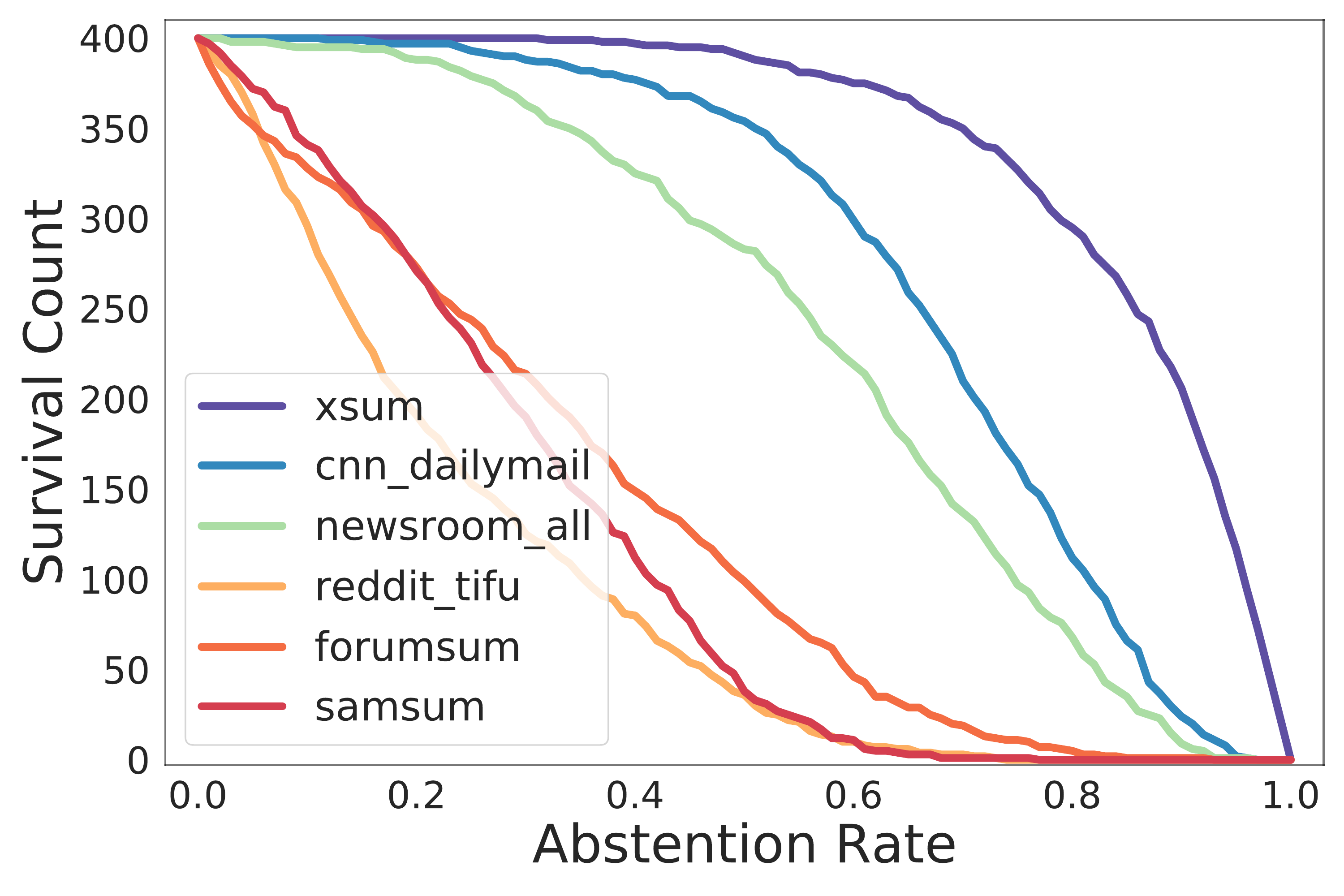}
\end{subfigure} 
% \vspace{-0.1em}
% \vspace{-0.8em}
\caption{(a)  The summarization quality \rougeone vs abstention curve for single scores, including input and output RMD OOD scores, output perplexity score, and NLI score, and combined scores, including linear regression machine learning model, percentile sum of RMD OOD scores and perplexity score. The corresponding area under the curve is in Table \ref{table:abstention_AUC_sum_rouge1}. (b) The survival count of each dataset as the joint dataset is abstained. Each dataset is sub-sampled to 400 examples for this analysis. }
\label{fig:abs_vs_quality_rouge1}
\end{figure}

\begin{figure}[h]
\centering
% \vspace{-0.1em}
\includegraphics[width=0.5\textwidth]{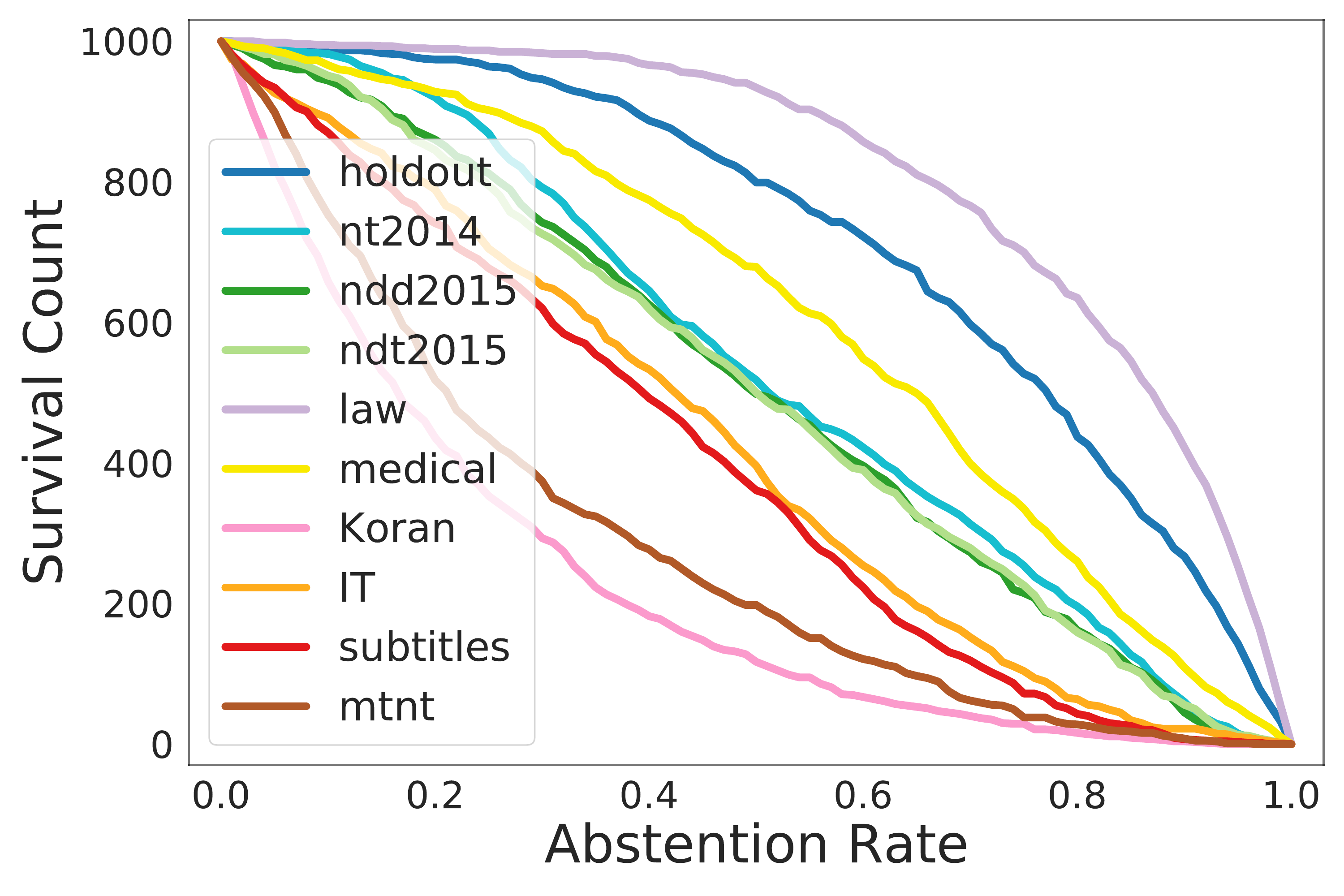} % s_abstain_vs_quality_methods8.pdf}
\caption{The translation survival count of each dataset as the joint dataset is abstained. Complete results for Figure \ref{fig:abs_vs_quality} (d).}
\end{figure}

\begin{table}[h]
\caption{Area under the quality (\texttt{human eval}) vs abstention curve for summarization for various single scores and the proposed combined scores. }
\scriptsize
\center
\begin{tabular}{lc}
Measure                                        & Area under the quality (human eval) vs abstention curve \\ \toprule
\multicolumn{2}{c}{Single Score} \\ \midrule    
\multicolumn{2}{c}{Input OOD}                                                             \\
MD                              & 0.464                                      \\
RMD                        & 0.466                                      \\
Binary logits                         & 0.445                                      \\
\multicolumn{2}{c}{Output OOD}                                                            \\ 
Perplexity (baseline)                        & 0.458                                      \\
NLI score (baseline)                         & 0.469                                      \\
MD                              & 0.469                                      \\
RMD                         & 0.474                                      \\
Binary logits                         & 0.441                                      \\ \midrule
\multicolumn{2}{c}{Combined Score}                                                        \\ \midrule
\prsum (perplexity, input RMD)                     & 0.468                                      \\
\prsum (perplexity, output RMD)                    & \underline{0.478}                                      \\
\prsum (perplexity, input \& output RMD)           & 0.476                                      \\
\prsum (perplexity, input binary logits)           & 0.461                                      \\
\prsum (perplexity, output binary logits)          & 0.461                                      \\
\prsum (perplexity, input \& output binary logits) & 0.456                                      \\
Linear regression (perplexity, input \& output RMD)                     & \textbf{0.481}                                      \\ \bottomrule 
\end{tabular}
\label{table:abstention_AUC_sum_heval}
\end{table}

\begin{table}[h]
\caption{Area under the quality (\rougeone) vs abstention curve for summarization for various single scores and the proposed combined scores. }
\scriptsize
\center
\begin{tabular}{lc}
Measure                                        & Area under the quality (\texttt{rouge1}) vs abstention curve \\ \toprule
\multicolumn{2}{c}{Single Score} \\ \midrule    
\multicolumn{2}{c}{Input OOD}                                                             \\
MD                              & 0.208                                      \\
RMD                        & 0.214                                      \\
Binary logits                        & 0.217                                      \\
\multicolumn{2}{c}{Output OOD}                                                            \\ 
Perplexity (baseline)                        & 0.221                                      \\
NLI score (baseline)                         & 0.207                                      \\
MD                             & 0.219                                      \\
RMD                        & 0.221                                      \\
Binary logits                         & 0.207                                      \\ \midrule
\multicolumn{2}{c}{Combined Score}                                                        \\ \midrule
\prsum (perplexity, input RMD)                     & 0.222                                      \\
\prsum (perplexity, output RMD)                    & \underline{0.228}                                      \\
\prsum (perplexity, input \& output RMD)           & 0.224                                      \\
\prsum (perplexity, input binary logits)           & 0.225                                      \\
\prsum (perplexity, output binary logits)          & 0.221                                      \\
\prsum (perplexity, input \& output binary logits) & 0.220                                      \\
Linear regression (perplexity, input \& output RMD)                     & \textbf{0.229}                                      \\ \bottomrule 
\end{tabular}
\label{table:abstention_AUC_sum_rouge1}
\end{table}

\begin{table}[h]
\caption{Area under the quality (\bleurt) vs abstention curve for translation using various single scores and the proposed combined scores.
% \todo{need to update with new MTNT}
}
\scriptsize
\center
\begin{tabular}{lc}
Names                                        & Area under the quality vs abstention curve \\ \toprule
\multicolumn{2}{c}{Single Score} \\ \midrule    
\multicolumn{2}{c}{Input OOD}                                                             \\ 
MD                              & 0.583                                      \\
RMD                         & 0.623                                      \\
Binary logits                         & 0.621                                      \\
\multicolumn{2}{c}{Output OOD}                                                            \\
Perplexity (baseline)                        & 0.627                                      \\
Comet (baseline)                                       & 0.644                                      \\
Prism (baseline)                                     & 0.638                                      \\
MD                             & 0.601                                      \\
RMD                         & 0.618                                      \\
Binary logits                        & 0.608                                      \\
\midrule
\multicolumn{2}{c}{Combined Score}                                                        \\ \midrule
\prsum (perplexity, input RMD)                     & \textbf{0.647}                                      \\
\prsum (perplexity, output RMD)                    & \underline{0.646}                                      \\
\prsum (perplexity, input \& output RMD)           & 0.641                                      \\
\prsum (perplexity, input binary logits)           & 0.639                                      \\
\prsum (perplexity, output binary logits)          & 0.632                                      \\
\prsum (perplexity, input \& output binary logits) & 0.633                                      \\
Linear regression (ppx, input \& output)                     & 0.645           \\
% \prsum (comet, input RMD) &	0.647 \\
% \prsum (prism, input RMD)	0.641 \\
% \prsum (comet, input RMD)	0.644 \\
% \prsum (prism, output RMD)	0.638 \\ 
\bottomrule                           
\end{tabular}
\label{table:abstention_AUC_translate}
\end{table}

% \begin{figure}[h]
%     \centering
%     \begin{subfigure}[t]{0.455\textwidth} % changed for arxiv
%         \centering
%         \caption{RMD vs Binary logits output OOD score}
%         % \vspace{-0.1em}
%         \includegraphics[width=\textwidth]{figures/2d_scatter_ood_score_output_vs_binary_logits_output.pdf} 
%     \end{subfigure} 
%     \vspace{-0.8em}
%     \begin{subfigure}[t]{0.48\textwidth} % changed for arxiv
%         \centering
%         \caption{RMD input vs RMD output OOD score}
%         % \vspace{-0.1em}
%         \includegraphics[width=\textwidth]{figures/2d_scatter_ood_score_output_vs_ood_score_input.pdf}
%     \end{subfigure} 
%     % \vspace{-0.1em}
%     \vspace{-0.8em}
%     \caption{Comparison between (a) RMD vs Binary logits OOD scores (b) RMD input vs output OOD scores. Comparing RMD with Binary logits, binary OOD score is more sensitive to dataset shift than RMD so that it does not have a good distinction between near shift OOD datasets \texttt{cnn\_dailymail} and \texttt{newsroom} and far OOD datasets \texttt{reddit\_tifu}, \texttt{forumsum}, and \texttt{samsum}. RMD has the resolution for distinguishing near-OOD and far-OOD, such that the near-OOD datasets have scores distributed in the middle of in-domain and far-OOD. Comparing Input OOD and output OOD score, }
% \end{figure}

\subsection{Investigation of the n-gram overlap between law dataset and in-domain datasets} 

\begin{table}[h]
\center
\scriptsize
\begin{tabular}{lccccc}
\toprule
\multirow{2}{*}{domain/split} & \multirow{2}{*}{overall average} & \mc{4}{c}{$n$-gram overlap} \\ 
\cmidrule(lr){3-6}                               
& & $n=1$ & $n=2$ & $n=3$ & $n=4$ \\ 
\midrule
\texttt{holdout} & \textbf{8.3} & \underline{45.4} & \textbf{16.8} & \textbf{4.8} & \textbf{1.3} \\
\texttt{nt2014} & 4.9 & 39.0 & 12.3 & 2.7 & 0.5 \\
\texttt{ndd2015} & 5.1 & 40.7 & 12.9 & 2.7 & 0.5 \\
\texttt{ndt2015} & 4.6 & 39.0 & 12.8 & 2.6 & 0.3 \\
\rowcolor{Gray}\texttt{law} & \underline{7.7} & \textbf{48.8} & \underline{16.1} & \underline{4.2} & \underline{1.1} \\
\texttt{medical} & 4.3 & 33.5 & 10.7 & 2.4 & 0.4 \\
\texttt{Koran} & 2.8 & 32.6 & 8.7 & 1.4 & 0.2 \\
\texttt{IT} & 4.0 & 35.9 & 10.6 & 2.2 & 0.3 \\
\texttt{sub} & 2.8 & 38.6 & 10.9 & 1.4 & 0.1 \\ 
\texttt{MTNT} & 2.5 & 31.4 & 8.4 & 1.2 & 0.1 \\
\bottomrule
\end{tabular}
% \raisebox{-0.5\height}{\includegraphics[width=0.4\textwidth]{figures/translation/abstain_vs_quality_ngram.pdf}}
\caption{$n$-gram overlap analysis between the various test sets including \texttt{law} and the in-domain training data, 
we observe that \texttt{law} has the highest unigram overlap rate (48.8\%) and the second highest overall overlap (defined as the geometric mean) with the in-domain data.
% \todo{add MTNT, explain what is overall}
}
\label{tab:translate_ngram}
\end{table}

%\clearpage

% \begin{figure}[h]
% \centering
% % \begin{subfigure}[t]{0.45\textwidth} % changed for arxiv
% % \centering
% % \caption{Survival count per dataset}
% % % \vspace{-0.1em}
% % \includegraphics[width=1\textwidth]{figures/translation/survivorship_line_full_addmtnt.png} % s_abstain_vs_quality_methods8.pdf}
% % \end{subfigure} 
% % % \vspace{0.8em}
% % \caption{Survival count per dataset}
% \vspace{0.6em}
% \includegraphics[width=0.5\textwidth]{figures/translation/abstain_vs_quality_ngram.pdf}
% \caption{Quality vs abstention curve based on $n$-gram overlap rate as the abstention score.}
% \end{figure}

%\clearpage

\subsection{Quantitative analysis using n-gram overlap to determine near- and far-OOD datasets in summarization} 
\revision{
To support our claim that the news related test datasets, \texttt{cnn\_dailymail} and \texttt{newsroom} are closer to the in-domain \texttt{xsum} than the other dialogue datasets \texttt{reddit\_tifu}, \texttt{samsum}, and \texttt{forumsum}, we compute the $n$-gram overlap between each of the test datasets and the in-domain dataset.
We use Jaccard similarity score, $J(\mathcal{A}, \mathcal{B})=\frac{|\mathcal{A} \cap \mathcal{B}|}{|\mathcal{A} \cup \mathcal{B}|}$, where $\mathcal{A}$ and $\mathcal{B}$ are the set of $n$-gram in dataset $A$ and dataset $B$, to measure the similarity between two datasets. 
Table \ref{tab:summary_ngram} shows the similarity scores based on $1-4$ grams. It is clear to see that \texttt{cnn\_dailymail} and \texttt{newsroom} have significantly higher similarity with the in-domain \texttt{xsum} data than other three datasets.
Therefore, we call the news-related datasets \textit{near}-OOD and the other dialogue based datasets \textit{far-} OOD.}

\begin{table}[h]
\center
\scriptsize
\begin{tabular}{lccccc}
\toprule
\multirow{2}{*}{domain/split} & \multirow{2}{*}{overall average} & \mc{4}{c}{$n$-gram overlap} \\ 
\cmidrule(lr){3-6}                               
& & $n=1$ & $n=2$ & $n=3$ & $n=4$ \\ 
\midrule
\texttt{xsum}           & 7.3 &	32.4 &	13.3 &	4.6 &	1.4 \\
\rowcolor{Gray}
\texttt{cnn\_dailymail} & 6.2 &	31.1 &	12.7 &	4.0 &	0.9  \\
\rowcolor{Gray}
\texttt{newsroom}  & 5.3 &	28.8 &	11.1 &	3.3	& 0.7    \\
\texttt{reddit\_tifu}   & 2.8 &	17.2 &	6.9	& 1.8 &	0.3    \\
\texttt{forumsum} & 2.7	& 18.0 &	6.5 &	1.6 &	0.3      \\
\texttt{samsum} & 1.2 &	10.4 &	3.1 &	0.7 &	0.1   \\
\bottomrule
\end{tabular}
% \raisebox{-0.5\height}{\includegraphics[width=0.4\textwidth]{figures/translation/abstain_vs_quality_ngram.pdf}}
\caption{Jaccard similarity based on $n$-gram overlap between the various test sets and the in-domain \texttt{xsum} training data. 
We observe that the news-related datasets \texttt{cnn\_dailymail} and \texttt{newsroom} have significantly higher similarity scores with the in-domain \texttt{xsum} data than the other three OOD datasets \texttt{reddit\_tifu}, \texttt{forumsum}, and \texttt{samsum}.
% \todo{add MTNT, explain what is overall}
}
\label{tab:summary_ngram}
\end{table}

\subsection{Visualization of OOD score on shifted dataset}
\label{sec:visualization}
\begin{figure}[h]
\centering
\includegraphics[width=0.9\textwidth]{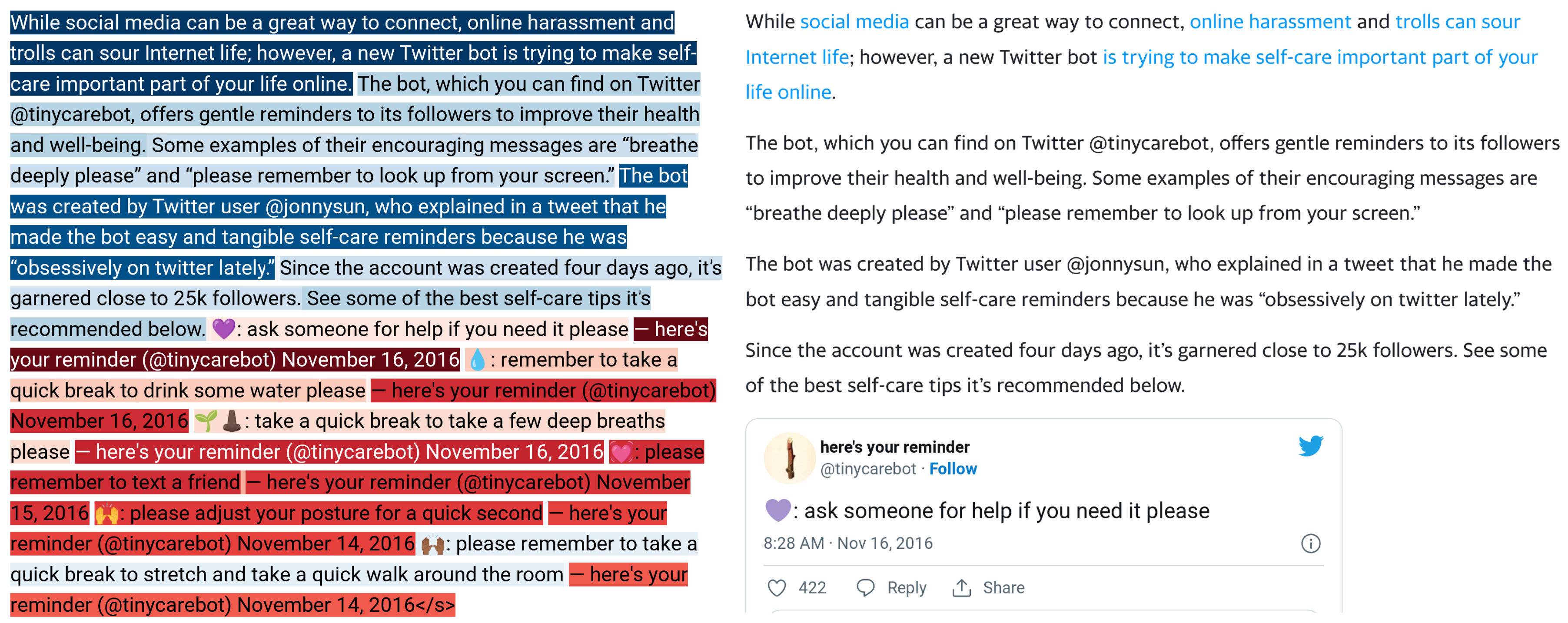}
\caption{OOD score can be attributed to individual sentences to highlight the out-of-domain noisy parts of text (red denotes out-of-domain and blue denotes in-domain text), e.g. tweets present in articles scraped from internet. Example taken from Newsroom dataset.}
\label{fig:ood_viz}
\end{figure}

We explore how individual parts of an input text contribute to the OOD score, which can help us visualize which parts of the text are OOD.
We define the OOD score of each sentence in the text using a leave-one-out strategy:
For any given sentence, we compute the 
OOD score of the article with and without that sentence in it.
The negative of the change in the OOD score after removing the sentence denotes the OOD score of that sentence. Intuitively, if removing the sentence decreases the overall OOD score, that sentence is assigned a positive OOD score and vice-versa.
Figure~\ref{fig:ood_viz} illustrates an example where an article contains noise in the form of tweets with emojis, and the OOD scoring mechanism described above assigns positive OOD scores to those tweets and negative scores to the main text.

\clearpage
\newpage

\subsection{Summarization examples with low/ high predicted quality scores}
\label{sec:cnn_example}

\revision{Besides the quantitative results, here we show a few real examples to better demonstrate how well our predicted quality score helps for selective generation on out-of-distribution examples. 
The model here was fine-tuned on \xsum but inference was run on examples from \cnn.}

\revision{Figure \ref{fig:low_quality_ex_1}, \ref{fig:low_quality_ex_2}, and \ref{fig:low_quality_ex_3} show 3 examples in \texttt{cnn\_dailymail} that have the highest \prsum (perplexity, output RMD) scores that predict for low quality summaries.} 

\revision{Figure \ref{fig:high_quality_ex_1}, \ref{fig:high_quality_ex_2}, and \ref{fig:high_quality_ex_3} show 3 examples in \texttt{cnn\_dailymail} that have the lowest \prsum (perplexity, output RMD) scores that predict for high quality summaries.}

%% bad examples
\begin{figure}[th]
\begin{mdframed}
\small
    \textbf{Document:} A man trying to elude police jumped into a Missouri creek overnight wearing only his underwear – but his daring gambit did not pay off. Responding officers and firefighters followed the fugitive into the murky waters of Brush Creek in Kansas City and fished him out early Friday morning. The 38-year-old suspect has been taken to an area hospital to be treated for injuries to his arm and leg. He may face charges in connection to a hit-and-run crash. Escape by water: A 38-year-old man stripped down to his skivvies and jumped into Brush Creek in Kansas City, Missouri, after being stopped by police. Up Brush Creek without a paddle: The suspect reached the middle of the creek and spent 10-15 minutes swimming back and forth. According to a Kansas City Police Department’s arrest report, officers were called to a gas station in the 4600 block of Prospect at around 2am after receiving complaints from neighbors about a car blasting loud music. The report states that when police approached the car, a grey 2007 Infinity, and asked to see the driver’s license, the man smiled, said, ‘I’m out!’ and took off from the scene. The Infinity promptly smashed into the north side of the Brush Creek bridge, after which the driver got out of the mangled car and jumped into the water. Police say the 38-year-old suspect stripped down to his underwear and spent 10-15 minutes swimming in chest-deep water, with officers waiting for him on north and south sides of the creek. Surrounded: When firefighters tried to pull him out, he threatened them with a log. Fish out of water: Police officers armed with a BB gun went after the nighttime bather and apprehended him. The bather was complaining of a broken leg, according to Fox4KC, so the Kansas City Fire Department’s water rescue crew were sent in to fish him out. But the half-naked man in the water was not going to go quietly. ‘The suspect picked up a large log and started swinging it at the firemen so they backed off as to not escalate the situation,’ the arrest report states. That is when uniformed police officers armed with a BB gun followed the man into the creek, got him in a choke hold and pulled him out of the creek. Police suspect the man may have been under the influence of drugs or alcohol. Prelude: Before he jumped in the water, the 38-year-old driver fled from police and smashed his 2007 Infinity into a bridge. Police suspect the man may have been under the influence of drugs or alcohol at the time. As of Friday morning, the 38-year-old has not been formally charged with any crime. \\
    \textbf{Reference Summary:} The 38-year-old suspect was questioned by Kansas City police after neighbors complained he was blasting music in his 2007 Infinity. 
Instead of handing over his ID, driver smiled, said 'I'm out!' and took off.
After crashing into bridge, the man stripped down to his underwear and jumped into Brush Creek.
It took cops armed with a BB gun 15 minutes to fish out the fugitive. \\
    \textbf{Model Summary:} All images are copyrighted. \\
    \textbf{Human rating score ($\uparrow$ means high quality):} 0.2 \\
    \textbf{\prsum (perplexity, output RMD) ($\downarrow$ means high quality): } 0.67
\normalsize
\end{mdframed}
\caption{Examples in \texttt{cnn\_dailymail} that have the highest \prsum (perplexity, output RMD) scores that predict for low quality summaries.}
\label{fig:low_quality_ex_1}
\end{figure}

\begin{figure}[th]
\begin{mdframed}
\small
    \textbf{Document:} A crisp fan who gets through 42 bags in a week has discovered a skull-shaped deep-fried potato snack in one of his packets. Barry Selby, 54, who lives with his dog in Poole, Dorset, was eating a bag of cheese and onion crisps when he made the bizarre discovery, which appears to be a profile of a human skull. The floor-fitter has decided to keep the two inches tall by two-and-a-half inches wide snack as he believes it is far more impressive than other oddly-shaped examples he has seen on the internet. Scroll down for video. Spooky find: Barry Selby was eating a bag of Tesco cheese and onion crisps when he found the 'skull' snack. Mr Selby said: 'I was shocked when I found it. I was just eating a bag of cheese and onion crisps from Tesco and when I pulled it out it did take me back a bit. 'I thought it was worth keeping as I don't think I will ever find one like it again. It must have been a very weird-shaped potato. 'It's about two inches tall and two-and-a-half inches wide and it's in perfect detail, it even has an eye socket. 'I sometimes give my dog, Max, crisps in a bowl, so it's lucky he didn't have this packet or I wouldn't have found it. Weird snack: Mr Selby has decided to keep the unusual find, which appears to show a jaw, nose and eye. Comparison: The 54-year-old said he was 'shocked' to make the discovery, although it is not his first. In the 1990s he came across a 3D heart-shaped crisp, which he kept until it broke. And it's not the first odd-shaped snack he has come across - in the 1990s he found a crisp shaped like a 3D heart, which he kept for several years until it broke. But he says this find was different: 'This one was a big one. I just thought "wow" and wanted to share it. 'I've been keeping it on top of my computer in the front room, but it should be in a protective box really. 'I'm going to keep it forever, it's just so spooky. I looked on the internet for other funny-shaped crisps but this is a one-off.' \\
    \textbf{Reference Summary:} Barry Selby from Dorset was eating bag of Tesco cheese and onion crisps.
The 54-year-old discovered a snack shaped like profile of the human skull.
He said he was 'shocked' with the find and has decided to 'keep it forever'
It's not his first weird food find - he once discovered a heart-shaped crisp. \\
    \textbf{Model Summary:} All images are copyrighted. \\
    \textbf{Human rating score ($\uparrow$ means high quality):} 0.2 \\
    \textbf{\prsum (perplexity, output RMD) ($\downarrow$ means high quality):} 0.66
\normalsize
\end{mdframed}
\caption{Examples in \texttt{cnn\_dailymail} that have the highest \prsum (perplexity, output RMD) scores that predict for low quality summaries.}
\label{fig:low_quality_ex_2}
\end{figure}

\begin{figure}[th]
\begin{mdframed}
\small
    \textbf{Document:} Last week she was barely showing – but Demelza Poldark is now the proud mother to the show’s latest addition. Within ten minutes of tomorrow night’s episode, fans will see Aidan Turner’s dashing Ross Poldark gaze lovingly at his new baby daughter. As Sunday night’s latest heartthrob, women across the country have voiced their longing to settle down with the brooding Cornish gentleman – but unfortunately it seems as if his heart is well and truly off the market. Scroll down for video. Last week she was barely showing – but Demelza Poldark is now the proud mother to the show’s latest addition. He may have married his red-headed kitchen maid out of duty, but as he tells her that she makes him a better man, audiences can have little doubt about his feelings. What is rather less convincing, however, is the timeline of the pregnancy. With the climax of the previous episode being the announcement of the pregnancy, it is quite a jump to the start of tomorrow’s instalment where Demelza, played by Eleanor Tomlinson, talks about being eight months pregnant. Just minutes after – once again without any nod to the passing of time – she is giving birth, with the last month of her pregnancy passing in less than the blink of an eye. With the climax of the previous episode being the announcement of the pregnancy, it is quite a jump to the start of tomorrow’s instalment where Demelza, played by Eleanor Tomlinson, talks about being eight months pregnant. As Sunday night’s latest heartthrob, women across the country have voiced their longing to settle down with Poldark – but unfortunately it seems as if his heart is well and truly off the market. Their fast relationship didn't go unnoticed by fans. One posted on Twitter: ‘If you are pregnant in Poldark times expect to have it in the next 10 minutes’ It is reminiscent of the show’s previous pregnancy that saw Elizabeth, another contender for Ross’s affection, go to full term in the gap between two episodes. This didn’t go unnoticed by fans, who posted on Twitter: ‘Poldark is rather good, would watch the next one now. Though if you are pregnant in Poldark times expect to have it in the next 10 minutes.’ \\
    \textbf{Reference Summary:} SPOILER ALERT: Maid gives birth to baby on Sunday's episode.
Only announced she was pregnant with Poldark's baby last week. \\
    \textbf{Model Summary:} It’s all change in the world of Poldark. \\
    \textbf{Human rating score ($\uparrow$ means high quality):}  0.4\\
    \textbf{\prsum (perplexity, output RMD) ($\downarrow$ means high quality):}  0.62
\normalsize
\end{mdframed}
\caption{Examples in \texttt{cnn\_dailymail} that have the highest \prsum (perplexity, output RMD) scores that predict for low quality summaries.}
\label{fig:low_quality_ex_3}
\end{figure}

%%%%%% good examples

\begin{figure}[th]
\begin{mdframed}
\small
    \textbf{Document:} Rangers boss Stuart McCall says he is already working on a dossier of signing targets for next season - even though he may not be around to parade them. The interim Ibrox manager still does not know if he will be in charge beyond the current campaign after being lured back to his old club to kick-start their faltering promotion bid. So far, everything is going to plan with Gers second in the Scottish Championship table and destined for a semi-final play-off slot. Stuart McCall says he is already looking at transfer targets for next season, though he may not be at Rangers. But with 12 players out of contract, McCall knows the Light Blues will need to strengthen if they have any chance of keeping pace with rivals Celtic next season - if they go up - and is already piecing together a wish list of potential new arrivals. He said: 'I've been speaking to a lot of agents and putting things in place for if and when... Even if I'm not here, if I'm getting players put to me who would like to come to Rangers regardless of the manager, then we build a little portfolio of positions that we will be needing next year. 'It's not a case of us standing still and then thinking come June 1, 'Oh we need to get into action'. 'No, there are a lot of agents who come to us and we build a little dossier of players that as a staff, we think will be good for next season, regardless of what league we are in. 'It would be slightly naive [if we were not doing that]. If I'm in charge or not, I still want the club to do well and I will put my view across to the board on who I think should be coming into the club and who should be here.' McCall is compiling a dossier on targets as he looks to put the club in the best possible position. Rangers have operated a haphazard transfer policy since re-emerging from the embers of liquidation. The club's team of scouts were jettisoned under the disastrous Craig Whyte regime and former boss Ally McCoist was largely forced to turn to a list of former Ibrox servants he had personal knowledge of when trying to bolster his squad. But McCall revealed the club's new board are now starting the process of re-establishing their spying network - albeit on a smaller level than before. 'I think there has been discussions behind the scenes with different people,' said the former Motherwell boss. 'I don't think we are at the stage where we were 10 or 15 years ago where we were aiming to get into the Champions League and bringing players in for three and four million yet. 'I don't think Rangers will be at the stage yet next year where we need international scouts everywhere. Rangers have expanded their scouting network after a haphazard system over the past few years. 'But certainly a scouting network needs to be put in place. 'Having said that, I spoke to Craig Levein at Hearts and they do a lot of their scouting with [online service] Wyscout. When I brought Henrik Ojamaa in at Motherwell, that was after I'd seen a clip of him on YouTube. I sold him for £350,000 after signing him for nothing. That was great. 'So you can still do your own background work. Personally I would always like to see the player myself. I've only ever signed one player without watching him first and slightly regretted it. 'So yeah we need a scouting network but at this moment where Rangers are, not to the extent where we have scouts all over Europe.' McCall admitted he still does not know if he will rejoin Gordon Strachan's Scotland staff for the June 13 Euro 2016 qualifier with Ireland in Dublin. And he also confessed to uncertainties ahead of Saturday's match with Falkirk. McCall's side are still in line for promotion, sitting in the play-off positions in the Scottish Championship. Peter Houston's Bairns - five points behind fourth-placed Queen of the South with two games to play - need an unlikely series of results to make the play-offs but McCall says that raises more questions than answers. He said: 'Housty is a wily old fox who has done terrifically well in his career so I don't know what to expect. 'It will take a difficult set of results for them to get into the play-offs so I don't know if they will come here and think the pressure is off and play care free. 'They don't lose many goals so we may have to be patient through the 90 minutes. We have had a couple of decent results against them but they have capable players and we will need to be at our best.'\\
    \textbf{Reference Summary:} Rangers are currently second in the Scottish Championship.
Stuart McCall's side are in pole position to go up via the play-offs.
But McCall is still not certain of his future at the club next season.
Rangers boss says he is still trying to build the squad for next year.
Rangers have begun to expand their scouting after several poor years. \\
    \textbf{Model Summary:} Stuart McCall says he is already looking at transfer targets for next season, though he may not be at Rangers. \\
    \textbf{Human rating score ($\uparrow$ means high quality):} 0.8 \\
    \textbf{\prsum (perplexity, output RMD) ($\downarrow$ means high quality):} 0.10
\normalsize
\end{mdframed}
\caption{Examples in \texttt{cnn\_dailymail} that have the lowest \prsum (perplexity, output RMD) scores that predict for high quality summary.}
\label{fig:high_quality_ex_1}
\end{figure}

\begin{figure}[th]
\begin{mdframed}
\small
    \textbf{Document:} An Alberta student who'd accidentally left his headlights on all day was greeted by what may have been the world's friendliest note from a stranger when he returned to his car. But Derek Murray, a University of Alberta law student, found more than just the note that cold November day in Edmonton--he also found an extension cord and battery charger left by the stranger to bring his dead Acura back to life. Now that Murray's life-affirming tale has now gone viral, he says 'It just shows you how such a pure act of kindness from one person can just spread through everyone and help make everyone’s day a little brighter.' Good Samaritan: A friendly stranger left this unbelievably friendly letter to Alberta law student Derek Murray in order to help him get his car started after he left the headlights on all day. At first, though, he assumed the letter was from an angry fellow motorist, he told the National Post. 'When I first saw the note, I was expecting it to be an angry letter from someone telling me not to park there. Instead, I got someone just totally brightening my day. My day could have been ruined but, because of this guy, it was the highlight of my day.' The note reads, in part:. I noticed you left your lights on. The battery will probably not have enough charge to start your vehicle. I left a blue extension cord on the fence and a battery charger beside the fence in the cardboard box. If you know how to hook it up, use it to start your car. What followed was a detailed explanation of how to use the equipment. 'Sure enough,' Derek recalled to the National Post, 'I looked over at the house my car was parked beside, and there was a blue extension cord plugged into an outlet behind the guy’s house with a battery charger right there beside it.' Derek was able to get his car started, but when he rang the good Samaritan's doorbell, there was no answer. So, Derek left his own note as a thank you for the kind gesture. He later snapped a photo of the stranger's friendly note to post to Facebook, where it has now gone viral. The note has been viewed millions of times and even Edmonton Mayor Don Iveson retweeted the photo. Derek snapped a photo of the note for Facebook and it has since gone viral. e 'It just shows you how such a pure act of kindness from one person can just spread through everyone and help make everyone’s day a little brighter,’ Derek said. \\
    \textbf{Reference Summary:} Derek Murray, a University of Alberta law student, could have had his day ruined by the mistake by a stranger's kindness brightened it up.
Murray posted his story and the note online and the random act of kindness has now gone viral. \\
    \textbf{Model Summary:} A Canadian student who accidentally left his headlights on all day was greeted by what may have been the world's friendliest note from a stranger when he returned to his car. \\
    \textbf{Human rating score ($\uparrow$ means high quality):}  0.8\\
    \textbf{\prsum (perplexity, output RMD) ($\downarrow$ means high quality):} 0.11
\normalsize
\end{mdframed}
\caption{Examples in \texttt{cnn\_dailymail} that have the lowest \prsum (perplexity, output RMD) scores that predict for high quality summary.}
\label{fig:high_quality_ex_2}
\end{figure}

\begin{figure}[th]
\begin{mdframed}
\small
    \textbf{Document:} Bayern Munich had to make do without FOUR important first-team stars as Pep Guardiola's side attempted to overturn a 3-1 deficit against Porto on Tuesday night. Injured quartet Franck Ribery, Mehdi Benatia, David Alaba and Arjen Robben were forced to watch on from the sidelines as the German giants bid to reach the Champions League semi-finals. However, the absence of Robben and Co appeared to make no difference as Bayern raced into a 5-0 lead at half-time before claiming a 6-1 victory to win the tie 7-4 on aggregate. Injured trio Franck Ribery, Mehdi Benatia and David Alaba chat ahead of Bayern's clash with Porto. Injured Ribery acknowledges a steward before taking a seat at the Allianz Arena on Tuesday night. Ribery looks on as former Roma defender Benatia chats with the France international in the dugout. While Ribery, Benatia and Alaba chatted in the home dugout ahead of kick-off, Holland international Arjen Robben was in front of the mic doing some punditry alongside Bayern goalkeeping legend Oliver Kahn. Ribery missed the game after failing to recover from a recent ankle injury while former Roma defender Benatia faces another two weeks out with a groin problem. Robben was unavailable for the encounter with an abdominal injury. David Alaba, meanwhile, is set for a month on the sidelines having partially ruptured knee ligaments playing for Austria at the start of April. Bayern had just 14 fit players to choose from against Porto in the first leg but tore the Portuguese giants apart at the Allianz Arena to progress. Holland international Arjen Robben was pictured doing punditry alongside Bayern legend Oliver Kahn (right) Bayern Munich wideman Robben was unavailable for the Champions League clash with an abdominal injury. \\
    \textbf{Reference Summary:} Bayern Munich beat Porto 6-1 at the Allianz Arena on Tuesday night.
German giants were without Franck Ribery, David Alaba and Mehdi Benatia.
Arjen Robben was also sidelined and did some punditry for the tie. \\
    \textbf{Model Summary:} Arjen Robben, Mehdi Benatia, Franck Ribery and David Alaba all missed Bayern Munich's Champions League quarter-final second leg against Porto. Holland international Arjen Robben was pictured doing punditry alongside Bayern legend Oliver Kahn (right) Bayern Munich wideman Robben was unavailable for the Champions League clash with an abdominal injury. \\
    \textbf{Human rating score ($\uparrow$ means high quality):} 0.8\\
    \textbf{\prsum (perplexity, output RMD) ($\downarrow$ means high quality):} 0.11
\normalsize
\end{mdframed}
\caption{Examples in \texttt{cnn\_dailymail} that have the lowest \prsum (perplexity, output RMD) scores that predict for high quality summary.}
\label{fig:high_quality_ex_3}
\end{figure}

\end{document}